\documentclass{article}

\usepackage{amsmath}
\usepackage{natbib}
\usepackage[table]{xcolor}

\usepackage{microtype}
\usepackage{graphicx}
\usepackage{booktabs} %

\usepackage{hyperref}

\usepackage{microtype}
\usepackage{graphicx}
\usepackage{booktabs} %

\usepackage{mathtools}
\usepackage{algorithm}
\usepackage{tikz}
\usetikzlibrary{arrows,automata,positioning}
\usepackage{multirow} 
\usepackage{xspace}

\usepackage{comment}
\usepackage{amssymb} %

\usepackage{dsfont}
\usepackage{tabulary}
\usepackage{tabularx}
\usepackage{makecell}
\usepackage{wrapfig}

\usepackage{dsfont}
\usepackage{tabulary}
\usepackage{makecell}
\usepackage{wrapfig}
\usepackage{subcaption}

\usepackage{amssymb}
\usepackage{mathtools}
\usepackage{amsthm}
\usepackage{array}

\usepackage[capitalize,noabbrev]{cleveref}

\usepackage{titletoc}

\usepackage{booktabs}
\usepackage{colortbl}
\usepackage{tabularx} %
\usepackage{pifont}

\theoremstyle{plain}

\theoremstyle{definition}

\theoremstyle{remark}

\newcommand{\maxfu}[1]{{\color{blue}{}}}
\newcommand{\jyu}[1]{{\color{orange}{}}}
\newcommand{\raven}[1]{{\color{red}{}}}
\newcommand{\karim}[1]{{\color{purple}{}}}
\newcommand{\ethan}[1]{{\color{green}{}}}
\newcommand{\haoru}[1]{{\color{pink}{}}}
\newcommand{\wenli}[1]{{\color{yellow}{}}}
\newcommand{\yuke}[1]{{\color{purple}{}}}

\newcommand{\sysname}{CaP-Agent0\xspace}
\newcommand{\benchname}{CaP-Bench\xspace}
\newcommand{\gymname}{CaP-Gym\xspace}
\newcommand{\rlname}{CaP-RL\xspace}

\definecolor{orange}{rgb}{1,0.5,0}
\definecolor{lightsalmonpink}{rgb}{1.0, 0.6, 0.6}
\definecolor{verylightsalmonpink}{rgb}{0.966, 0.805, 0.797}
\definecolor{lightblue}{rgb}{0.862, 0.906, 0.984}
\definecolor{lightyellow}{rgb}{1.0, 0.945, 0.797}
\definecolor{lightgreen}{rgb}{0.835, 0.91, 0.828}
\definecolor{lightpurple}{rgb}{0.879, 0.832, 0.902}

\newcommand{\fsize}{small}
\usepackage[font=\fsize,labelfont=bf,skip=5pt]{caption}
\usepackage[export]{adjustbox}
\usepackage{wrapfig}

\usepackage{ragged2e}

\usepackage{listings}

\definecolor{codegreen}{rgb}{0,0.6,0}
\definecolor{codegray}{rgb}{0.5,0.5,0.5}
\definecolor{codepurple}{rgb}{0.58,0,0.82}
\definecolor{backcolour}{rgb}{0.95,0.95,0.92}

\lstdefinestyle{mystyle}{
    backgroundcolor=\color{backcolour},   
    commentstyle=\color{codegreen},
    keywordstyle=\color{magenta},
    numberstyle=\tiny\color{codegray},
    stringstyle=\color{codepurple},
    basicstyle=\ttfamily\footnotesize,
    breakatwhitespace=false,         
    breaklines=true,                 
    captionpos=b,                    
    keepspaces=true,                 
    numbers=left,                    
    numbersep=5pt,                  
    showspaces=false,                
    showstringspaces=false,
    showtabs=false,                  
    tabsize=2
}

\usepackage{url}

\hypersetup{
    colorlinks=true,
    linkcolor=black,
    citecolor=black,
    filecolor=cyan,
    urlcolor=black
}



\usepackage{listings}
\usepackage{xcolor}

\definecolor{codegreen}{rgb}{0,0.6,0}

\usepackage{tabularx}

\usepackage[accepted]{icml2026}

\icmltitlerunning{CaP-X}

\begin{document}

\twocolumn[
  \icmltitle{CaP-X: A Framework for Benchmarking and\\ Improving Coding Agents for Robot Manipulation}

  \icmlsetsymbol{equal}{*}
  \icmlsetsymbol{lead}{$\dagger$}

  \begin{icmlauthorlist}
    \icmlauthor{Letian Fu}{equal,nvidia,berkeley}
    \icmlauthor{Justin Yu}{equal,berkeley}
    \icmlauthor{Karim El-Refai}{equal,berkeley}
    \icmlauthor{Ethan Kou}{equal,berkeley}
    \icmlauthor{Haoru Xue}{equal,nvidia,berkeley}
    \icmlauthor{Huang Huang}{stanford}
    \icmlauthor{Wenli Xiao}{cmu}
    \icmlauthor{Guanzhi Wang}{nvidia}
    \icmlauthor{Dantong Niu}{nvidia,berkeley}
    \icmlauthor{Fei-Fei Li}{stanford}
    \icmlauthor{Guanya Shi}{cmu}
    \icmlauthor{Jiajun Wu}{stanford}
    \icmlauthor{Shankar Sastry}{berkeley}
    \icmlauthor{Yuke Zhu}{nvidia}
    \\
    \icmlauthor{Ken Goldberg}{lead,berkeley}
    \icmlauthor{Linxi ``Jim'' Fan}{lead,nvidia}
  \end{icmlauthorlist}

  \icmlaffiliation{nvidia}{NVIDIA}
  \icmlaffiliation{berkeley}{UC Berkeley}
  \icmlaffiliation{stanford}{Stanford University}
  \icmlaffiliation{cmu}{Carnegie Mellon University}

  \icmlcorrespondingauthor{Letian Fu}{max.fu.letian@berkeley.edu}

  \icmlkeywords{Machine Learning, ICML}

  \vskip 0.3in
]

\printAffiliationsAndNotice{\icmlEqualContribution\ \ $\dagger$Equal advising}

\newcommand{\cmark}{\checkmark}

\newcolumntype{Y}{>{\centering\arraybackslash}X}

\setlength{\aboverulesep}{0pt}
\setlength{\belowrulesep}{0pt}
\setlength{\extrarowheight}{.75ex}

\def\capbenchleveltable#1{%
\begin{table*}[#1]
\centering
\caption{\textbf{CaP-Bench evaluation tiers.} Each column specifies a tier by how the agent accesses environment state, primitive abstraction, in-context primitive usage examples, and visual-grounding modality. S1--S4 are single-turn; M1--M4 allows multi-turn interaction.}
\label{tab:capbench-levels}
\small
\begin{tabularx}{\textwidth}{llYYYY|YYYY}
\toprule
& & \multicolumn{4}{c|}{\textbf{Single-Turn}} & \multicolumn{4}{c}{\textbf{Multi-Turn}} \\
\cmidrule(r){3-6} \cmidrule(l){7-10}
\textbf{Category} & \textbf{Characteristic} & \textbf{S1} & \textbf{S2} & \textbf{S3} & \textbf{S4} & \textbf{M1} & \textbf{M2} & \textbf{M3} & \textbf{M4} \\
\midrule
\multirow{2}{*}{Perception} & Noiseless (State-Based)& \cmark & & & & & & & \\
\arrayrulecolor{black!10}\cmidrule{2-10}\arrayrulecolor{black}
& Noisy & & \cmark & \cmark & \cmark & \cmark & \cmark & \cmark & \cmark\\
\arrayrulecolor{black!10}\cmidrule{2-10}\arrayrulecolor{black}
\midrule
\multirow{2}{*}{Primitive Abstraction} & High-level & \cmark & \cmark & & & \cmark & \cmark & \cmark & \\
\arrayrulecolor{black!10}\cmidrule{2-10}\arrayrulecolor{black}
& Low-level & & & \cmark & \cmark & & & & \cmark\\
\midrule
In-Context Learning & Primitive Usage Examples & & & \cmark & & & & & \cmark\\
\midrule
\multirow{2}{*}{Visual-Grounding Modality} & Multimodal Feedback & & & & & & \cmark & & \\
\arrayrulecolor{black!10}\cmidrule{2-10}\arrayrulecolor{black}
& Visual Diff. Module (VDM) & & & & & & & \cmark & \cmark\\
\bottomrule
\end{tabularx}
\end{table*}
}

\def\benchmarkcomparison#1{
\begin{table*}[#1]
\centering
\begin{tabular}{l c c c c c}
\toprule
\textbf{Gym/Benchmark} & \textbf{Code Env. Design} & \textbf{Embodied} & \textbf{Multi-Modal} & \textbf{RL Rewards} & \textbf{Multi-API} \\
\hline
LIBERO         & $\boldsymbol{\times}$ & $\checkmark$ & $\checkmark$ & $\checkmark$ & $\boldsymbol{\times}$ \\
Robosuite      & $\boldsymbol{\times}$ & $\checkmark$ & $\checkmark$ & $\checkmark$ & $\boldsymbol{\times}$ \\
BEHAVIOR-1K    & $\boldsymbol{\times}$ & $\checkmark$ & $\checkmark$ & $\checkmark$ & $\boldsymbol{\times}$ \\
VLABench       & $\boldsymbol{\times}$ & $\checkmark$ & $\checkmark$ & $\checkmark$ & $\boldsymbol{\times}$ \\
SWE-bench      & $\checkmark$ & $\boldsymbol{\times}$ & $\boldsymbol{\times}$ & $\boldsymbol{\times}$ & $\boldsymbol{\checkmark}$ \\
SWE-Gym      & $\checkmark$ & $\boldsymbol{\times}$ & $\boldsymbol{\times}$ & $\boldsymbol{\checkmark}$ & $\boldsymbol{\checkmark}$ \\
HumanEval      & $\checkmark$ & $\boldsymbol{\times}$ & $\boldsymbol{\times}$ & $\boldsymbol{\times}$ & $\boldsymbol{\times}$ \\
MMLU           & $\boldsymbol{\times}$ & $\boldsymbol{\times}$ & $\checkmark$ & $\boldsymbol{\times}$ & $\boldsymbol{\times}$ \\
OVMM (HomeRobot) & $\boldsymbol{\times}$ & $\checkmark$ & $\checkmark$ & $\checkmark$ & $\boldsymbol{\times}$ \\
EmbodiedBench  & $\boldsymbol{\times}$ & $\checkmark$ & $\checkmark$ & $\boldsymbol{\times}$ & $\boldsymbol{\times}$ \\
\hline
\textbf{CaP-Bench (ours)} & $\checkmark$ & $\checkmark$ & $\checkmark$ & $\checkmark$ & $\checkmark$ \\
\bottomrule
\end{tabular}
\caption{Comparison of coding and robotics benchmarks \maxfu{remove Multi-API as it is unclear? @ethan thinks it is an important differentiator. maybe we can reword it. from Raven: remove this table}}
\end{table*}
}

\def\liberoproGOALcomparison#1{%
\begin{table}[h!]
    \centering
    \footnotesize %
    \caption{Task-wise LIBERO-PRO performance of OpenVLA, $\pi_0$, $\pi_{0.5}$ and \sysname{} on the \textbf{libero-goal} benchmark. \textbf{Action notation:} $Open(x, y)$ = open target $y$ of container $x$; $Put(obj, loc)$ = place object $obj$ onto/into location $loc$; $Push(obj, loc)$ = push object $obj$ toward location $loc$; $TurnOn(obj)$ = activate object $obj$.}
    \label{tab:liberoproGOALcomparison}
    \begin{tabular}{>{\scriptsize}lcccccccc}
        \toprule
        \multicolumn{1}{l}{\textbf{Task (Symbolic Form)}} & \multicolumn{2}{c}{OpenVLA} & \multicolumn{2}{c}{$\pi_{0}$} & \multicolumn{2}{c}{$\pi_{0.5}$} & \multicolumn{2}{c}{\sysname{}} \\ 
        \cmidrule(lr){2-3} \cmidrule(lr){4-5} \cmidrule(lr){6-7} \cmidrule(lr){8-9} & Pos & Task & Pos & Task & Pos & Task & Pos & Task \\ 
        \midrule
        $Open(cabinet, drawer_{mid})$ & 0.00 & 0.00 & 0.00 & 0.00 & 0.00 & 0.04 & 0.00 & 0.00 \\
        $Put(bowl, drawer_{top})$  & 0.00 & 0.00 & 0.00 & 0.00 & 0.94 & 0.02  & 0.04 & 0.00 \\
        $Push(plate, stove_{front})$ & 0.00 & 0.00 & 0.00 & 0.00 & 0.00 & 0.00  & 0.00 & 0.10 \\
        $Put(bowl, plate)$ & 0.00 & 0.00 & 0.00 & 0.00 & 0.00 & 0.02 & 0.36 & 0.38 \\
        $Put(bowl, stove)$ & 0.00 & 0.00 & 0.00 & 0.00 & 0.00 & 0.04 & 0.22 & 0.12 \\
        $Put(bowl, cabinet_{top})$ & 0.00 & 0.00 & 0.00 & 0.00 & 0.00 & 0.02 & 0.60 & 0.04 \\
        $Put(cream\_cheese, bowl)$ & 0.00 & 0.00 & 0.00 & 0.00 & 0.98 & 0.02  & 0.04 & 0.34 \\
        $Put(wine\_bottle, rack)$  & 0.00 & 0.00 & 0.00 & 0.00 & 0.88 & 0.02  & 0.02 & 0.12 \\
        $Put(wine\_bottle, cabinet_{top})$ & 0.00 & 0.00 & 0.00 & 0.00 & 0.98  & 0.02 & 0.62 & 0.40 \\
        $TurnOn(stove)$ & 0.00 & 0.00 & 0.00 & 0.00 & 0.00 & 0.00  & 0.66 & 0.18 \\
        \midrule
        \rowcolor[gray]{0.92} \textbf{Average} & 0.00 & 0.00 & 0.00 & 0.00  & \textbf{0.38} & 0.00 & 0.256 & \textbf{0.168} \\
        \bottomrule
    \end{tabular}
    \end{table}
}

\def\liberoproSPATIALcomparison#1{%
\begin{table}[h!]
\centering
\footnotesize %
\caption{Task-wise LIBERO-PRO performance of OpenVLA, $\pi_0$, $\pi_{0.5}$ and \sysname{} on the \textbf{libero-spatial} benchmark. \textbf{Action notation:} $Pick(src, dst)$ = pick up object $bowl_{black}$ from location $src$ and
place $bowl_{black}$ onto/into target $dst$.}
\label{tab:liberoproSPATIALcomparison}
\begin{tabular}{>{\scriptsize}lcccccccc} 
\toprule
\multicolumn{1}{l}{\textbf{Task (Symbolic Form)}} & \multicolumn{2}{c}{OpenVLA} & \multicolumn{2}{c}{$\pi_{0}$} & \multicolumn{2}{c}{$\pi_{0.5}$} & \multicolumn{2}{c}{\sysname{}} \\ 
        \cmidrule(lr){2-3} \cmidrule(lr){4-5} \cmidrule(lr){6-7} \cmidrule(lr){8-9}
         & Pos & Task & Pos & Task & Pos & Task & Pos & Task \\ 
        \midrule
$Pick(between(plate, ramekin), plate)$ & 0.00 & 0.00 & 0.00 & 0.00 & 0.02 & 0.00 & 0.22 & 0.14 \\
$Pick(table\_center, plate)$                & 0.00 & 0.00 & 0.00 & 0.00 & 0.00 & 0.02 & 0.22 & 0.14 \\
$Pick(drawer_{top}(cabinet_{wood}), plate)$  & 0.00 & 0.00 & 0.00 & 0.00 & 0.00 & 0.00 & 0.02 & 0.10 \\
$Pick(next\_to(cookie\_box), plate)$         & 0.00 & 0.00 & 0.00 & 0.00 & 0.00 & 0.02 & 0.00 & 0.10 \\
$Pick(next\_to(plate), plate)$               & 0.00 & 0.00 & 0.00 & 0.00 & 0.00 & 0.00 & 0.10 & 0.20 \\
$Pick(next\_to(ramekin), plate)$             & 0.00 & 0.00 & 0.00 & 0.00 & 0.12 & 0.02 & 0.30 & 0.14 \\
$Pick(on(cookie\_box), plate)$               & 0.00 & 0.00 & 0.00 & 0.00 & 0.00 & 0.00 & 0.14 & 0.08 \\
$Pick(on(ramekin), plate)$                   & 0.00 & 0.00 & 0.00 & 0.00 & 0.98 & 0.02 & 0.02 & 0.20 \\
$Pick(on(stove), plate)$                     & 0.00 & 0.00 & 0.00 & 0.00 & 0.02 & 0.00 & 0.08 & 0.14 \\
$Pick(on(cabinet_{wood}), plate)$            & 0.00 & 0.00 & 0.00 & 0.00 & 0.90 & 0.00 & 0.08 & 0.16 \\
\midrule
\rowcolor[gray]{0.92} \textbf{Average} & 0.00 & 0.00 & 0.00 & 0.00 & \textbf{0.20} & 0.01 & 0.118 & \textbf{0.14} \\
\bottomrule
\end{tabular}
\end{table}
}

\def\liberoproOBJECTcomparison#1{
\begin{table}[h!]
    \centering
    \footnotesize 
    \caption{Task-wise LIBERO-PRO performance of OpenVLA, $\pi_0$, $\pi_{0.5}$ and \sysname{} on the \textbf{libero-object} benchmark. \textbf{Action notation:} $Place(obj, loc)$ = pick up object $obj$ and place $obj$ into/onto target $loc$.}
    \label{tab:liberoproOBJECTcomparison}
    \begin{tabular}{>{\scriptsize}lcccccccc} 
    \toprule
    \multicolumn{1}{l}{\textbf{Task (Symbolic Form)}} & \multicolumn{2}{c}{OpenVLA} & \multicolumn{2}{c}{$\pi_{0}$} & \multicolumn{2}{c}{$\pi_{0.5}$} & \multicolumn{2}{c}{\sysname{}} \\ 
    \cmidrule(lr){2-3} \cmidrule(lr){4-5} \cmidrule(lr){6-7} \cmidrule(lr){8-9}
     & Pos & Task & Pos & Task & Pos & Task & Pos & Task \\ 
    \midrule
    $Place(\text{alphabet\_soup}, \text{basket})$ & 0.00 & 0.00 & 0.00 & 0.00 & 0.00 & 0.00 & 0.02 & 0.04 \\
    $Place(\text{bbq\_sauce}, \text{basket})$ & 0.00 & 0.00 & 0.00 & 0.00 & 1.00 & 0.02 & 0.12 & 0.42 \\
    $Place(\text{butter}, \text{basket})$ & 0.00 & 0.00 & 0.00 & 0.00 & 0.54 & 0.00 & 0.26 & 0.18 \\
    $Place(\text{chocolate\_pudding}, \text{basket})$ & 0.00 & 0.00 & 0.00 & 0.00 & 0.00 & 0.02 & 0.18 & 0.48 \\
    $Place(\text{cream\_cheese}, \text{basket})$ & 0.00 & 0.00 & 0.10 & 0.00 & 0.00 & 0.00 & 0.12 & 0.06 \\
    $Place(\text{ketchup}, \text{basket})$ & 0.00 & 0.00 & 0.00 & 0.00 & 0.20 & 0.02 & 0.32 & 0.12 \\
    $Place(\text{milk}, \text{basket})$ & 0.00 & 0.00 & 0.00 & 0.00 & 0.00 & 0.00 & 0.38 & 0.02 \\
    $Place(\text{orange\_juice}, \text{basket})$ & 0.00 & 0.00 & 0.00 & 0.00 & 0.00 & 0.02 & 0.30 & 0.02 \\
    $Place(\text{salad\_dressing}, \text{basket})$ & 0.00 & 0.00 & 0.10 & 0.00 & 0.00 & 0.00 & 0.32 & 0.00 \\
    $Place(\text{tomato\_sauce}, \text{basket})$ & 0.00 & 0.00 & 0.00 & 0.00 & 0.00 & 0.00 & 0.16 & 0.48 \\
    \midrule
    \rowcolor[gray]{0.92} \textbf{Average} & 0.00 & 0.00 & 0.00 & 0.00 & 0.17 & 0.01 & \textbf{0.218} & \textbf{0.182} \\
    \bottomrule
    \end{tabular}
    \end{table}
}

\def\liberoproAVGcomparison#1{
    \begin{table}[#1]
    \centering
    \tiny
    \caption{LIBERO-PRO~\cite{zhou2025liberopro} performance of OpenVLA~\cite{kim2024openvlaopensourcevisionlanguageactionmodel}, $\pi_0$~\cite{black2024pi0visionlanguageactionflowmodel}, $\pi_{0.5}$~\cite{intelligence2025pi05} and \sysname{} on the \textbf{libero-object}, \textbf{libero-goal}, and \textbf{libero-spatial} benchmarks under initial position perturbations (Pos) and instruction perturbations (Task) averaged across tasks. Detailed individual task-wise performance can be found in \cref{app:liberoproresults}.}
    \label{tab:liberoproAVG}
    \setlength{\tabcolsep}{3pt}
    \begin{tabular}{>{\scriptsize}lcccccc} 
    \toprule
    \multicolumn{1}{l}{\textbf{Method}} & \multicolumn{2}{c}{\textbf{libero-object}} & \multicolumn{2}{c}{\textbf{libero-goal}} & \multicolumn{2}{c}{\textbf{libero-spatial}} \\ 
            \cmidrule(lr){2-3} \cmidrule(lr){4-5} \cmidrule(lr){6-7}
             & Pos (Avg.) & Task (Avg.) & Pos (Avg.) & Task (Avg.) & Pos (Avg.) & Task (Avg.)\\ 
            \midrule
    OpenVLA & 0.00 & 0.00 & 0.00 & 0.00 & 0.00 & 0.00\\
    $\pi_0$ & 0.00 & 0.00 & 0.00 & 0.00 & 0.00 & 0.00\\
    $\pi_{0.5}$ & 0.17 & 0.01 & \textbf{0.38} & 0.00 & \textbf{0.20} & 0.01\\
    \sysname{} & \textbf{0.22} & \textbf{0.18} & 0.26  & \textbf{0.17} & 0.12 & \textbf{0.14}\\
    \bottomrule
    \end{tabular}%
    \end{table}
}

\def\simbktask#1{
    \begin{table}[#1]
    \centering
    \caption{Results on BEHAVIOR~\cite{li2024behavior1k} Tasks}
    \label{tab:b1k}
    \resizebox{\columnwidth}{!}{%
        \begin{tabular}{lcccccc}
        \toprule
        \textbf{Task} & \multicolumn{3}{c}{\textbf{Nav. Success Rate}} & \multicolumn{3}{c}{\textbf{Task. Success Rate}} \\
        \cmidrule(lr){2-4} \cmidrule(lr){5-7}
        &Human&  S3  &\sysname{} &Human  & S3  &\sysname{} \\
        \midrule
        Pick up Radio & \textbf{88\%} &72\%& 80\% & 36\% & 24\% & \textbf{56\%}   \\
        Pick up Soda Can & 80\% & 52\% & \textbf{84\%} & \textbf{72\%} & 32\% & \textbf{72\%} \\
        \bottomrule
        \end{tabular}%
    }
    \end{table}
}

\def\caprltab#1{
    \begin{table}[#1]
    \centering
    \caption{\textbf{Impact of RL Post-Training in Sim and Real.} Comparison of success rates between the base model, our RL post-trained agent, and human experts. Simulation results are averaged over 100 trials per task, while real-world deployment on a Franka Emika robot is evaluated over 25 trials.}
    \label{tab:rl_combined}
    \resizebox{\columnwidth}{!}{%
    \begin{tabular}{lccccc}
    \toprule
     & \multicolumn{3}{c}{\textbf{Simulation (N=100)}} & \multicolumn{2}{c}{\textbf{Real World (N=25)}} \\
    \cmidrule(lr){2-4} \cmidrule(lr){5-6}
    \textbf{Method} & \textbf{Cube Lift} & \textbf{Cube Stack} & \textbf{Spill Wipe} & \textbf{Cube Lift} & \textbf{Cube Stack} \\
    \midrule
    \rowcolor[gray]{0.92} Human Expert       & 93\% & 73\% & 100\% & 92\% & 84\% \\
    Qwen 2.5 Coder 7B  & 25\% & 4\%  & 30\%  & 24\% & 12\% \\
    Qwen w/ CaP-RL          & \textbf{80\%} & \textbf{44\%} & \textbf{93\%}  & \textbf{84\%} & \textbf{76\%} \\
    \bottomrule
    \end{tabular}
    }
    \end{table}
}

\def\splashfigure#1{
    \begin{figure*}[#1]
        \centering
        \includegraphics[width=\textwidth]{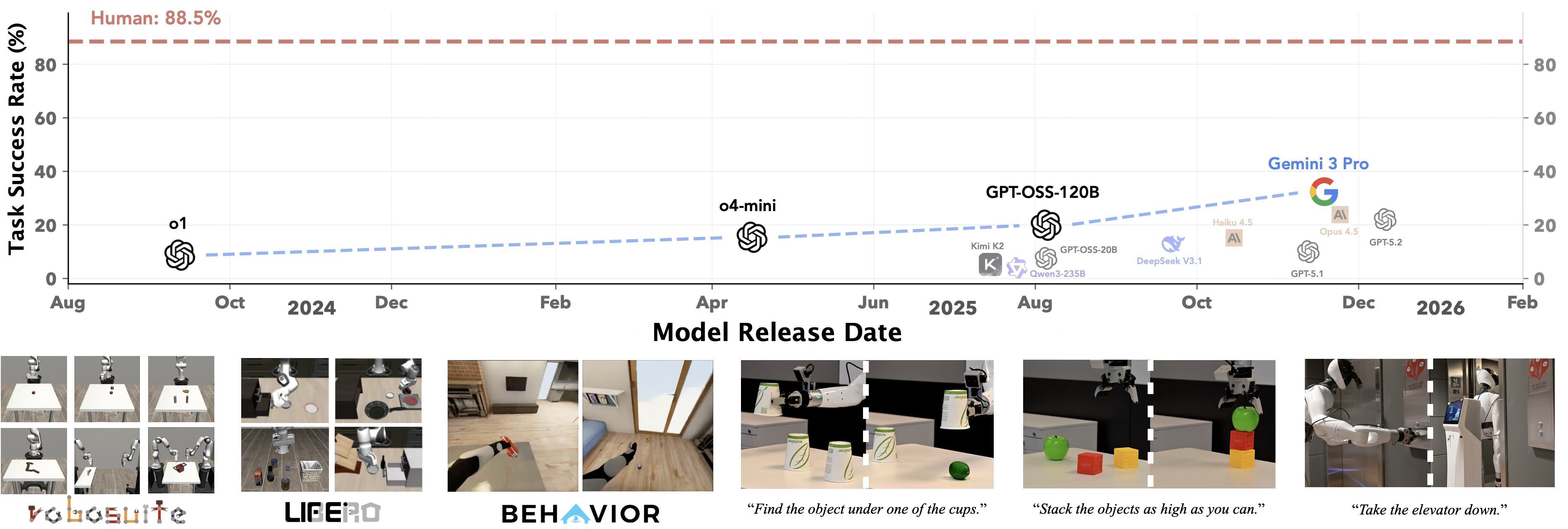}
        \caption{
        \textbf{(Top)} CaP-Bench Task Success Rate over Model Release Date: across 7 tasks and 12 models, we compare the success rates of model-generated robot-control programs against those of human expert-written programs. We find that while (vision-) language models have achieved capabilities comparable to humans in other domains~\cite{jimenez2024swebench, rein2024gpqa, hendrycks2020mmlu}, they still trail behind human performance in writing code that controls robots for manipulation tasks. \textbf{(Bottom)}: \gymname{} integrates Robosuite~\cite{zhu2020robosuite}, LIBERO-PRO~\cite{zhou2025liberopro}, and BEHAVIOR~\cite{li2024behavior1k}. We further present \sysname{} (\cref{sec:capagent}), a training-free agentic framework that recovers near human-level performance on several manipulation tasks and achieves success rates comparable to—and in some cases exceeding—those of post-trained VLAs, without any task-specific training data.
        }
        \label{fig:splash_fig}
    \end{figure*}
}

\def\capgymenvfigure#1{
    \begin{figure*}[#1]
        \centering
        \includegraphics[width=\textwidth]{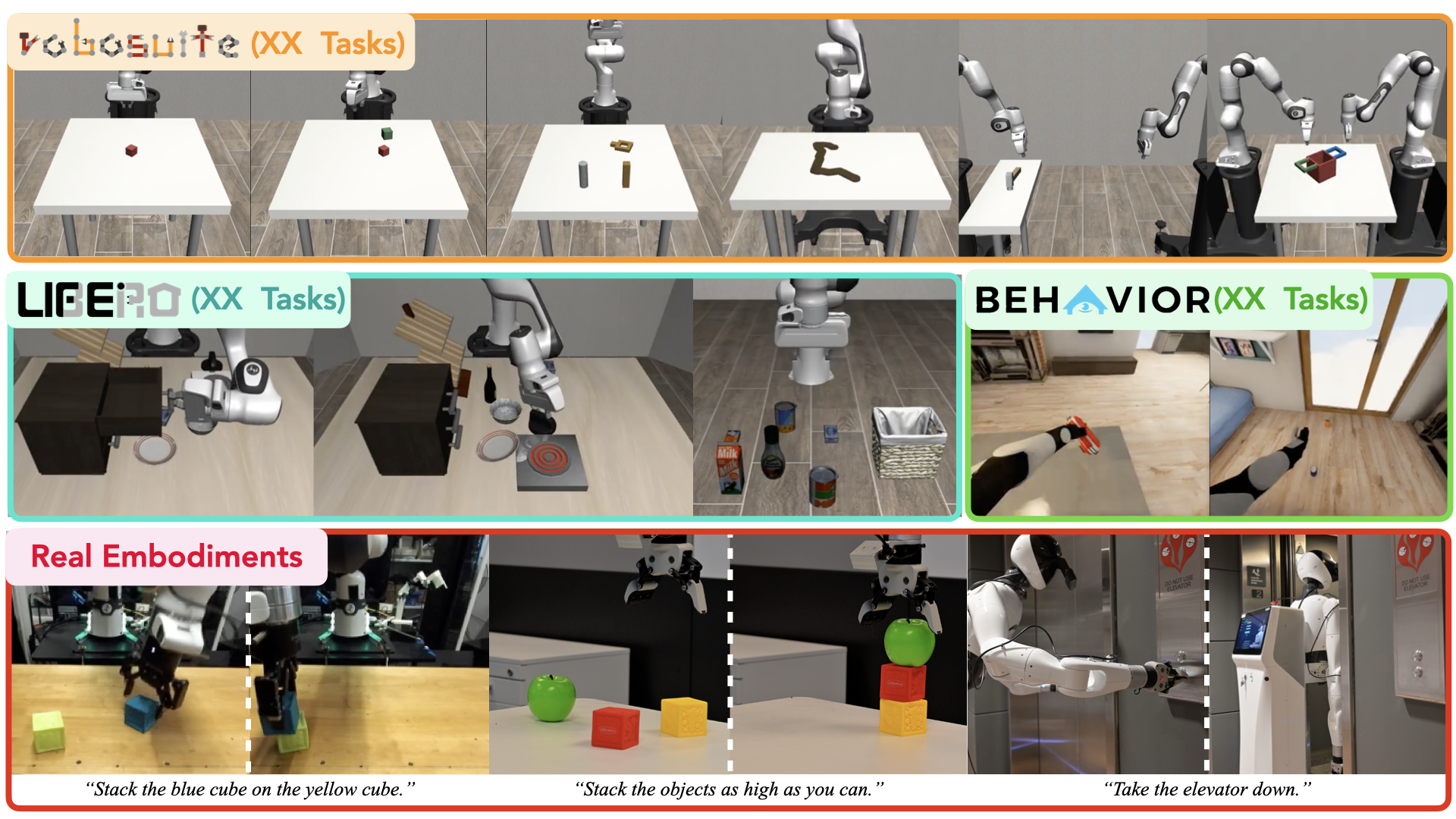}
        \caption{
        \textbf{(Top and Middle Row) \gymname{} environments.}
        \gymname{} integrates XX tasks from RoboSuite~\cite{zhu2020robosuite}, LIBERO-PRO~\cite{zhou2025liberopro}, and BEHAVIOR~\cite{li2024behavior1k}. \textbf{(Bottom Row) CaP-Agent0 in real-world manipulation.} We deploy \sysname{} on a real-world AgiBot G1~\cite{bu2025agibot_iros} robot to perform a wide variety of robot manipulation tasks zero-shot (see Appendix \cref{app:real-world-agibot}).
        }
        \label{fig:capgym_envs}
    \end{figure*}
}

\def\capagentfigure#1{
    \begin{figure*}[#1]
        \centering
        \includegraphics[width=\textwidth]{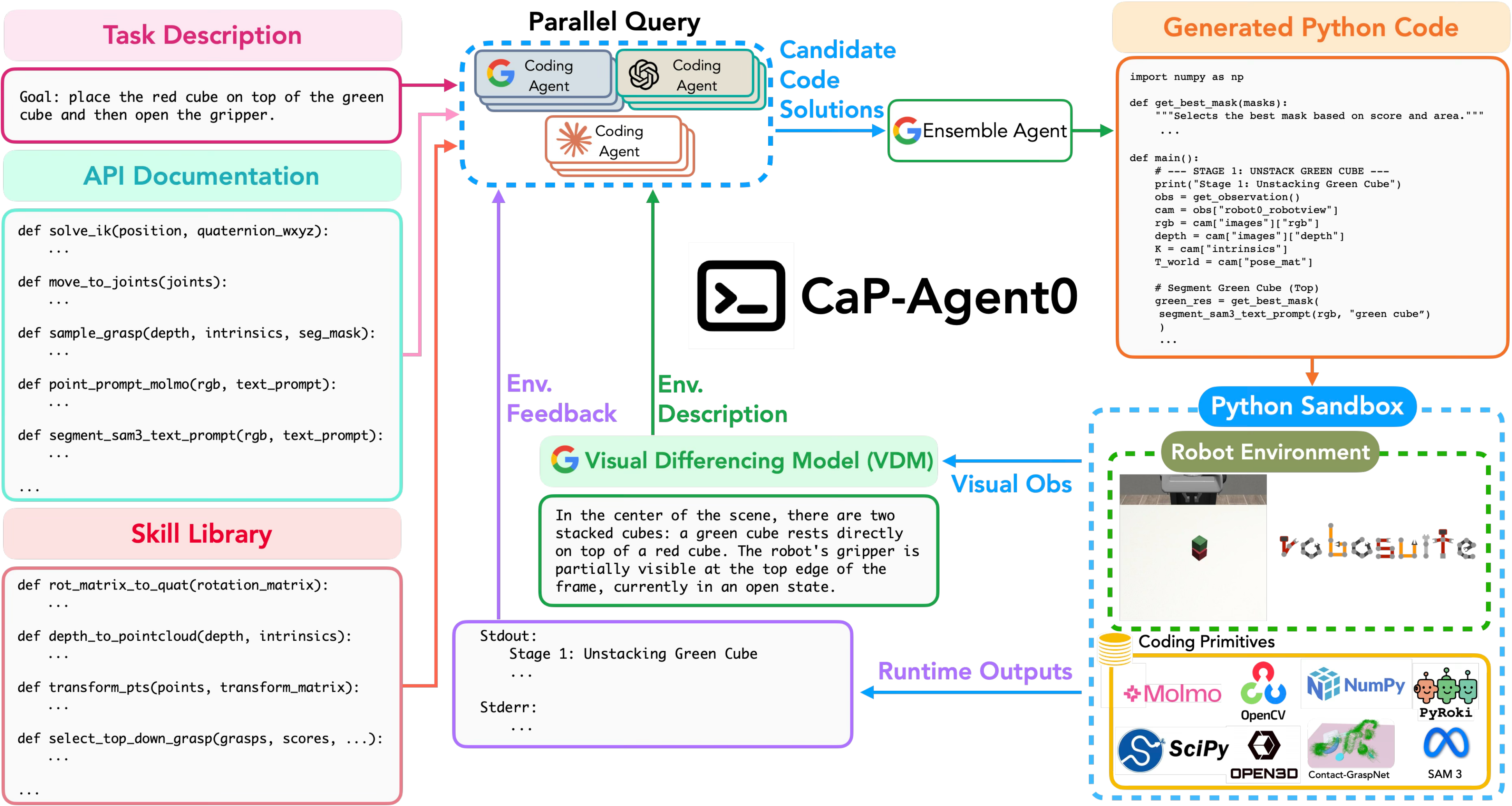}
        \caption{
        \textbf{CaP-Agent0.} CaP-Agent0 incorporates an auto-synthesized skill library of auxiliary utility generated by coding agents during CaP-Bench, a visual differencing model (VDM) which provides a textual description of the initial scene and for each subsequent turn what changes have occurred in the scene since, and a parallel reasoning system where multiple coding agents are provided the same prompt. These coding agents then generate candidate code solutions that may solve the task and the ensemble agent must then synthesize these code generations into a final code snippet which is then executed in the Python sandbox containing the robot environment, whether that environment may be a simulator like Robosuite or on a real robot. For more details see Section \ref{sec:capagent}.
        }
        \label{fig:capagent0_sys}
    \end{figure*}
}

\def\abstractionlevelfigure#1{
    \begin{figure*}[#1]
        \centering
        \includegraphics[width=\textwidth]{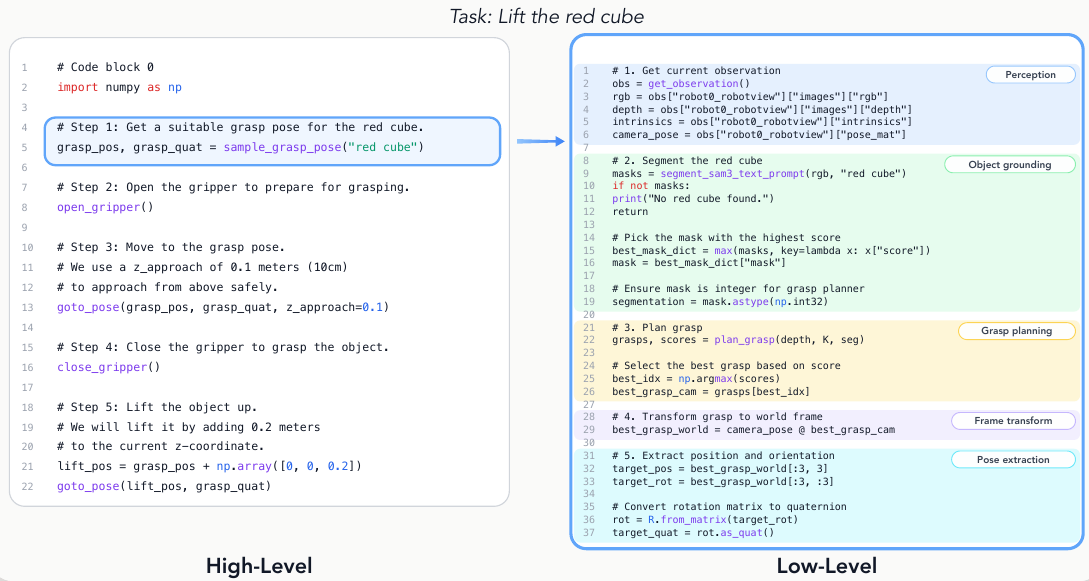}
        \caption{
        \textbf{(Left)} An example of code generated by Gemini-3-Pro for completing the task \textit{``lift the red cube"} using the high-level primitives. \textbf{(Right)} Code generated by Gemini-3-Pro using just low-level primitives to achieve the same functionality as a single high-level primitive. For more details on the differences between high and low-level primitives, see Section \ref{subsec:singleturn} and Appendix \ref{app:apis}.
        }
        \label{fig:abstractionlevel_diff}
    \end{figure*}
}

\def\humanvsmodels#1{
    \begin{figure*}[#1]
        \centering
        \includegraphics[width=\textwidth]{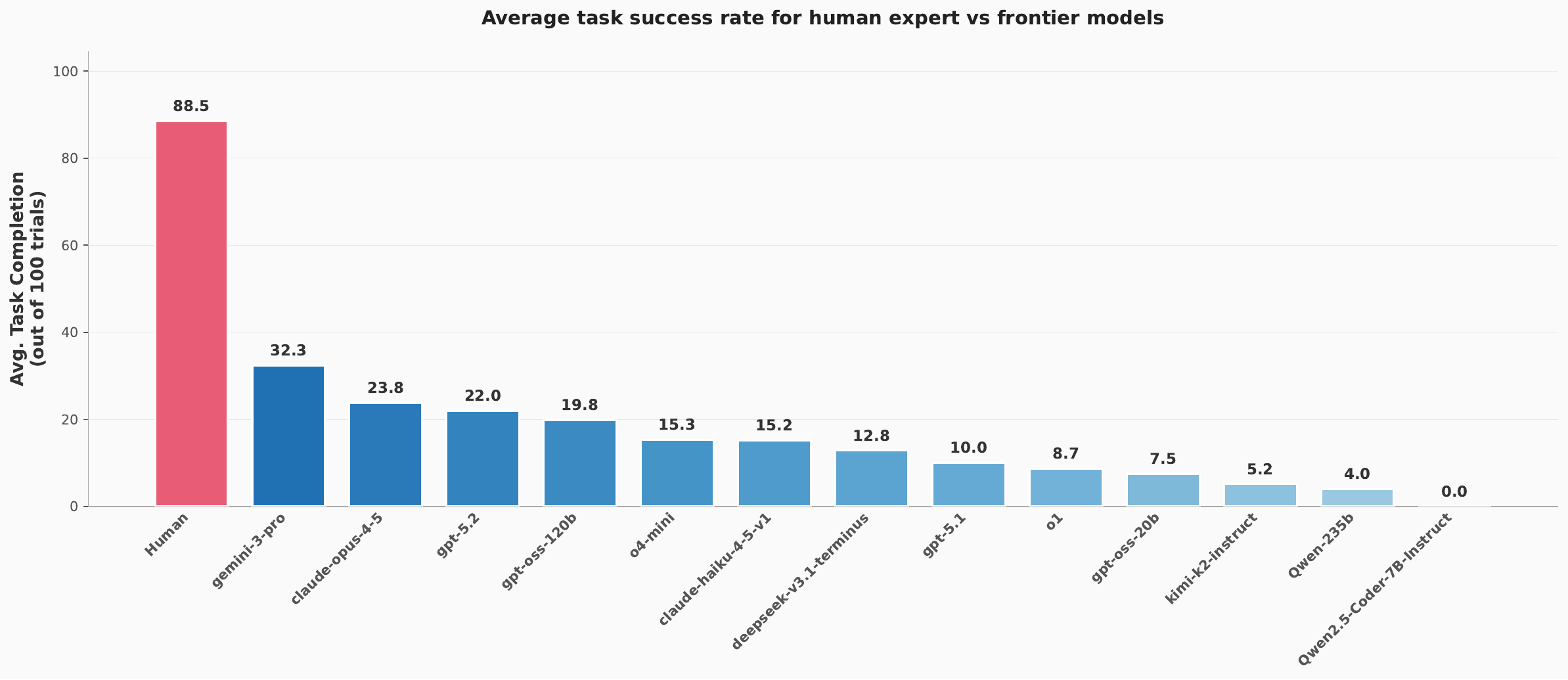}
        \caption{Performance comparison between human experts and frontier models on abstraction tier S4.}
        \label{fig:human_vs_single_turn}
    \end{figure*}
}

\def\abstractionincrease#1{
    \begin{figure}[#1]
        \centering
        \includegraphics[width=0.45\textwidth]{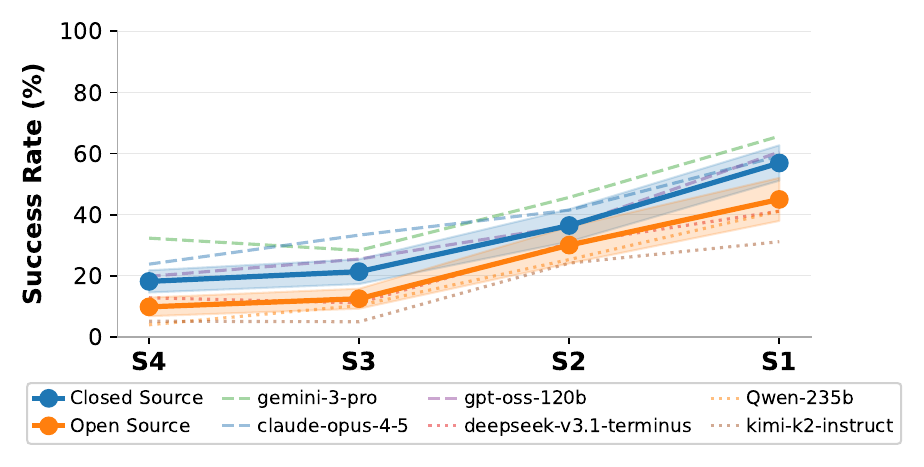}
        \caption{Average task success rate across open-source and closed-source models as primitive abstraction increases. As primitive abstraction increases S4 (per model performance illustrated in \cref{fig:splash_fig}) to S1 (high level primitives with full state observation), the success rate increases. This can only in part be attributed to reduced code correctness at S3-S4, see \cref{fig:codecompilation}.}
        \label{fig:abstraction_increase}
    \end{figure}
}

\def\codecompilation#1{
    \begin{figure}[#1]
        \centering
        \includegraphics[width=0.45\textwidth]{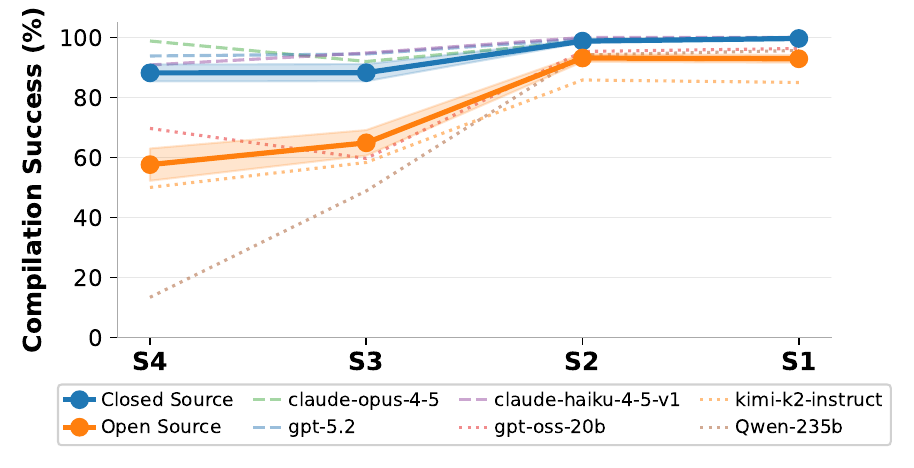}
        \caption{Code Execution Success Rate vs primitive abstraction.}
        \label{fig:codecompilation}
    \end{figure}
}

\def\iclapiusage#1{
    \begin{figure}[#1]
        \centering
        \includegraphics[width=\textwidth]{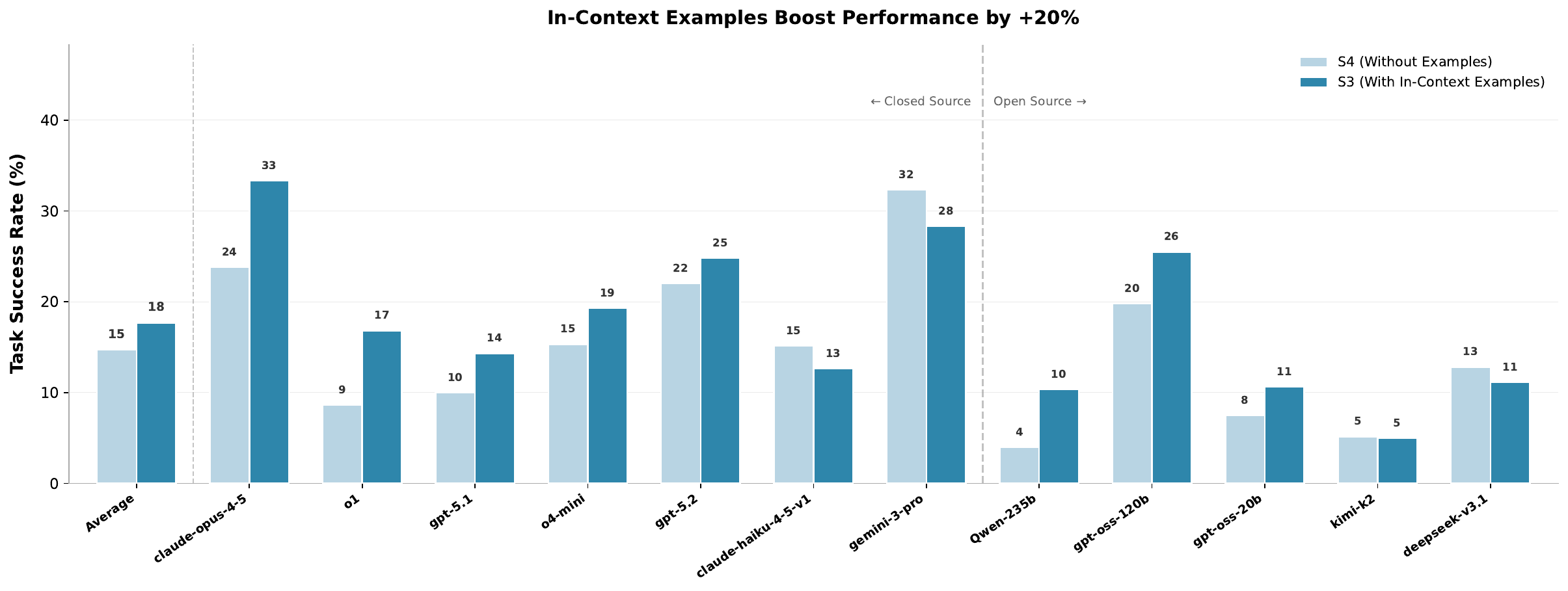}
        \caption{Per-model breakdown of usefulness of in-context API Usage}
        \label{fig:iclapiusage}
    \end{figure}
}

\def\pertaskicl#1{
    \begin{figure}[#1]
        \centering
        \includegraphics[width=\textwidth]{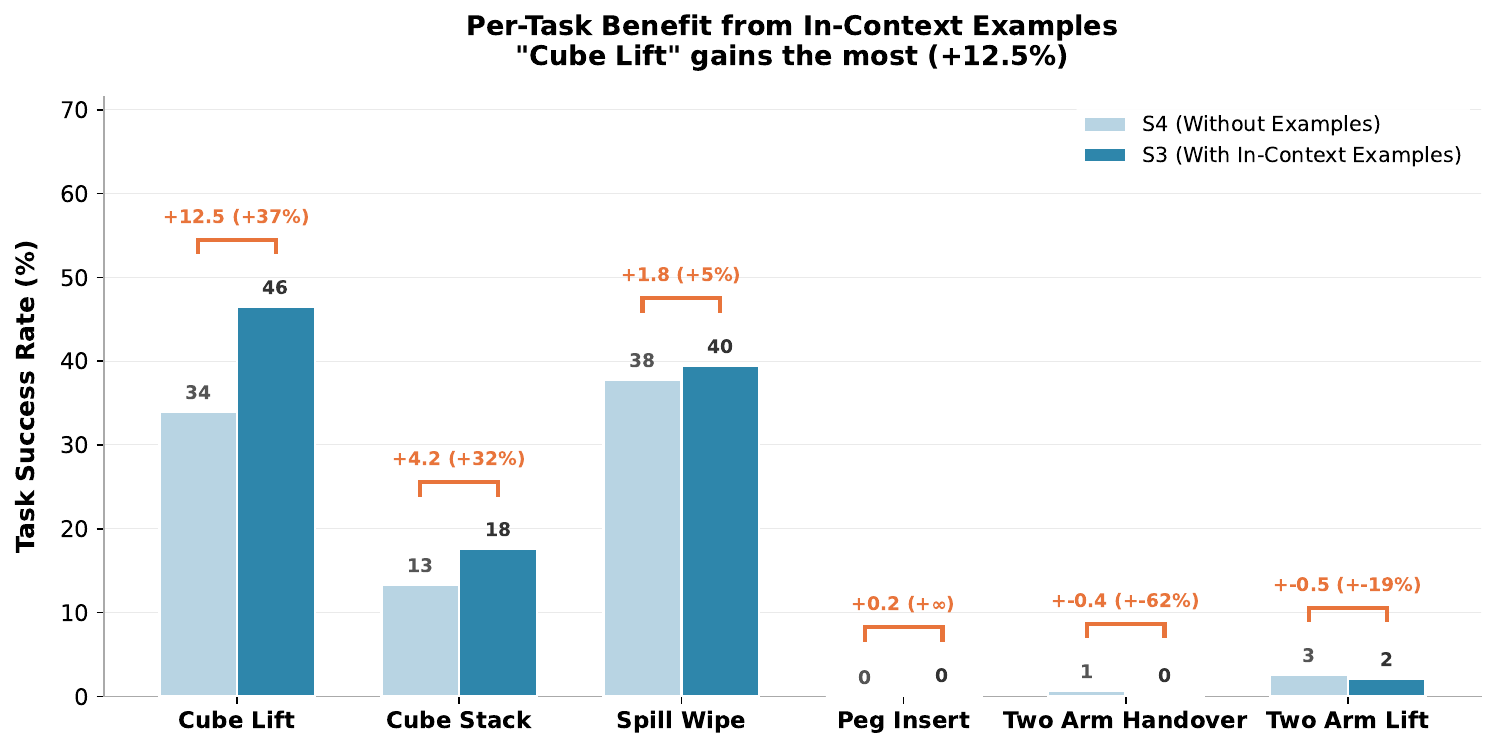}
        \caption{Average model performance improvement from In-Context API Usage Examples. The ``Cube Lift" task across all models sees an improvement of 12.5\% improvement in success rate.}
        \label{fig:pertaskicl}
    \end{figure}
}

\def\multiturnvssingle#1{
    \begin{figure}[#1]
        \centering
        \includegraphics[width=0.45\textwidth]{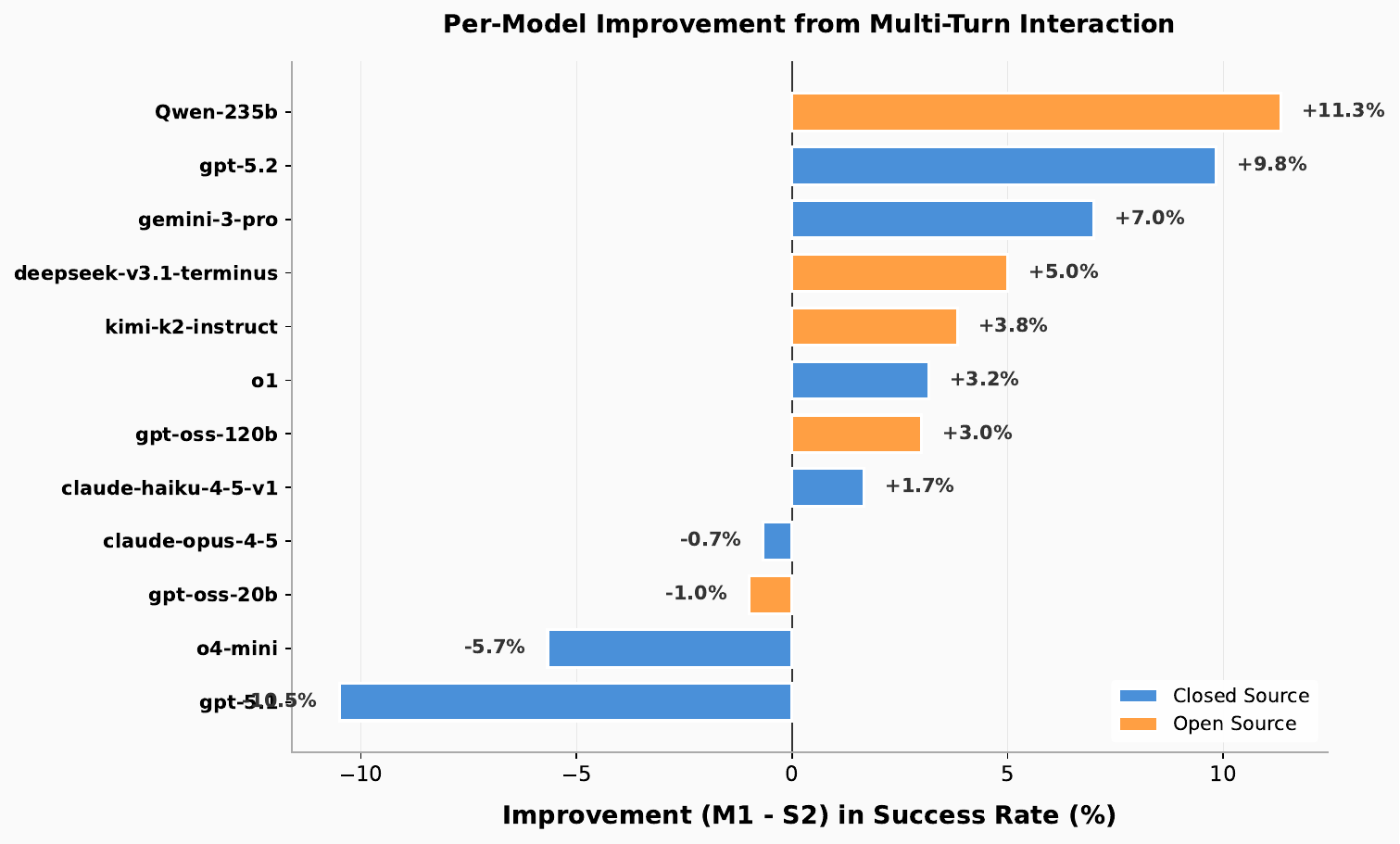}
        \caption{Enabling multi-turn interaction with the environment improves task performance.}
        \label{fig:multiturnvssingle}
    \end{figure}
}

\def\visualfeedbackcomp#1{
    \begin{figure}[#1]
        \centering
        \includegraphics[width=0.45\textwidth]{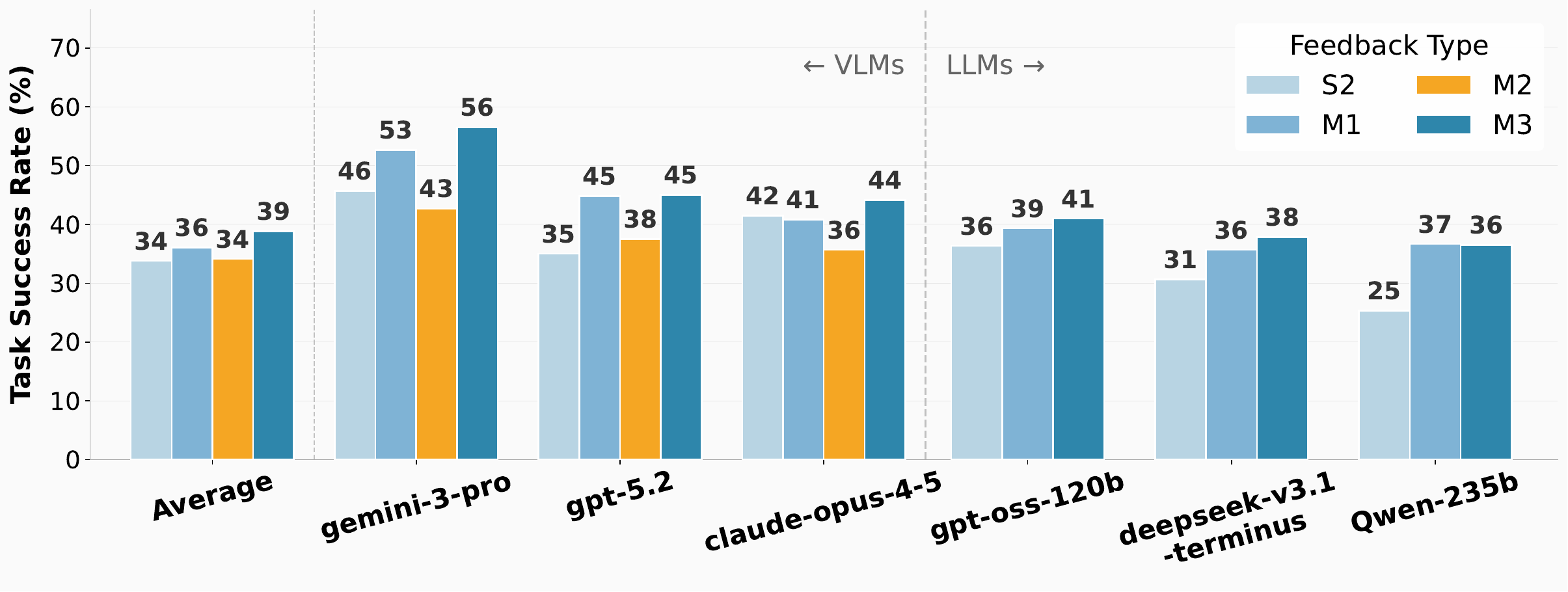}
        \caption{Comparison of single-turn (S2) and multi-turn tiers (M1-M3) across models. Enabling textual feedback through multi-turn (M1) improves task success rate in most models. Direct multimodal (M2) visual grounding reduced task success rate. We find that visual differencing into text (M3) instead of direct multimodal visual input consistently improves task success rate across both close and open source models. }
        \label{fig:visualfeedbackcomp}
    \end{figure}
}

\def\vdmlvslvsnp#1{
    \begin{figure}[#1]
        \centering
        \includegraphics[width=0.45\textwidth]{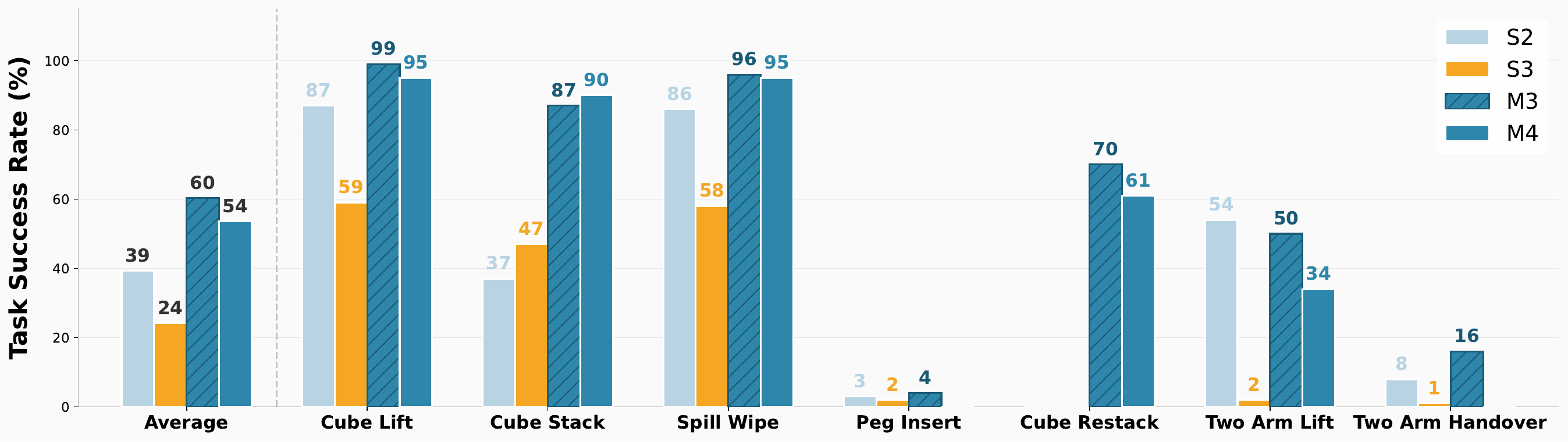}
        \caption{Evaluation of Gemini-3-Pro across four tiers of the benchmark. Multi-turn with visual differencing and low-level API (M4) significantly outperforms both single-turn low-level API (S3), and more notably, single-turn high-level API (S2).}
        \label{fig:vdmlvslvsnp}
    \end{figure}
}

\def\tennisballcasestudy#1{
    \begin{figure*}[#1]
        \centering
        \includegraphics[width=0.19\textwidth]{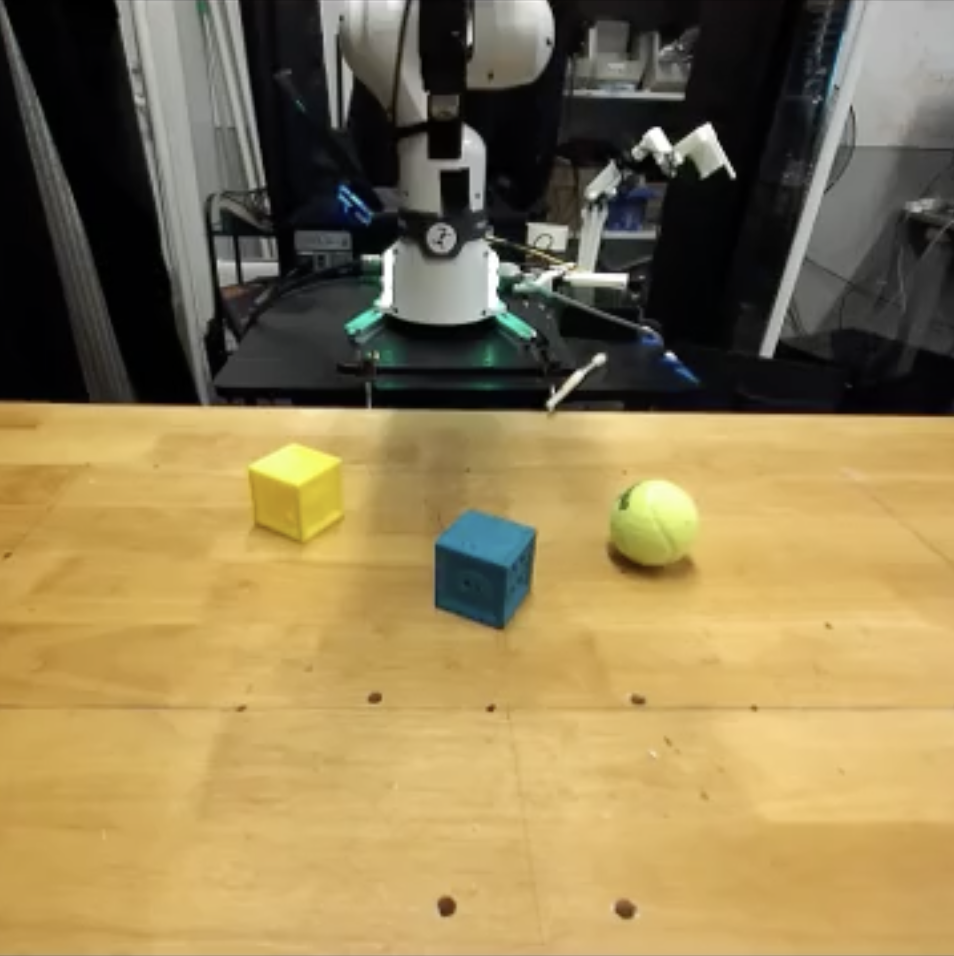}
        \hfill
        \includegraphics[width=0.19\textwidth]{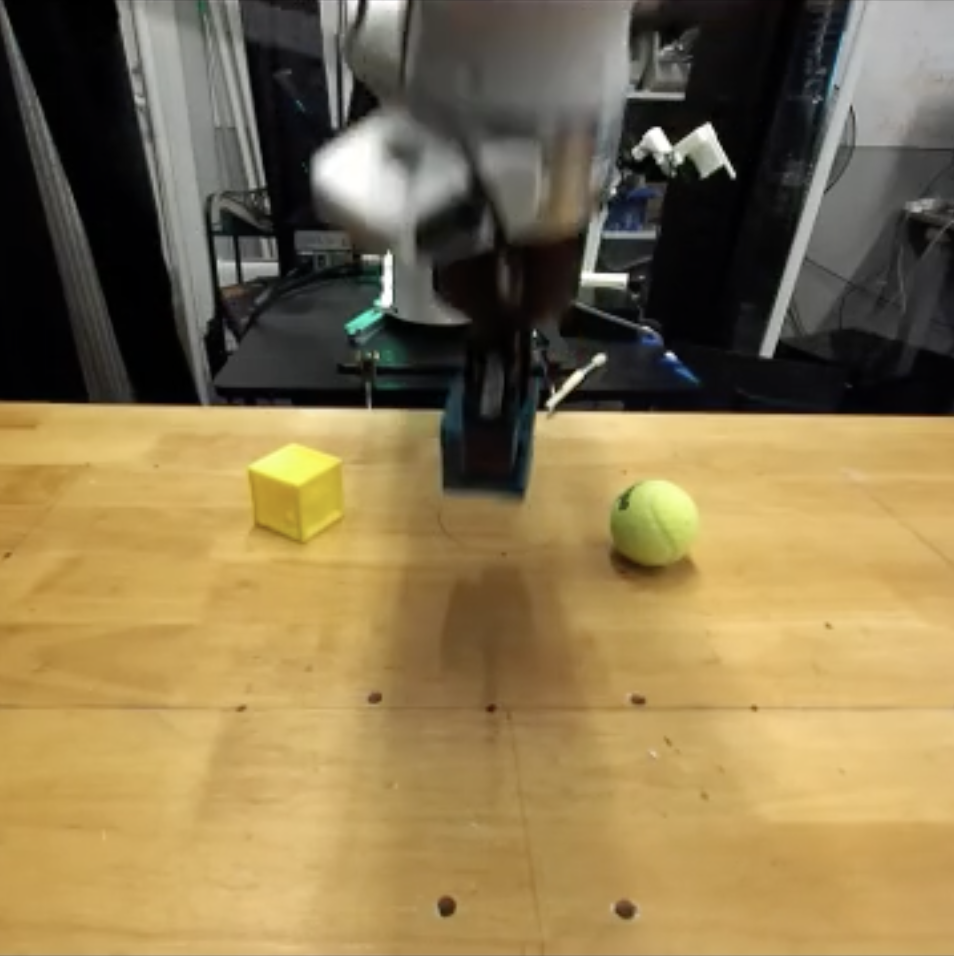}
        \hfill
        \includegraphics[width=0.19\textwidth]{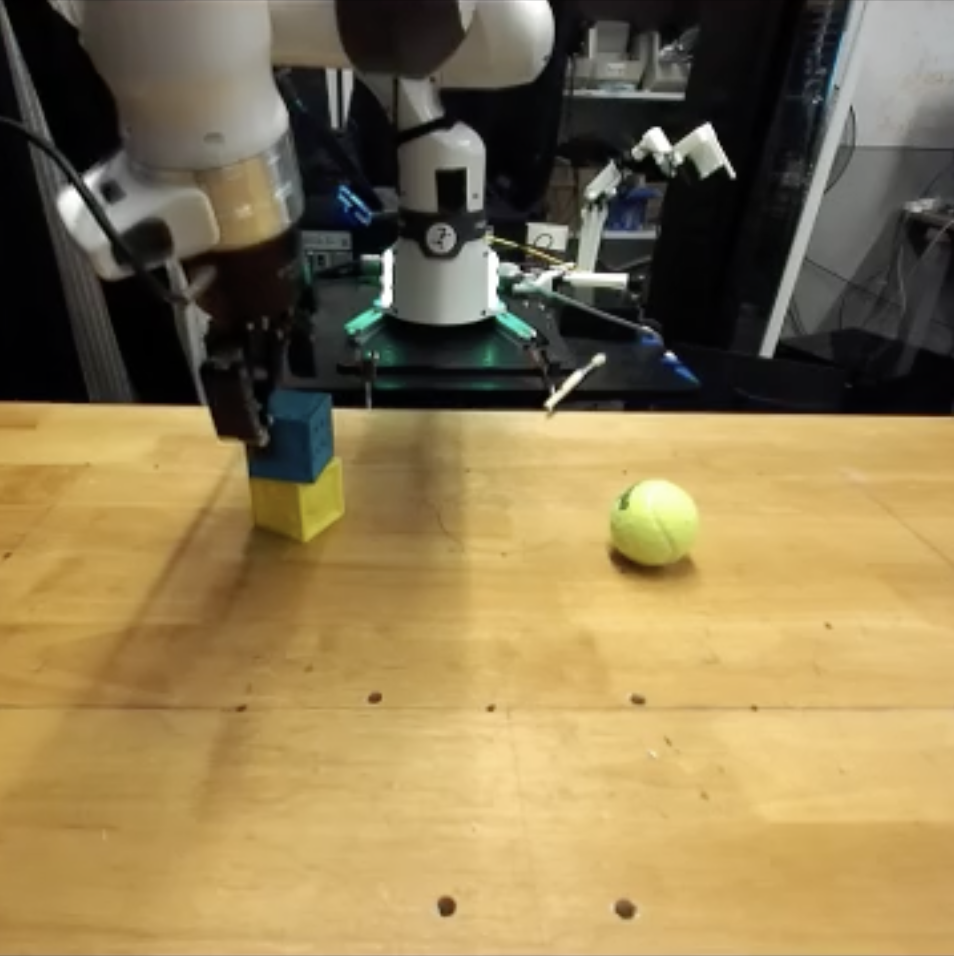}
        \hfill
        \includegraphics[width=0.19\textwidth]{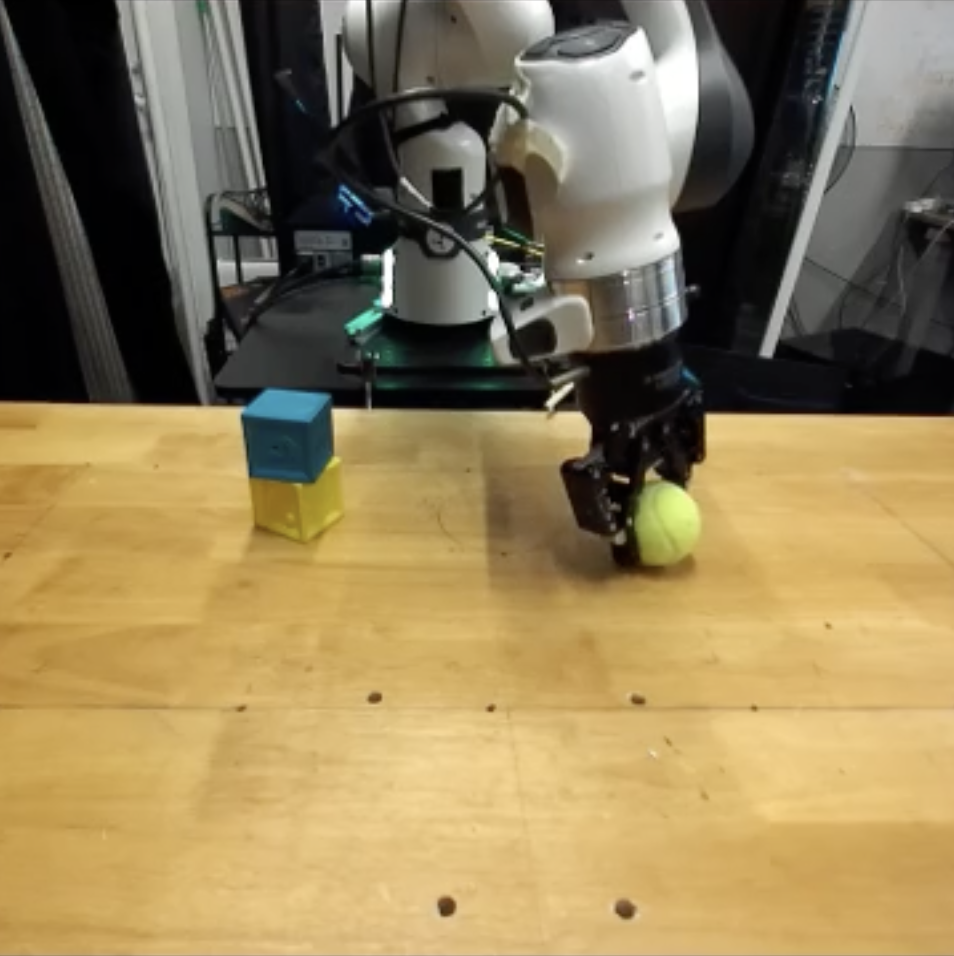}
        \hfill
        \includegraphics[width=0.19\textwidth]{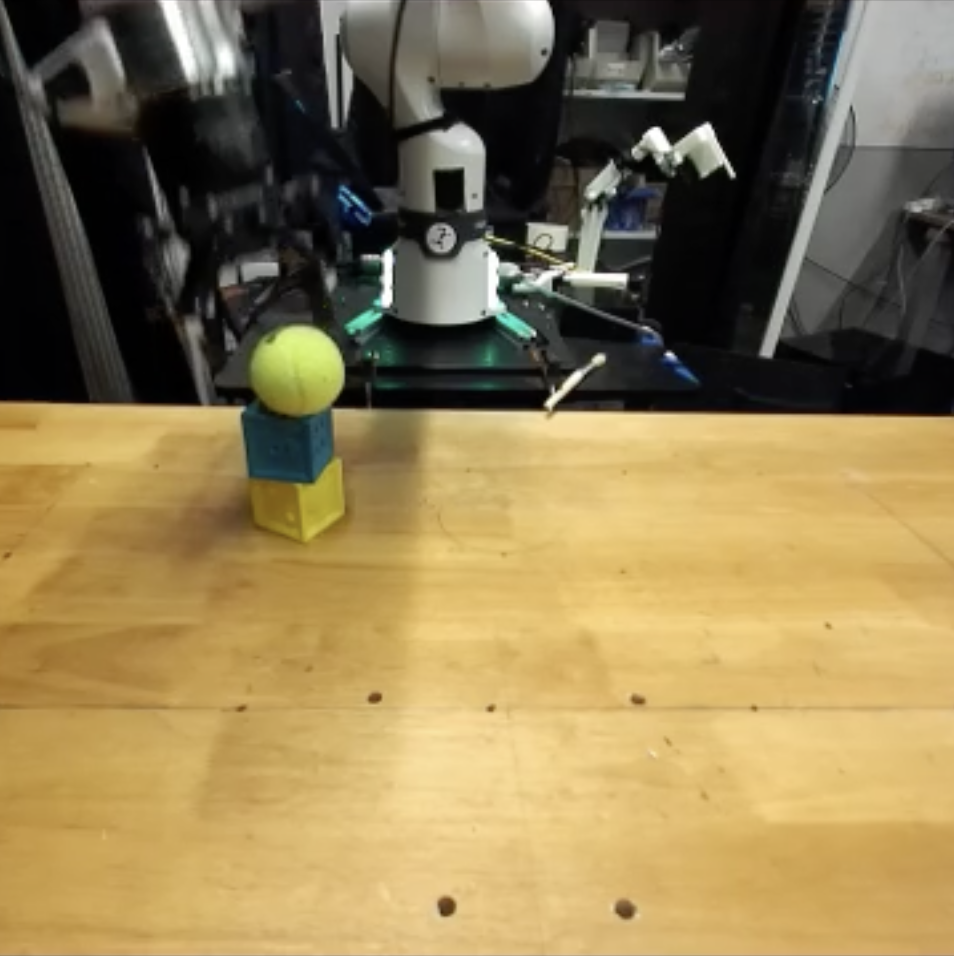}
        
        \caption{"Stack these as high as you can."}
        \label{fig:tennis_ball_real}
    \end{figure*}
}
\def\restackrealcasestudy#1{
    \begin{figure*}[#1]
        \centering
        \includegraphics[width=0.19\textwidth]{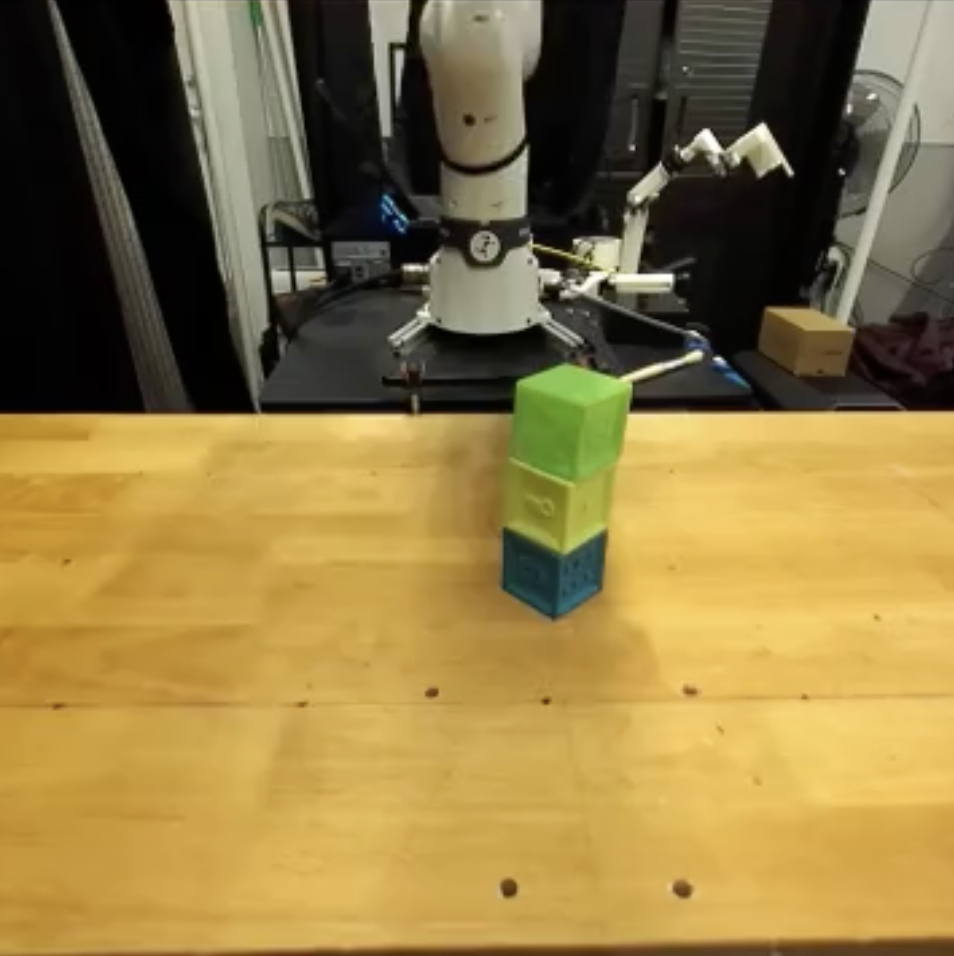}
        \hfill
        \includegraphics[width=0.19\textwidth]{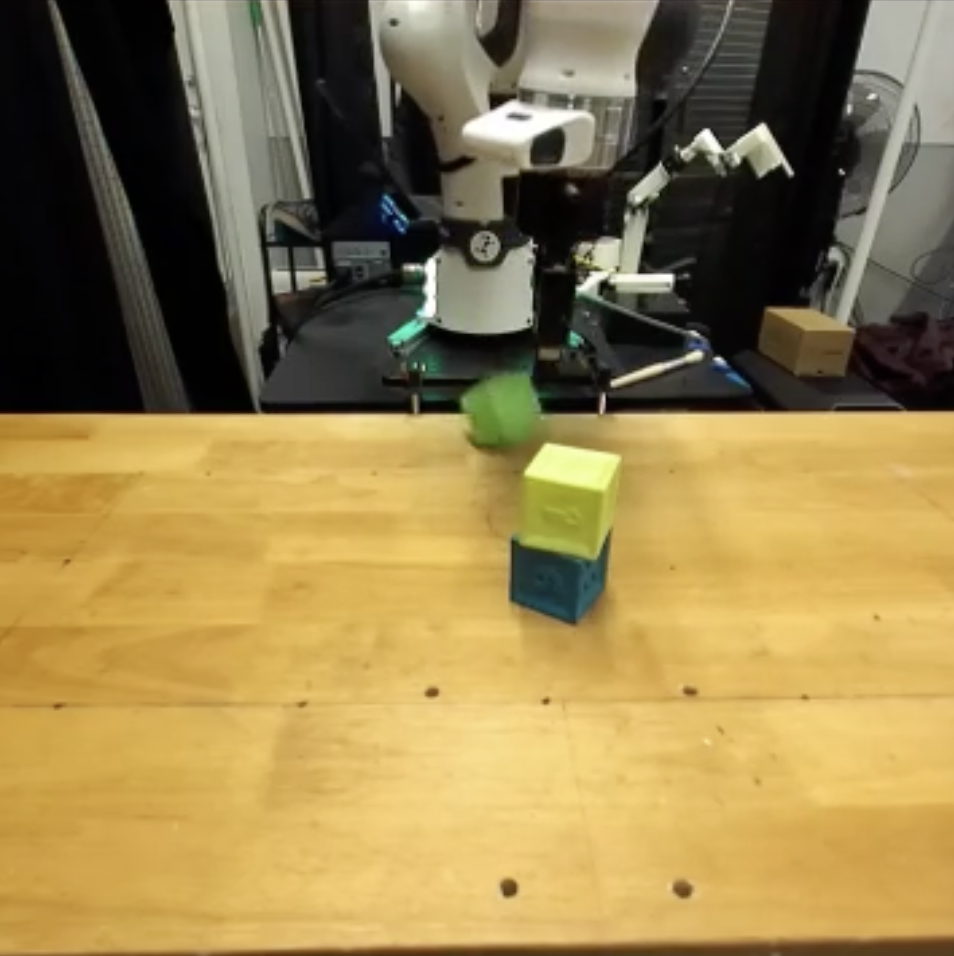}
        \hfill
        \includegraphics[width=0.19\textwidth]{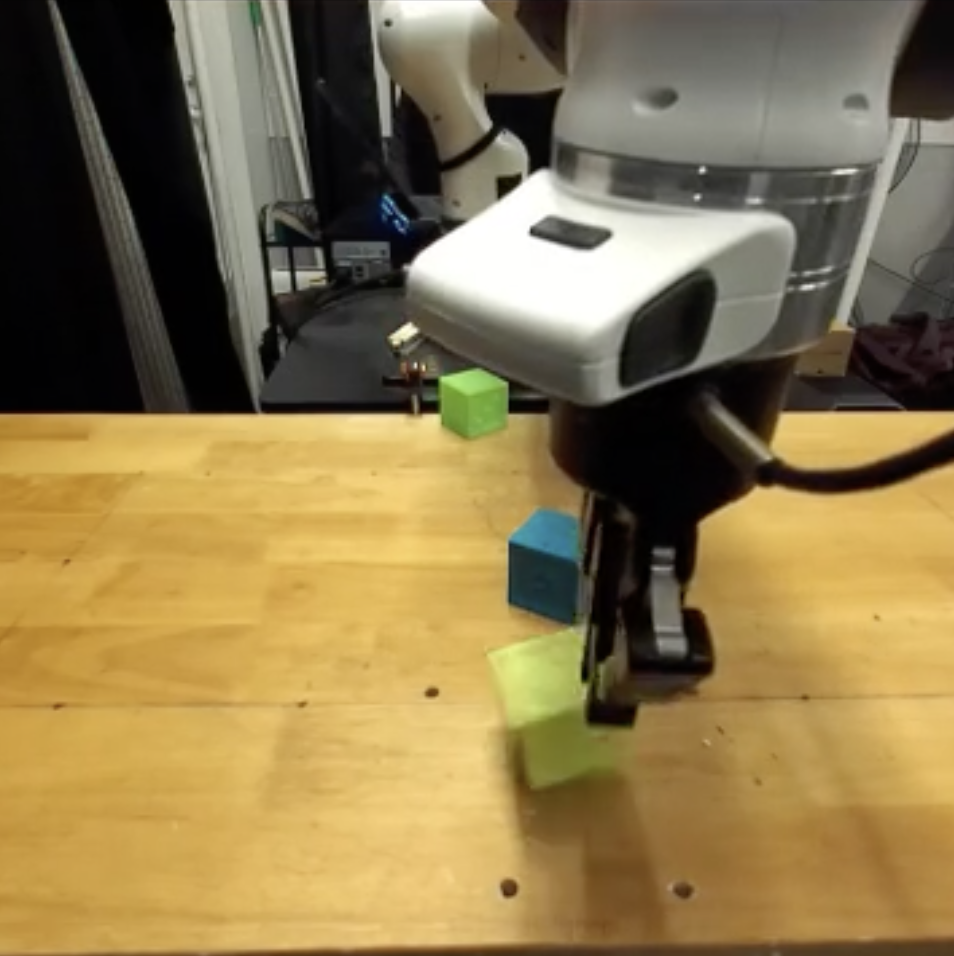}
        \hfill
        \includegraphics[width=0.19\textwidth]{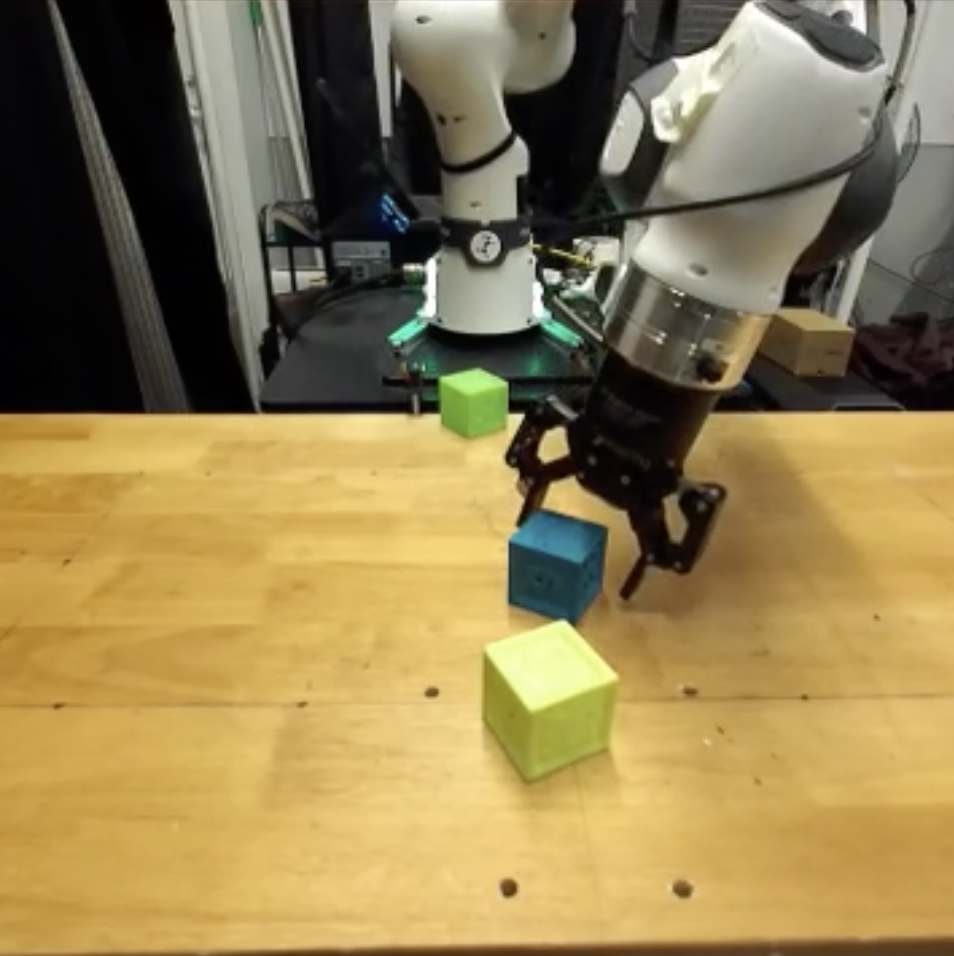}
        \hfill
        \includegraphics[width=0.19\textwidth]{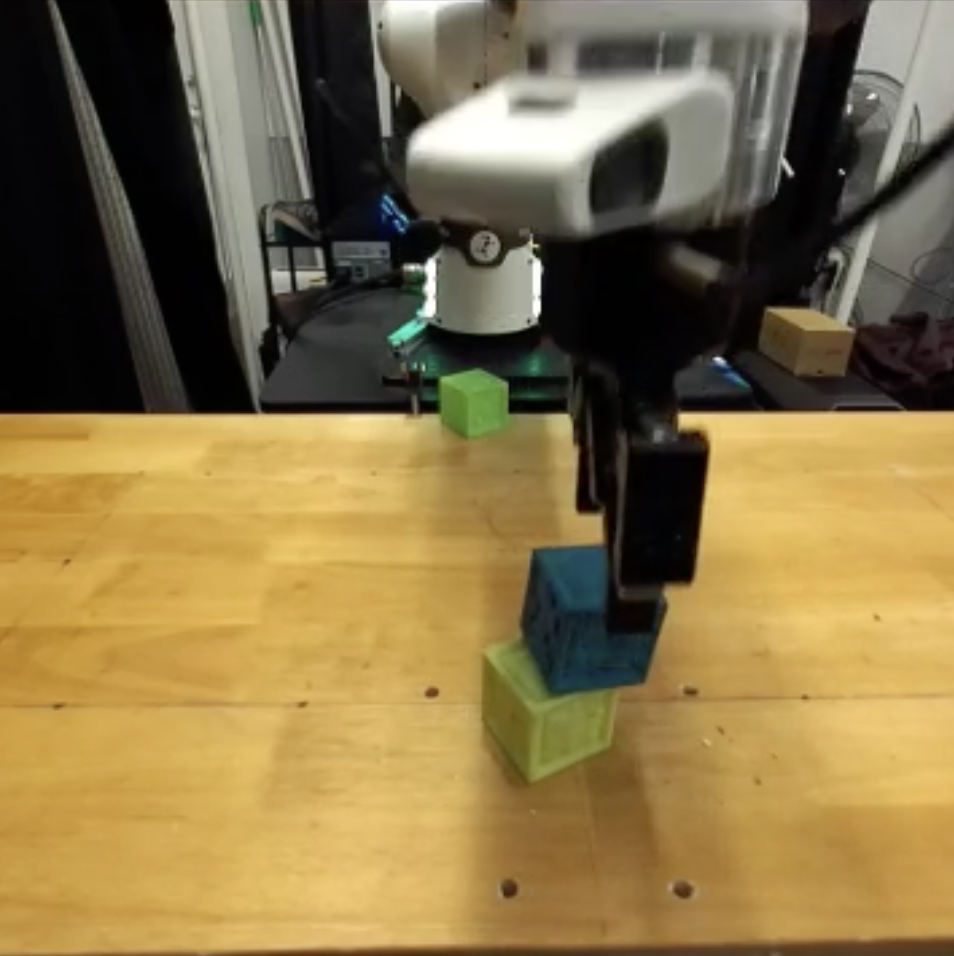}
        
        \caption{"Place the blue cube on top of the yellow cube."}
        \label{fig:restack_real}
    \end{figure*}
}

\def\capagentmain#1{
    \begin{figure}[#1]
        \centering
        \includegraphics[width=0.45\textwidth]{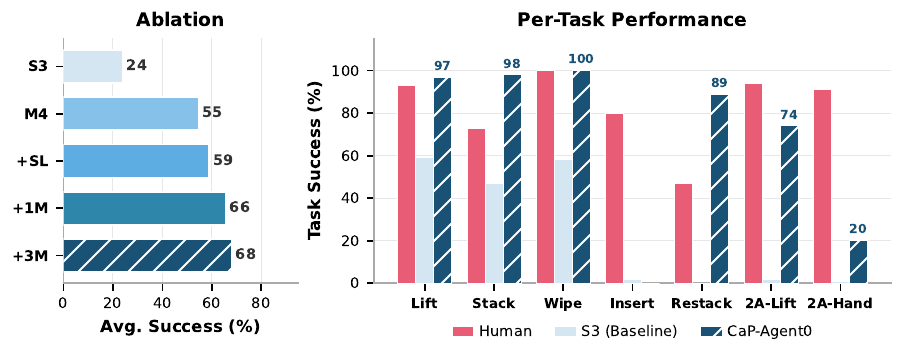}
        \caption{Ablation for \sysname{}. (Left): Combining VDM (M4), skill library (+SL), and parallel queries (+1M: Gemini-3-Pro, +3M: Gemini-3-Pro, GPT-5.2, and Claude Opus) significantly improves over single-turn setup with low-level API. (Right): On 4 out of 7 \benchname{} tasks, \sysname{} achieves comparable or better success rates than human expert code in a single-turn setting.
        }
        \label{fig:capagentmain}
    \end{figure}
    \vspace{-5.0pt}
}

\def\prepostrlreal#1{
    \begin{figure}[#1]
        \centering
        \includegraphics[width=\textwidth]{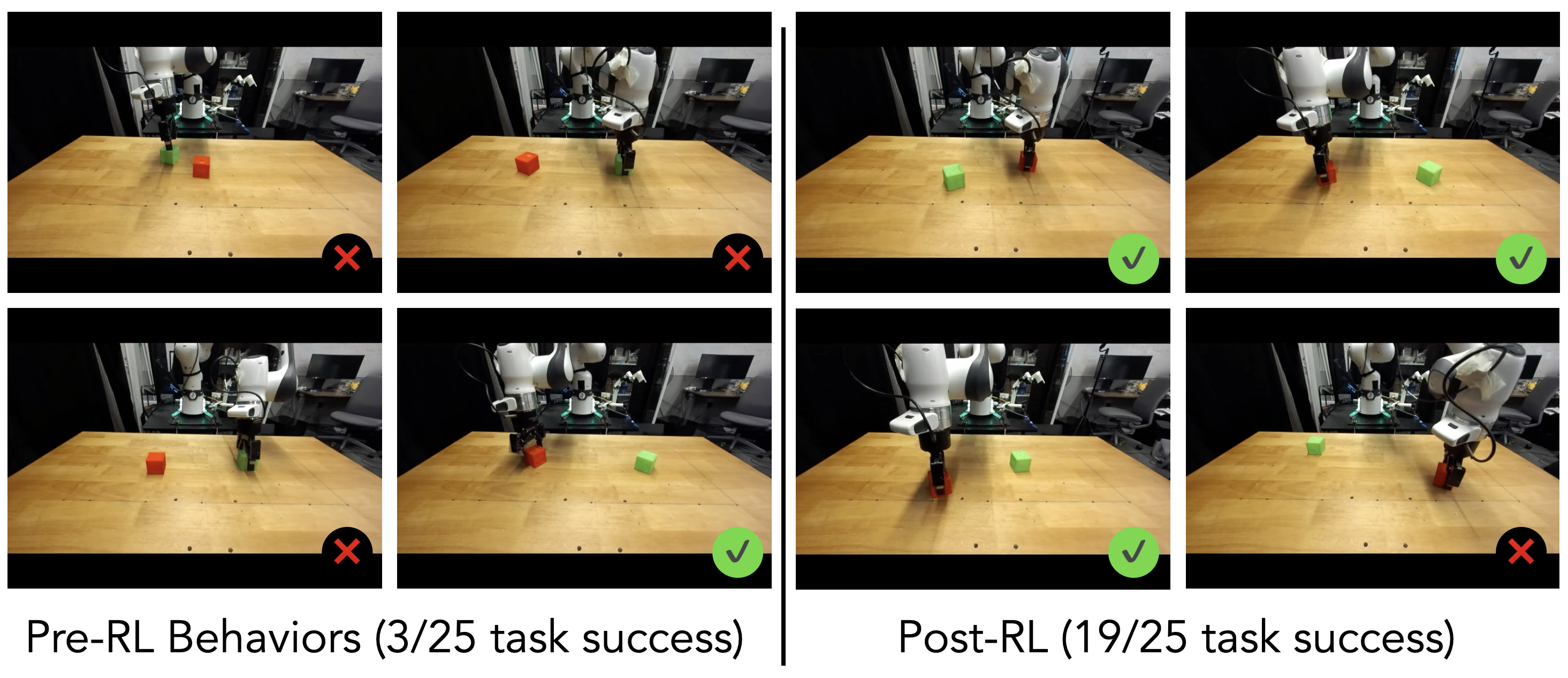}
        \caption{CaP-RL Finetuned Qwen-2.5-Coder-7B-Instruct deployed on the real task \textit{put the red cube on the green cube}. Human Oracle code gets 21/25 task successes in real. Prior to reinforcement learning, we see behaviors such as grasp approaches on the green cube as the first action, rather than a more sensible approach on the red cube first.}
        \label{fig:RL_real_franka}
    \end{figure}
}

\def\postrlgeneralization#1{
    \begin{figure}[#1]
        \centering
        \includegraphics[width=\textwidth]{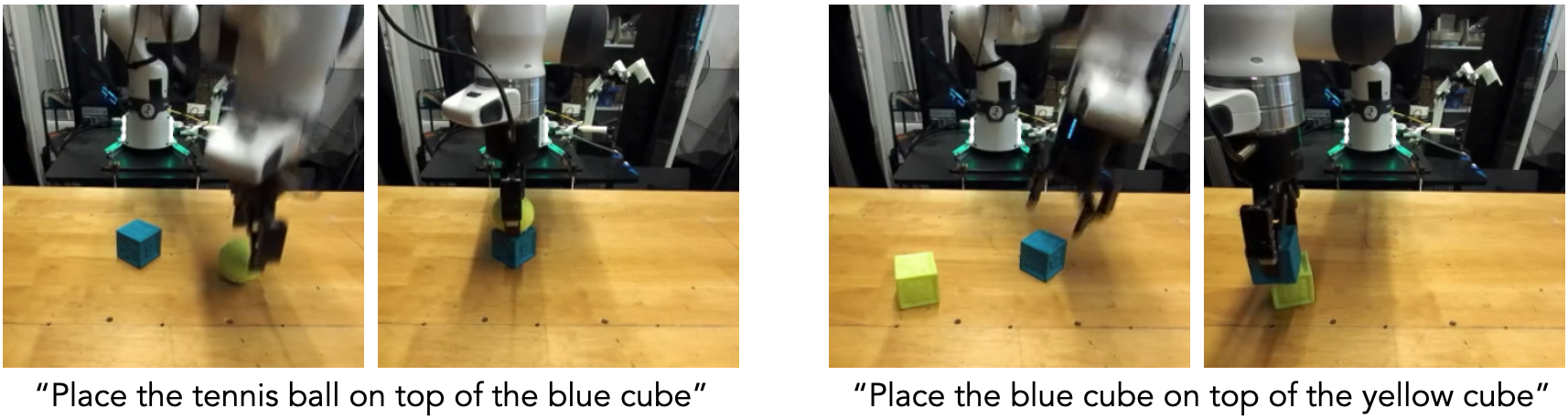}
        \caption{CaP-RL Finetuned Qwen-2.5-Coder-7B-Instruct successfully deployed on similar task variants. Demonstrates that base model is still capable of instruction following.}
        \label{fig:postRLgeneral}
    \end{figure}
}

\def\fullbenchtable#1{
    \begin{figure}[#1]
        \centering
        \includegraphics[width=\textwidth]{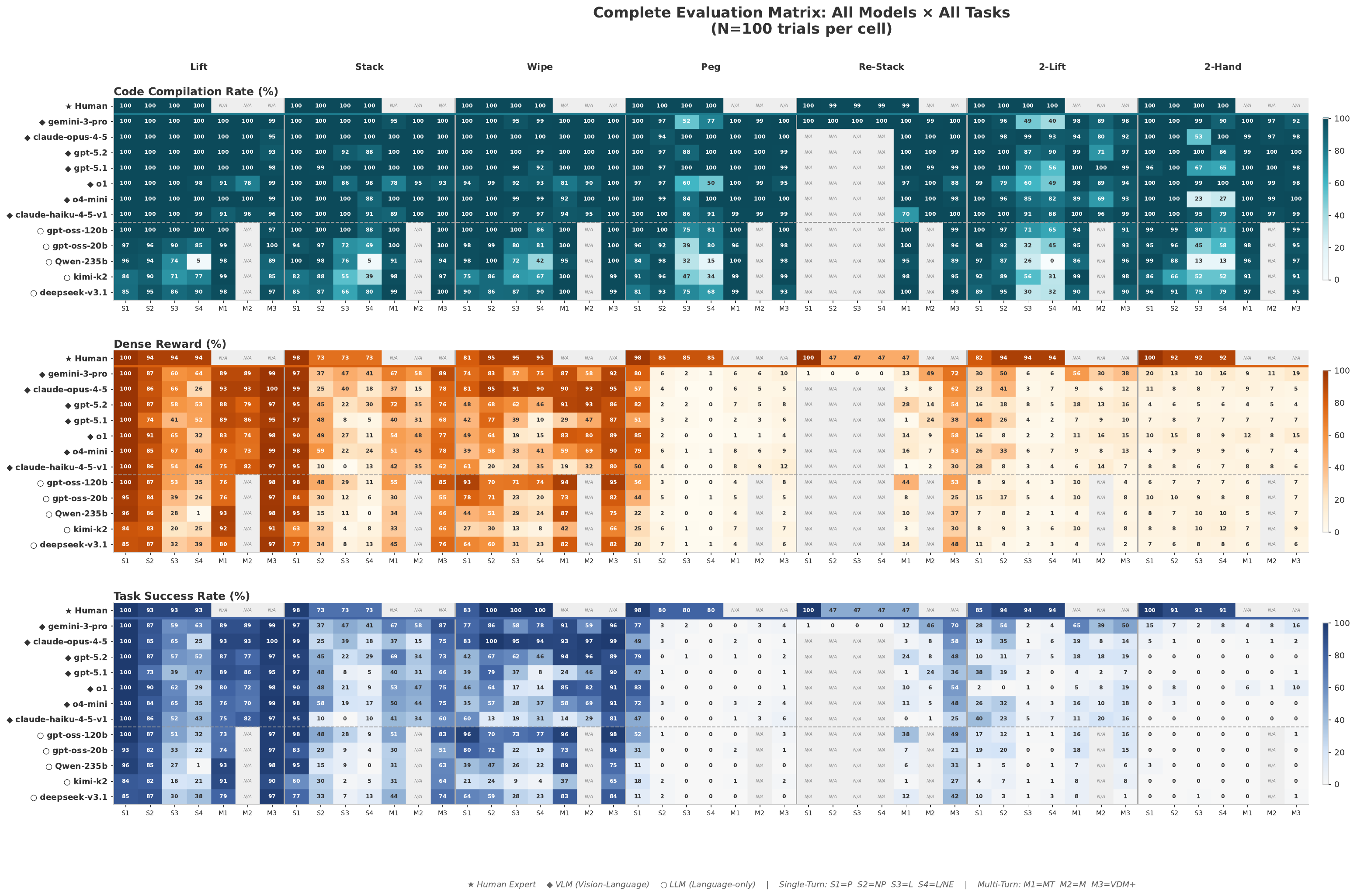}
        \caption{Full \benchname{} results. From top to bottom are code compilation success rates, average task reward, and average task success rate. Note that for \texttt{Re-Stack}, with a simple task prompt, without visual input, we do not observe model writing code to first check the initial condition and then perform the task, resulting in failures. Therefore, we only evaluatd all models for the multi-turn setup.}
        \label{fig:fullbenchtable}
    \end{figure}
}

\begin{abstract}

Code-as-Policy (CaP) is a paradigm in which a language or vision-language model generates executable robot control programs, yet its effectiveness as an autonomous controller for embodied manipulation remains underexplored. Prior CaP systems often rely on high-level, human-designed primitives, making it difficult to separate agent capability from designer-provided scaffolding. We present \textbf{CaP-X}, an open-access framework for systematically studying Code-as-Policy agents in robot manipulation. CaP-X includes four components. \textbf{\gymname{}} is an interactive environment in which coding agents control robots by synthesizing and executing programs that compose perception and control primitives. Building on this foundation, \textbf{\benchname{}} evaluates frontier language and vision-language models across varying levels of abstraction, interaction, and perceptual grounding. 
Across 
12 models, the task success rates improve with human-crafted abstractions but degrade as these priors are removed, exposing a dependence on designer scaffolding. At the same time, we observe that scaling test-time computation with multi-turn interaction, structured execution feedback, visual differencing, automatic skill synthesis, and ensembled reasoning can substantially improve robustness even when agents operate over low-level primitives. These findings motivate \textbf{\sysname{}}, a training-free framework that achieves near human-level reliability on several manipulation tasks in simulation and on real embodiments. \textbf{\rlname{}} explores reinforcement learning with verifiable rewards to improve success rates and supports sim-to-real transfer through a shared code-as-action-space interface. Together, CaP-X provides an open-access platform for advancing embodied coding agents.
Project page: \href{https://capgym.github.io}{https://capgym.github.io}

\end{abstract}

\section{Introduction}
\splashfigure{ht!}
Robots have long been controlled through explicit programs that combine perception, geometry, planning, and feedback~\citep{fikes1971strips, murray1994mathematical, aeronautiques1998pddl, siciliano2008springer}. As robots advanced to continuous, higher-dimensional spaces, these representations were integrated with geometric motion planning~\citep{khatib1986real}, evolving into Task-and-Motion Planning (TAMP)~\citep{kaelbling2011hierarchical}. These \textit{classical} control paradigms achieve robustness through explicit structure: human engineers manually write software that decomposes high-level goals into subtasks, compose perception and control modules, and handle failure and edge cases through trial-and-error and explicit logic. While this can offer strong interpretability and geometric precision guarantees, it relies heavily on human skill and expertise. The manual process can be very time-consuming and often produces task-specific solutions that are difficult to generalize to open-ended environments. 

Driven by successes in foundation models~\citep{devlin2018bert, radford2018improving, radford2019language, brown2020language, chowdhery2023palm, achiam2023gpt, radford2021learning, li2023blip}, another robot control paradigm has emerged in the form of Vision-Language-Action (VLA) models~\citep{brohan2023rt2, kim2024openvlaopensourcevisionlanguageactionmodel, octo_2023, jang2022bc, jiang2022vima, reed2022generalist, 2024rtx, shah2023vint, fu2024icrt, huang2025otter, bjorck2025gr00t, lbmtri2025, intelligence2025pi05}. These approaches learn from large-scale visuomotor datasets to achieve impressive performance on contact-rich tasks such as shirt folding and whole-body loco-manipulation. However, VLAs inherit the limitations of their training data and design: they lack interpretability and struggle to generalize to changes in the environment, new robot embodiments, and long-horizon tasks without additional data collection and retraining.

Recent advances in coding capability of Large Language Models (LLMs) suggest a way to bridge these paradigms: using coding agents to replace the human engineer. Modern coding agents have demonstrated the ability to synthesize executable code, define functions, and debug failures on software engineering benchmarks~\citep{jimenez2024swebench}. Unlike earlier language-conditioned planners limited to function calling~\citep{tellex2011understanding}, today's agents can construct mid- to low-level logic that closely resembles expert code. 

Code-as-Policy (CaP) pioneers~\citep{liang2023code, singh2023progprompt} explored this approach in applications to robotics; they used human-tuned high-level primitives (e.g., \verb|stack_objs_in_order()|) that offer significant task-specific simplifications. 
As a result, it remains unclear how much of the observed performance in robot control stems from the agent itself versus the structure imposed by these primitives. In particular, prior work does not systematically characterize how agent performance changes as such scaffolding is reduced, nor to what extent increased test-time computation—through iterative debugging, skill synthesis, ensembled reasoning, or multimodal grounding—can compensate for operating over lower-level interfaces.

To address this gap, we introduce \textbf{CaP-X}, a unified framework for systematically evaluating and improving code-based robot control agents. At its core is \textbf{\gymname{}}, an interactive environment in which agents directly control robots by generating and executing programs that compose perception and control primitives. \gymname{} integrates 187 tasks from standard robot manipulation simulators (Robosuite~\citep{zhu2020robosuite}, LIBERO-PRO~\citep{liu2023libero, zhou2025liberopro}, and BEHAVIOR~\citep{li2024behavior1k}) under a shared primitive design that is intentionally compatible with both simulation and physical robot systems. 

Building on \gymname{}, we construct \textbf{\benchname{}}, a benchmark designed to systematically study agentic capability along three axes: \textbf{Abstraction Level:} Varying the action space from human-crafted macros (High-Level) to atomic, fundamental primitives (Low-Level); \textbf{Temporal Interaction:} Comparing zero-shot single-turn program generation against multi-turn interaction to quantify capabilities in failure recovery and iterative reasoning; and \textbf{Perceptual Grounding:} Evaluating how different modalities of visual feedback impact the agent's ability to ground task-relevant visual features into code generation.
We instantiate \benchname{} by focusing on a subset of 7 environments from \gymname{} and evaluate 12 state-of-the-art open- and closed-source language models and vision-language models. 

Guided by insights from \benchname{} results, we derive \textbf{\sysname{}}, a training-free agentic framework that augments coding models with multi-turn interaction, visual grounding into text, an automatically synthesized task-agnostic skill library, and parallelized multi-model code generation. 
\sysname{} achieves performance comparable to, and in some cases exceeding, human expert baselines on \benchname{} tasks. Finally, we show that \gymname{} supports \textbf{\rlname{}}--reinforcement learning on the coding agent itself. On-policy post-training with environment rewards improves task success, while synthesized programs transfer directly to real robots. 
This paper makes the following contributions:
\capbenchleveltable{t!}
\begin{enumerate}
    \item \gymname{}, a unified suite of interactive robot coding environments spanning tabletop, bimanual, and mobile manipulation tasks, designed for evaluating and training code-generating multimodal embodied agents.
    \item CaP-Bench, a systematic benchmark that measures robot control performance across tasks and levels of primitives and modalities.
    \item \sysname{}, a training-free, agentic harness combining multi-turn visual differencing, ensembled reasoning, and automatic skill library synthesis.
    \item \rlname{}, reinforcement learning on the coding agent via environment reward.
\end{enumerate}

\section{\gymname{}} 
\gymname{} is a hierarchical control framework built on the standard \textbf{Gymnasium} interface~\citep{brockman2016openai}. It binds a Low-Level Environment loop (a physics simulator or the real world) with a stateful Code Executor loop. 
Architecturally, \gymname{} preserves the native dynamics of underlying simulators, e.g., Robosuite~\citep{zhu2020robosuite}, LIBERO-PRO~\citep{liu2023libero, zhou2025liberopro}, and BEHAVIOR~\citep{li2024behavior1k}, while exposing them through a Read-Eval-Print Loop (REPL) paradigm tailored for coding agents. In CaP-Gym, a code environment ``turn" corresponds to one interaction between the coding agent and a specific robot task instance: the agent receives observations, generates a Python program, and the environment executes it to completion. The program may invoke multiple perception and control primitives, each of which can run the simulator or robot controller for multiple internal updates.

\subsection{Low-Level Perception and Control Primitives}
All computationally intensive perception and control primitives are implemented as stateless services~\cite{uvicorn}, enabling high-throughput parallel evaluation.

\textbf{Perception Primitives} 
Agents access perceptual data from the environment through modular perception primitives that abstract raw sensor data into structured semantic objects, e.g., SAM3~\cite{carion2025sam3} for language-conditioned segmentation and Molmo 2~\cite{clark2026molmo2} for open-vocabulary pointing, alongside standard vision libraries like OpenCV~\citep{opencv_library} and Open3D~\citep{zhou2018open3d}. 

\textbf{Control Primitives} Instead of directly emitting joint-space action commands, agents call motion planners or inverse kinematics solvers such as PyRoki~\cite{kim2025pyroki}. They can handle collision checking, reachability constraints, and action-space transformations, allowing agents to reason in a task-oriented Cartesian space while delegating execution feasibility to the controller. 

\section{CaP-Bench: Evaluating Frontier Models}
\abstractionlevelfigure{t}

\textbf{Models.} We evaluate 12 state-of-the-art vision-language and language models, including closed-source frontier models (Gemini-3-Pro~\cite{google_deepmind_gemini3pro_modelcard_2025}, OpenAI GPT o1~\cite{openai_o1_system_card_2024}, o4-mini~\cite{openai_o4mini_system_card_2025}, 5.1~\cite{openai_gpt5_1_2025} and 5.2~\cite{openai_gpt5_2_2025}, and Claude Haiku 4.5~\cite{anthropic_claude_haiku_4_5_2025} and Opus 4.5~\cite{anthropic_claude_opus_4_5_2025}, and open source models (OpenAI GPT-OSS-20B and 120B~\cite{openai_gpt_oss_120b_2025}, Qwen3 235B~\cite{qwen_qwen3_235b_2025}, Qwen-2.5-Coder-7B-Instruct, Kimi K2 Instruct~\cite{moonshot_ai_2025}, and DeepSeek-V3.1-Terminus~\cite{deepseekai2024deepseekv3technicalreport}). 

\textbf{Simulation Task Suite.} Primary analysis is performed across 7 core tasks ranging from single-arm manipulation to bimanual coordination: \textit{Cube Lift, Cube Stack, Spill Wipe, Peg Insertion, Cube Re-stack, Two-Arm Lift,} and \textit{Two-Arm Handover}. Each task is evaluated with 100 trials per tier, where each tier specifies the available primitives, interaction mode, and feedback/grounding signal. These 7 tasks are an intentionally controlled core for ablating abstraction, iteration, and grounding under matched conditions; the full release of \gymname{} ships 187 tasks (7 Robosuite + 130 LIBERO-PRO + 50 BEHAVIOR) for broader community evaluation. In \cref{sec:capagent}, we further extend this analysis to diverse long-horizon settings using tasks from LIBERO-PRO~\cite{zhou2025liberopro} and BEHAVIOR~\cite{li2024behavior1k}.

\textbf{Protocol}. We evaluate models using \textbf{Zero-Shot Pass@1}~\citep{chen2021codex,jimenez2024swebench}. Within a trial, agents may interact with the environment over one or multiple turns, observing execution feedback and generating subsequent code to recover from errors or extract additional information; the environment is never reset during the trial. Coding agent task performance is compared against that of human expert-written reference solutions under identical environments and primitives (see Appendix~\ref{app:human_expert}).
We introduce 4 single-turn tiers (S1-S4) and 4 multi-turn tiers (M1-M4) to \benchname{}. Refer to \cref{tab:capbench-levels} for a tabular comparison. The tiers vary three axes faced by any code-as-policy practitioner: (1) \textit{primitive abstraction} (human macros vs.\ low-level primitives), (2) \textit{level-of-iteration} (single-turn vs.\ multi-turn with structured execution feedback), and (3) \textit{mode of grounding} (raw visual inputs vs.\ text descriptions from a separate VLM). CaP-Bench isolates these axes via independently controllable tiers.

\subsection{Single-Turn Benchmarks (S1-S4)}
\label{subsec:singleturn}

\textbf{High-Level (S1 \& S2)}: These tiers evaluate the agent's ability to reason with human-designed primitives. We evaluate this in two modes: \textbf{Privileged (S1)}, which uses ground-truth simulation state (masks and object poses), and \textbf{Non-Privileged (S2)}, which relies on real perception modules processing raw RGB-D inputs--the default setting for most prior work. We introduce S1 to disentangle high-level planning from perception noise, establishing a reasoning upper bound that allows us to distinguish between algorithmic failures and visual estimation errors.

\textbf{Low-Level (S3 \& S4)}: In these tiers, human-designed abstractions are replaced by their constituent low-level primitives (e.g., \texttt{solve\_ik()}, \texttt{sam3\_text\_prompt()}), drawn directly from the APIs of each underlying package—reflecting the interfaces that human developers use to control robots. We evaluate two settings: \textbf{(S3)}, where documentation includes usage examples to scaffold low-level composition, and \textbf{(S4)}, where examples are removed and the agent must reason about program structure solely from interface definitions (function signatures and docstrings). See \cref{fig:abstractionlevel_diff} for an illustration of high-level (S1 \& S2) versus low-level (S3 \& S4) primitives; full API details at each abstraction level are provided in Appendix~\ref{app:apis}.

\subsection{Multi-turn Benchmarks (M1-M4)} 

\textbf{Text-Only Multi-turn (M1)}: In this setting, the agent receives the standard output (\texttt{stdout}) and error traces (\texttt{stderr}) from the Python sandbox after each execution turn. This enables a state introspection loop: agents can proactively inject diagnostic print statements to surface hidden symbolic variables (e.g., perception estimates) and utilize these traces to diagnose logical failures and refine code without access to visual ground truth. All other multiturn tiers (M2-M4) retain access to these code execution traces.

\textbf{Multimodal (M2)}: The environment pipes the current RGB observation back into the agent's context window. The M2 tier is only available to multimodal foundation models that accept raw RGB images as input.

\textbf{Visual Differencing Module (M3)}: We introduce the Visual Differencing Module (VDM), which uses a vision–language model to convert visual observations into structured natural language. VDM is provided with the task instruction alongside visual observations. In the first turn, it generates a scene description and extracts task-relevant visual attributes. In subsequent turns, it explicitly describes differences between the previous and current image observations and whether the coding agent has completed the task. The resulting text from the VDM is provided as part of the coding agent’s observation context for code generation.

\textbf{Low-Level with VDM (M4)}: This tier has the same VDM as tier M3 and has access to the same low-level primitives and in-context usage examples as tier S3.  

\abstractionincrease{t!}

\subsection{Discussion}

\textbf{Takeaway 1: A Significant Gap Persists Between Frontier Models and Human Experts in Single-Turn Evaluation.} 
We benchmark open- and closed-source agents against human expert solutions under an identical set of perception and control primitives in a single-turn, zero-shot setting (S4). The human reference is a near-upper-bound: $N{=}7$ paper authors (each with 2+ years of robotics-programming experience) wrote single Python scripts at each tier using exactly the same API primitives available to the model, iterating and debugging until the solution achieved $88.5\%$ average single-turn success (see Appendix~\ref{app:human_expert} for the protocol and effort budget). As shown in \cref{fig:splash_fig}, while closed-source models consistently outperform open-source alternatives and newer architectures exhibit stronger capabilities, none yet match the success rate of human-crafted programs in a zero-shot Pass@1 setting.

\textbf{Takeaway 2: High-Level Abstractions Boost Performance but Limit Expressivity}. 
\cref{fig:abstraction_increase} shows a monotonic increase in task success as primitive abstraction increases, mirroring how prior Code-as-Policies~\cite{liang2023code} approaches relying on high-level primitives report strong zero-shot performance. By collapsing low-level perception, geometric reasoning, and control into human-designed primitives, these abstractions reduce the effective search space and allow models to focus on task sequencing.

However, this gain comes at the cost of expressivity. As abstraction increases, the agent’s action space is increasingly constrained by human priors, imposing a generality ceiling that masks failures in low-level reasoning. In contrast, performance degradation at lower abstraction levels (S3/S4) reflects the difficulty of code synthesis, while enabling expressive behaviors, such as hierarchical perception fallback strategies (\cref{app:perception_fallbacks}), that cannot be represented by fixed high-level primitives. 

This observation motivates a scalable middle ground: rather than relying on human-designed abstractions, agents should be able to recover structure from low-level primitives themselves. In \cref{sec:capagent}, we demonstrate this capability by enabling agents to distill successful execution traces into a reusable skill library. Consequently, we propose that generalist embodied coding agents be evaluated primarily on primitive-level performance, ensuring that success stems from robust reasoning rather than from the inductive biases of an over-engineered, often task-specific, API set.

\codecompilation{t!}

\textbf{Takeaway 3: Closing the Loop with Multi-turn and Visual Grounding Improves Performance.} We study how \textit{multi-turn interaction} and different forms of \textit{visual grounding} mitigate the performance gaps identified in Takeaways 1 and 2.
\visualfeedbackcomp{h}
Allowing agents to iterate and inspect their own execution traces (\texttt{stdout}/\texttt{stderr}, M1) consistently improves performance across all models (\cref{fig:visualfeedbackcomp}), highlighting the importance of explicit execution feedback for debugging and recovery.

Counter-intuitively, directly interleaving raw RGB observations at each turn (M2) degrades performance relative to the text-only M1 baseline. We hypothesize this degradation is due to a cross-modal alignment gap: foundation models are rarely trained to jointly reason over software coding and images of physical task execution, making raw visual inputs difficult to integrate effectively during code synthesis.

Prior work~\cite{hu2025lmgame, wang2026visgym} similarly observes that textually grounded feedback outperforms raw images, though primarily in structured environments with native text states. In contrast, robotic manipulation operates in unstructured, continuous settings without ground-truth textual descriptions. Here, the Visual Differencing Module (M3) bridges this gap by converting visual observations into structured natural language, substantially outperforming both naive image interleaving (M2) and execution-only feedback (M1) across all tasks (\cref{fig:visualfeedbackcomp}).

\textit{Low-Level Primitives with Multi-turn Feedback (M4).} \cref{fig:vdmlvslvsnp} shows that agents operating over low-level primitives augmented with multi-turn feedback not only surpass high-level single-turn (S2) but can reach parity with high-level multi-turn performance (M3). This supports a \emph{test-time compute scaling} hypothesis: robustness can be synthesized at runtime by increasing an agent’s capacity for reasoning, verification, and self-correction over atomic primitives.

Across all multi-turn settings (M1-M4), both coding errors (e.g., exceptions) and physical execution failures (e.g., unstable grasps) are frequently recoverable through iterative interaction. In the absence of explicit visual grounding mechanisms (M1), successful agents compensate by proactively instrumenting perception primitives to expose symbolic state--such as object poses--and performing explicit checks for task completion (e.g., verifying relative object heights to confirm stacking). Visual grounding (M2-M4) reduces the burden of such self-instrumentation but does not eliminate explicit state verification behavior through perception primitives. Instead, the strongest performance emerges when agents combine structured feedback with iterative reasoning, framing multi-turn interaction as a mechanism for hypothesis testing and error recovery in embodied control.
\vdmlvslvsnp{h}

\capagentfigure{t}

\section{\sysname{}: An Agentic Framework for Robot Control}
\label{sec:capagent}
Based on the failure modes and insights identified in \benchname{}, we introduce \textbf{\sysname{}}, a training-free, agentic infrastructure for robot control that augments base foundation models with a specialized multi-turn reasoning loop and a dynamically synthesized skill library.
The architecture of \sysname{} is designed directly to address three key gaps identified in the benchmark.
We demonstrate the efficacy of each design choice in \sysname{} by running ablation studies using the most capable model identified from CaP-Bench (Gemini-3-Pro) bootstrapped with each design choice, shown in Figure~\ref{fig:capagentmain}. A visual walkthrough of the agentic framework is presented in \cref{fig:capagent0_sys}.

\textbf{1. Multi-turn Visual Differencing (VDM):}
Directly adopting the insights from Takeaway 3, \sysname{} integrates the Visual Differencing Module as part of the per-turn observation. By grounding observations in structured text rather than raw pixels, the agent mitigates the specific cross-modal alignment failures identified in the M2 tier.

\capagentmain{t!}
\textbf{2. Auto-Synthesized and Persistent Skill Libraries:}
In analyzing low-level S3 and S4 code generations from CaP-Bench, we found that capable models routinely synthesize helper functions to perform robotics data manipulations.

Motivated by agentic systems that accumulate reusable tools over time~\cite{wang2023voyager}, \textbf{\sysname{}} introduces an \textit{automatically synthesized, task-agnostic skill library} that persists across trials. Rather than requiring the agent to repeatedly re-derive low-level utilities, the library onboards commonly recurring implementation patterns and offloads fragile low-level logic, allowing the coding agent to focus on high-level semantic planning. Importantly, unlike fixed human-designed high-level APIs, these skills are \emph{discovered}: they emerge from successful executions and retain the expressivity of low-level interfaces while improving robustness through reuse.

The library is constructed via an automated synthesis pipeline that can be executed by the agent itself. Specifically, we collect all successful S3-tier rollouts pooled across all 12 models and 7 Robosuite tasks (i.e., the library is not model-specific), extract function definitions via regular-expression matching, and prompt Gemini-3-Pro to identify frequently recurring, task-agnostic logic. This yields a compact library of 9 verified, task-agnostic primitives (Appendix~\ref{app:skill_lib}).
While the current implementation performs a single synthesis pass, the process is inherently \emph{iterative}. As additional successful executions are accumulated, the agent can continue to update its skill library over time. 
A complete list of the nine synthesized functions currently utilized by \sysname{} is provided in Appendix~\ref{app:skill_lib}.

\textbf{3. Parallel Reasoning:}
Results from tiers S2 and S3 indicate that failures often stem from insufficient test-time exploration rather than a lack of capability. To address this, \sysname{} employs a parallel reasoning strategy inspired by recent ensemble methods \cite{pan2025learning, jin2025learning, rodionov2025hogwild}. At each turn, the system concurrently samples candidate solutions via two configurations: Single-model: 9 queries to one model. Multi-model: 3 queries each to GPT-5.2, Claude Opus 4.5, and Gemini-3-Pro. To maximize output diversity, we vary sampling temperatures (see Appendix \ref{app:model_ensemble_details}). A central coding agent then synthesizes these candidates into a final code snippet. This parallel approach is applied to both code generation and to the decision-making process for multi-turn continuations.

\subsection{\sysname{} Performance on \benchname{}}
Evaluated over 100 trials per task, \sysname{} significantly outperforms single-turn baselines by integrating visual differencing, a self-synthesized skill library, and parallel reasoning (\cref{fig:capagentmain}). Despite operating solely on low-level primitives, the system achieves success rates comparable to or exceeding human-written programs on 4 out of 7 tasks, narrowing the gap toward expert-level performance.

\subsection{CaP-Bench++}
The 7 core tasks of CaP-Bench isolate abstraction, iteration, and grounding under matched conditions; CaP-Bench++ extends this to compare coding agents directly against (i) state-of-the-art VLA policies and (ii) iterated human-written code on a broader task distribution. Rather than running all 12 models across every extended task--prohibitively expensive at this scale--we focus the comparison on \sysname{}.
We further evaluate the performance of \sysname{} on subsets of tasks from LIBERO-PRO and BEHAVIOR. We summarize the results of \sysname{}  on 30 manipulation tasks from  LIBERO-PRO in Table \ref{tab:liberoproAVG} and compare them against three state-of-the-art VLA methods: OpenVLA~\cite{kim2024openvlaopensourcevisionlanguageactionmodel}, $\pi_0$~\cite{black2024pi0visionlanguageactionflowmodel}, and $\pi_{0.5}$~\cite{intelligence2025pi05}. Although these VLA methods are training-based, \sysname{} is a training-free method that has comparable or exceeds the performance of VLA post-training on these tasks. LIBERO-PRO~\cite{zhou2025liberopro} extends the LIBERO~\cite{liu2023libero} benchmark by increasing the perturbations of tasks by reasonable amounts along object attribute perturbations, initial position perturbations, instruction perturbations, and environmental perturbations. \sysname{} and these VLA methods are evaluated along initial position perturbations (Pos) and instruction perturbations (Task). Under Pos perturbations, the initial positions of objects in the scene are swapped with one another (ex: the position of the frypan and moka pot) and under Task perturbations, the instruction is changed so that another object in the scene is manipulated (ex: ``Put the moka pot on the stove" $\rightarrow$ ``Put the frypan on the stove"). Since VLAs are trained on a different instruction distribution, they perform poorly under task perturbations, whereas CaP-Agent0 remains robust to instruction variations. Furthermore, \sysname{} achieves performance comparable to $\pi_{0.5}$ under initial position perturbations.

\liberoproAVGcomparison{t!}
\simbktask{t!}
We evaluate our method on two long-horizon mobile manipulation tasks in BEHAVIOR, where an R1Pro wheel-based humanoid is required to pick up a radio from a table and a soda can from the floor. The results over 25 trials for each task are summarized in \cref{tab:b1k}, reporting both navigation success rates and task completion success rates. In both tasks, the robot may start with the target object outside its field of view and must actively search for and navigate toward it. Navigation is considered successful if the robot reaches a location within 1 m of the target.

The robot can adjust its camera in both horizontal and vertical directions and move both its base and arm to reach the object, resulting in a substantially larger action space than in tabletop manipulation settings. Moreover, potential collisions with surrounding furniture may prevent the robot from reaching the desired pose, leading to partial execution of planned trajectories and increased task complexity.

In the radio pickup task, although the robot often successfully locates and approaches the object, it may lose sight of it or encounter severe occlusions when navigating too closely, due to its limited field of view. This frequently results in missing or poor grasp poses and constitutes a major failure mode for both S3 and the human policy. In contrast, \sysname{} mitigates this issue by repositioning the robot to obtain a better view, achieving substantially higher success rates.
For the soda can pickup task, the small object size makes accurate vertical camera alignment critical for localization. S3 often fails by adjusting the camera vertically too early, leading to unsuccessful searches within the time limit. \sysname{} adapts its search strategy based on feedback, improving navigation success. Furthermore, grasping may fail when the can is knocked over, further reducing S3’s completion rate. \sysname{} can resample grasp poses after such disturbances and successfully recover, achieving significantly higher task success and similar performance as human expert.

\subsection{Let the Agent Experience the Real World}

The design of the \gymname{} environment loop deliberately allows it to directly interface with real-world robot perception and control interfaces, and we additionally demonstrate zero-shot performance of \sysname{} in unseen real-world tasks on real robot embodiments including the Franka Panda and AgiBot G1 without requiring any major cross-embodiment modifications (with the exception of single arm to bimanual control primitive modifications). With no post-training, off-the-shelf VLMs such as Gemini-3-Pro and Claude Opus 4.5 can perform complex long-horizon robotics reasoning and manipulation tasks following natural language instructions. For example, when asked to ``find the object under one of the cups" (Figure \ref{fig:mech_search}), it mechanically searches through all cups with closed-loop feedback from vision. When the robot is asked to solve a math problem presented in the physical world (Figure \ref{fig:symbolic}), it perceives the visual cue, thinks, and selects the correct blocks on the first attempt. We also demonstrate embodied reasoning ability in Figure \ref{fig:everything}, where \sysname{} exhibits common sense physics reasoning and understands the sensible stacking order for objects of various shapes. Optionally, a human-in-the-loop can also interactively correct the robot's behavior by providing additional feedback in between turns. More details are documented in Appendix \ref{app:real-world-agibot}.

\section{\rlname{}}

\gymname{} enables on-policy reinforcement learning with verifiable rewards (RLVR) directly on the coding agent. To demonstrate this, we apply Group Relative Policy Optimization (GRPO)~\cite{shao2024deepseekmath, guo2025deepseek} to post-train a Qwen2.5-Coder-7B-Instruct base model~\cite{hui2024qwen2}.

\textbf{Methodology.} We RL post-train on three tasks: \textit{Cube Lift, Cube Stack,} and \textit{Spill Wipe}. To ensure stable convergence, we train using the privileged state-based APIs of tier S1. This avoids the noisy reward signals present in tier S2, where compounding perception and control errors can cause otherwise correct programs to fail during execution, introducing credit assignment ambiguity similar to that observed in G1~\cite{chen2025g1}. 

\caprltab{}
\textbf{Simulation Results.} Post-training for 50 iterations per task significantly improves code compilation rates and strategic robustness. When evaluated on S2 (noisy perception), the RL post-trained model achieves substantial gains over the base model, as detailed in \cref{tab:rl_combined}.

\textbf{Sim-to-Real Transfer.} A key property of \rlname{} is that what transfers across the sim-to-real boundary is the \emph{code-as-action-space}: the agent learns to compose shared perception and control tools that are fixed across simulation and reality, rather than mapping raw visual features to motor commands. We validate this on a Franka Emika robot, where the agent retains high success rates for cube lifting (84\%) and stacking (76\%), demonstrating that strategies learned in simulation remain robust under real-world perception noise on these tasks. Please refer to \cref{app:reinforcement_learning} for comparing code generation before and after RL post-training.

\section{Related Work}
\textbf{Code as Policies.}
A growing line of work explores programmatic robot control, where LLMs generate executable code that orchestrates perception and control modules~\citep{gupta2023visprog, yao2022react}. In robotics, this paradigm spans grounding plans in affordances~\citep{ahn2022saycan}, composing APIs for closed-loop behaviors~\citep{singh2023progprompt, liang2023code}, generating Python programs over perception APIs~\citep{huang2023instruct2act, mu2024robocodex, goldberg2024bloxnet}, and modular vision-language agentic pipelines~\citep{shi2025maestro, huang2022inner, geminirobotics1d52025}. Structured intermediate representations such as PDDL~\citep{aeronautiques1998pddl} and Signal Temporal Logic~\citep{liu2023llmplanning, chen2024autotamp} and persistent state tracking~\citep{yoneda2024statler} improve plan reliability, while executable code is empirically a superior agent action representation~\citep{wang2024codeact}, with agents benefiting from iterative self-refinement~\citep{shinn2023reflexion, madaan2023selfrefine} and inference-time sampling~\citep{wang2023selfconsistency, snell2024scaling}. Despite this progress, most coding-agent work on robot control still relies on high-level, human-crafted APIs encoding significant task structure. \gymname{} exposes the full primitive stack down to joint-level perception and control, and \benchname{} evaluates frontier models across abstraction tiers and inference-time strategies.

\textbf{Skill Synthesis and RL with LLM-Generated Code.}
A large body of work uses LLMs as \emph{static} code generators for reward functions, curricula, and skill synthesis with a frozen LLM and a separate trained policy~\citep{ma2024eureka, yu2023l2r, liang2024eurekaverse, ma2024dreureka, du2023ellm, wang2024robogen, ahn2024autort}; complementarily, RL with verifiable rewards (RLVR) improves the model itself across reasoning, code, and agentic settings~\citep{guo2025deepseek, shao2024deepseekmath, wei2025swerl, pan2024training, feng2026retool}. \rlname{} extends RLVR to robot manipulation by directly fine-tuning the language model via GRPO on physics-simulation outcomes, rather than using the LLM to produce reward code for a separate policy.

\textbf{Benchmarks for robotics and embodied agents.}
Robotic manipulation benchmarks~\citep{zhu2020robosuite, liu2023libero, li2024behavior1k, mees2022calvin, james2019rlbench, yu2020metaworld} evaluate fixed policy interfaces rather than executable program synthesis with tiered APIs and multi-turn debugging; code benchmarks~\citep{chen2021codex, jimenez2024swebench, pan2024training} lack embodied perception; and embodied-agent benchmarks~\citep{yang2025embodiedbench, chen2025g1, wang2026visgym, liu2024agentbench, shridhar2020alfred} broaden to multimodal interactive environments but do not require executable robot-control code across abstraction tiers. \gymname{} targets this intersection.

\section{Conclusion}

We introduce CaP-X, a unified framework for benchmarking and improving coding agents for robot control. CaP-X consists of CaP-Gym, CaP-Bench, CaP-Agent0, and CaP-RL, enabling controlled evaluation across abstraction levels, interaction modes, and learning paradigms. CaP-X frames robot control as a problem of machine intelligence—where agent design, inference-time computation, perception, and control are co-studied—providing a testbed for evaluating and advancing general-purpose embodied intelligence.

\section{Impact Statements}

This paper presents work whose goal is to advance the field of machine learning. There are many potential societal consequences of our work, none of which we feel must be specifically highlighted here.

\bibliography{example_paper}
\bibliographystyle{icml2026}

\newpage
\newpage
\appendix
\onecolumn

\section{Future Works}
Programmatic control performs well on long-horizon, reasoning-heavy tasks, but remains brittle for contact-rich behaviors that require tight visual servoing and continuous feedback (e.g., insertion or pouring). One promising direction is hybrid CaP-VLA policies, in which a coding agent manages high-level task logic and recovery while deferring low-level execution to VLA policies. Results from CaP-Bench further highlight several avenues for improving language-model-based agents, including stronger embodiment-aware planning and reasoning, more effective grounding of task-relevant visual information into code generation, improved test-time search, and agent or prompt optimization methods~\cite{khattab2024dspy}. From a robotics perspective, robustness may further improve by incorporating optimization-based control primitives that allow agents to specify task-level constraints and account for collision avoidance during motion planning, rather than relying solely on inverse kinematics solutions that may be suboptimal when directly interpolated to in joint space. Expanding CaP-Gym to additional environments and classical robotics problems (e.g., active and interactive perception) would further stress-test agentic reasoning.

\section{Interactive Real-World Setup}
\label{app:real-world-agibot}

In this section, we document the interactive real-world experience with \sysname{} centered around a chat-based web UI, where the user can propose new tasks to the robot, view the step-by-step tool-use results, and offer feedbacks for multi-turn improvement. We will show how this system enables a real-world robot (AgiBot G1) to zero-shot complete novel tasks.

\subsection{User Interface}

\begin{figure}
    \centering
    \includegraphics[width=1.0\linewidth]{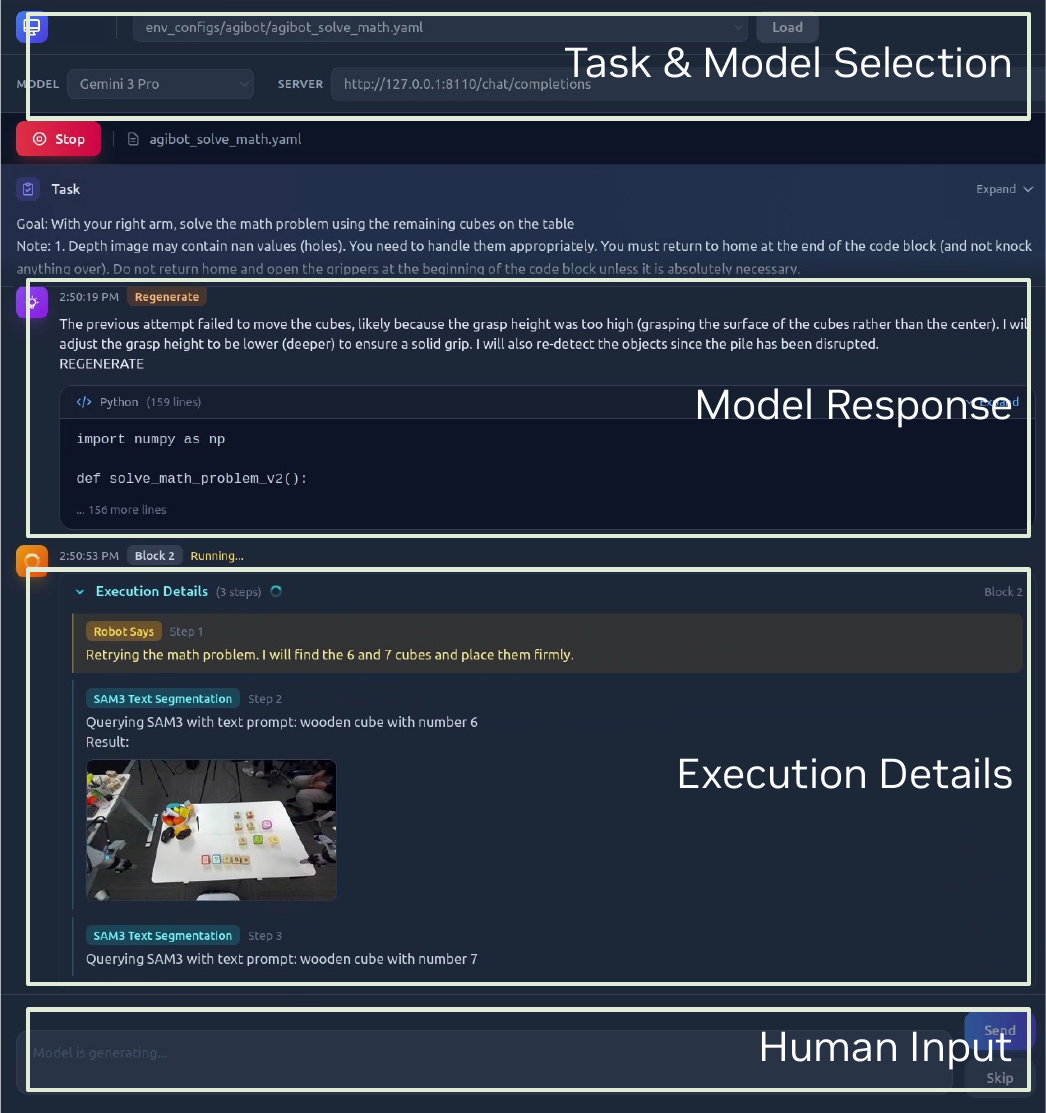}
    \caption{An example of the web UI, where users can select the task and model at the top, view the model’s responses and execution details in the middle, and provide instructions at the bottom.}
    \label{fig:ui}
\end{figure}

\begin{figure}
    \centering
    \includegraphics[width=0.8\linewidth]{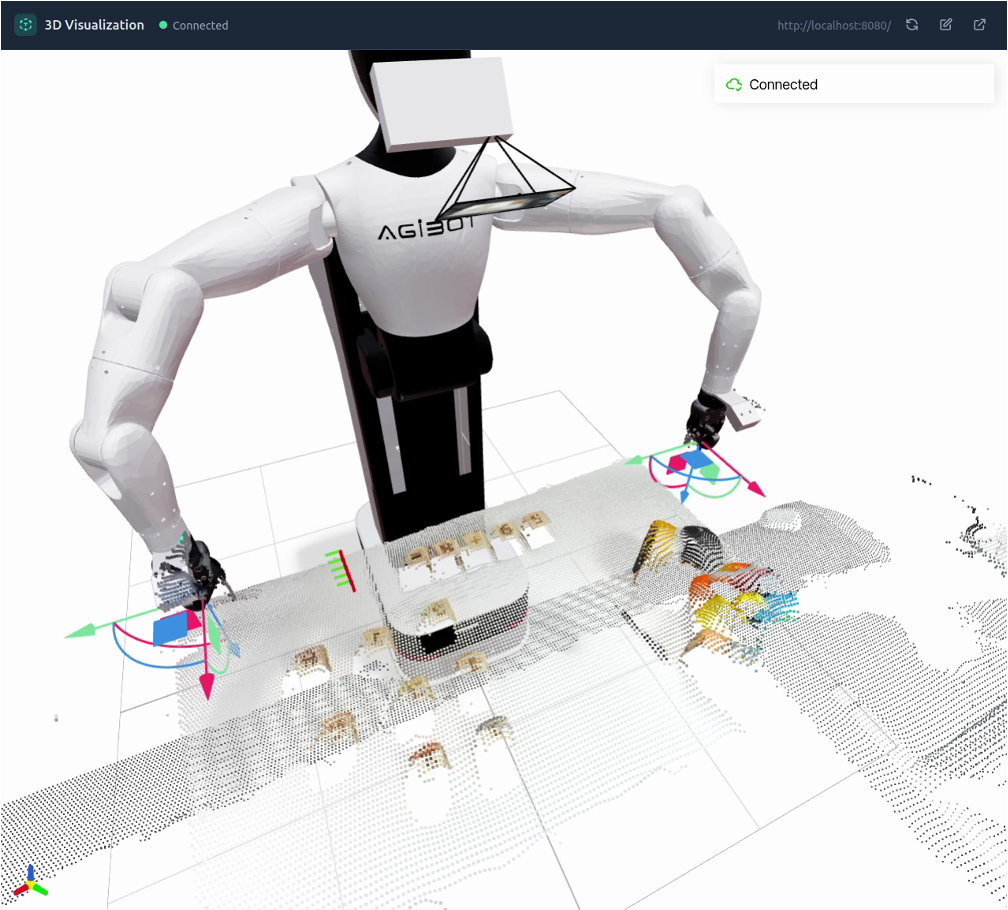}
    \caption{Viser~\citep{yi2025viser} visualization section of the web UI.}
    \label{fig:placeholder}
\end{figure}

On the chat UI shown in Figure \ref{fig:ui}, the user start with choosing task configurations and models. Then they can view the scene description and code generated by the model. As the code is executed on the robot, they also see the step-by-step progress of each tool call, such as the segmentation masks returned by SAM3, Points annotated by Molmo 2, and wrist camera views when the robot arm moves. In between turns, the user can provide additional feedback to the model by typing in the chatbox on the bottom.

The UI also features a 3D visualization powered by Viser~\citep{yi2025viser}, which shows the live robot URDF, trajectories, and depth pointcloud.

\subsection{Tools}
\label{app:tools}

\sysname{}, specifically on the AgiBot G1, has access to these APIs for perceiving, reasoning, and interacting.

\begin{itemize}
    \item \verb|get_observation|: returns a dictionary of the current head camera image (RGB, depth, and intrinsics), left and right wrist camera images, end-effector 6-DoF force-wrenches, and joint configuration.
    \item \verb|segment_sam3_text_prompt|: given an image and a text prompt, returns a list of segmentation masks and scores.
    \item \verb|segment_sam3_point_prompt|: given an image and a point coordinate on the image, returns a list of segmentation masks and scores.
    \item \verb|point_prompt_molmo|: given an image and a text prompt, returns a list of points on the image pointed out by Molmo 2.
    \item \verb|open_gripper|: opens gripper.
    \item \verb|close_gripper|: closes gripper.
    \item \verb|goto_pose|: Plans and executes IK to a pose specified by a position and orientation, optionally with a pre-grasp offset and max force-wrench.
    \item \verb|matrix_to_pose_wxyz_xyz|: Converts a $4\times4$ transformation matrix to a pose represented in a quaternion and a 3D position.
    \item \verb|pose_wxyz_xyz_to_matrix|: Converts a pose to a transformation matrix.
    \item \verb|euler_to_quaternion_wxyz|: Converts extrinsic XYZ Euler angles to a quaternion.
    \item \verb|quaternion_wxyz_to_euler|: Converts quaternion of extrinsic XYZ Euler angles.
    \item \verb|query_vlm|: Ask a question with text and images to an expert VLM model (Gemini-3-Pro)
    \item \verb|go_forward|: Move forward 1 meter.
    \item \verb|turn_left_45_degrees|: Turn left 45 degrees.
    \item \verb|turn_right_45_degrees|: Turn right 45 degrees.
    \item \verb|goto_planar_position|: Plan and execute IK to a position with planar motion constraint, but no orientation (yaw) requirement.
    \item \verb|convert_depth_to_pointcloud|: Given a depth image, intrinsics, and extrinsics, convert the depth image into a pointcloud.
    \item \verb|forward_kinematics|: Given a robot configuration, compute the end-effector pose.
    \item \verb|overlay_eef_axis_on_image|: Visualize end-effector pose on a camera image by rendering the 3D axis on top.
    \item \verb|overlay_segmentation_masks|: Visualize segmentation masks on an image.
    \item \verb|say_something|: Convey some intent to the user.
\end{itemize}

Among the tools, the 3D rigid body transformation helpers are provided although technically not strictly necessary. We find that when not provided, these functions are almost always written by the models who make mistakes from time to time.

Visualization tools such as \verb|overlay_eef_axis_on_image| and \verb|overlay_segmentation_masks| enables advanced multi-modal reasoning when the output images are passed to an expert VLM via \verb|query_vlm|. This allows, for example, code execution conditioning based on if the expert VLM answered "which of these segmentation masks should I pick?".

\subsection{Real World Tasks}

In this section, we show visualizations of \sysname{} completing tasks in the real world. All experiments are \textit{zero-shot}. The agent's execution details are condensed and presented as pseudo sub-steps, while \sysname{} writes the underlying code in greater detail using tools available in \ref{app:tools}. Raw code example is available in Appendix \ref{app:case_study}.

\subsubsection{Needle in a Haystack}
\begin{figure}
    \centering
    \includegraphics[width=0.9\linewidth]{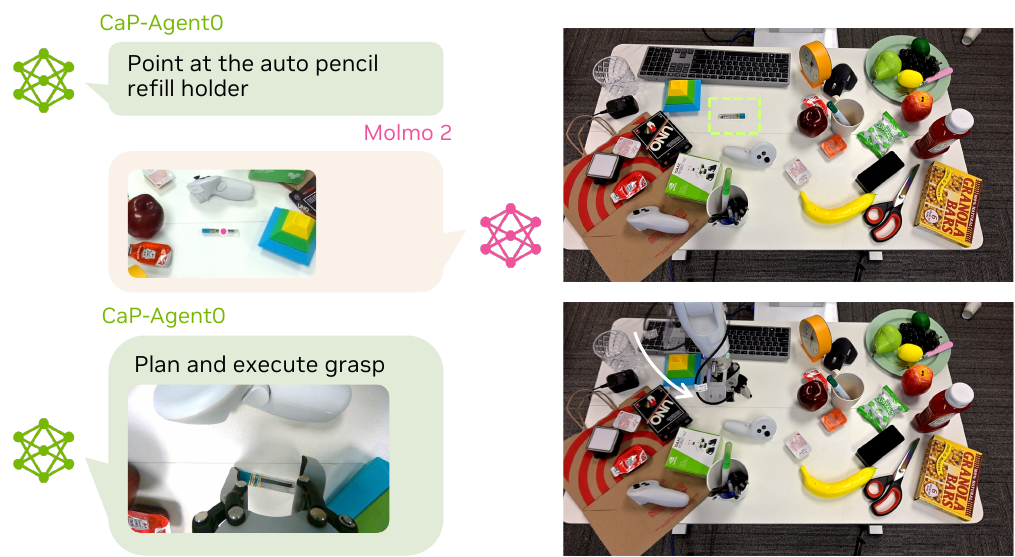}
    \caption{An example of \sysname{} locating and grasping an auto pencil refill holder, where it uses Molmo2 to locate the object.}
    \label{fig:needle}
\end{figure}

In this task, we evaluate the ability of \sysname{} to find “needles in a haystack.” Specifically, in a cluttered scene containing diverse objects, the robot is tasked with locating and grasping an auto pencil refill holder, a relatively uncommon item. Such uncommon objects are often challenging for end-to-end learning policies such as VLAs. In contrast, \sysname{} leverages pretrained VLMs to successfully localize and retrieve the target object, as shown in~\cref{fig:needle}.

\subsubsection{Mechanical Search}
\begin{figure}[h!]
    \centering
    \includegraphics[width=0.8\linewidth]{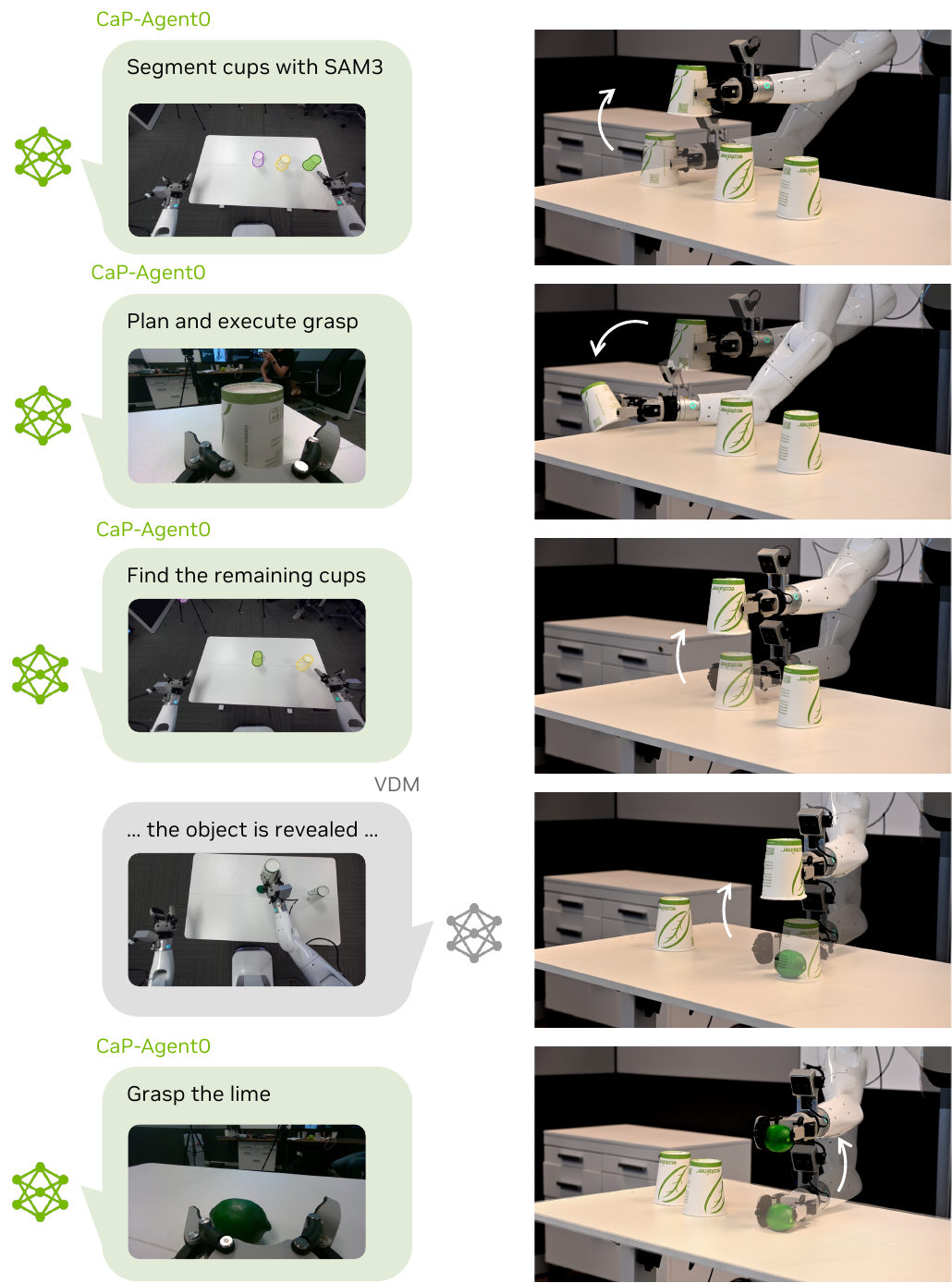}
    \caption{Mechanical search planning and execution example. The left column shows the agent planning process and the right column shows the physical robot execution.}
    \label{fig:mech_search}
\end{figure}

In this task, three inverted cups are placed on a table, and the robot is instructed to retrieve a green lime hidden beneath one of them. This occluded object retrieval problem is commonly referred to as mechanical search~\cite{huang2022mechanical}. Because the lime is concealed by the cups, the robot must systematically explore each cup to locate it. A representative planning and execution example is shown in~\cref{fig:mech_search}. 

\subsubsection{Multimodal Symbolic Reasoning}
\begin{figure}
    \centering
    \includegraphics[width=0.9\linewidth]{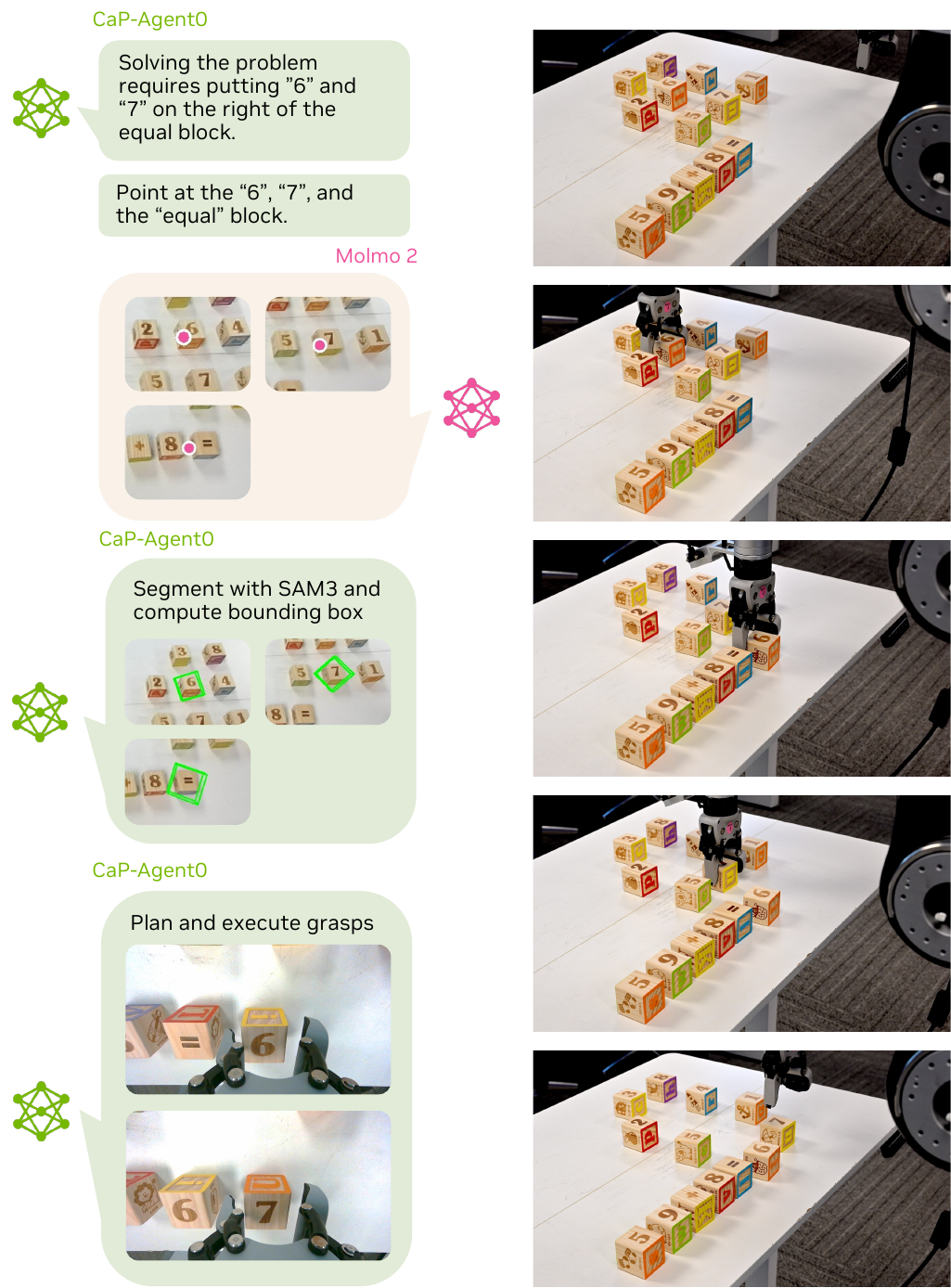}
    \caption{Example of a Multimodal Symbolic Reasoning task, where the robot is asked to solve the problem of 59 plus 8.}
    \label{fig:symbolic}
\end{figure}

In this task, we evaluate the multimodal symbolic reasoning capabilities of \sysname{} by requiring it to solve a mathematical equation formed by numbered wooden blocks. The robot must first perceive the scene to interpret the equation and then grasp the correct block and place it in correct location to complete the solution, as shown in~\cref{fig:symbolic}.

\subsubsection{Learning from Human Feedback}
\begin{figure}[h!]
    \centering
    \includegraphics[width=0.85\linewidth]{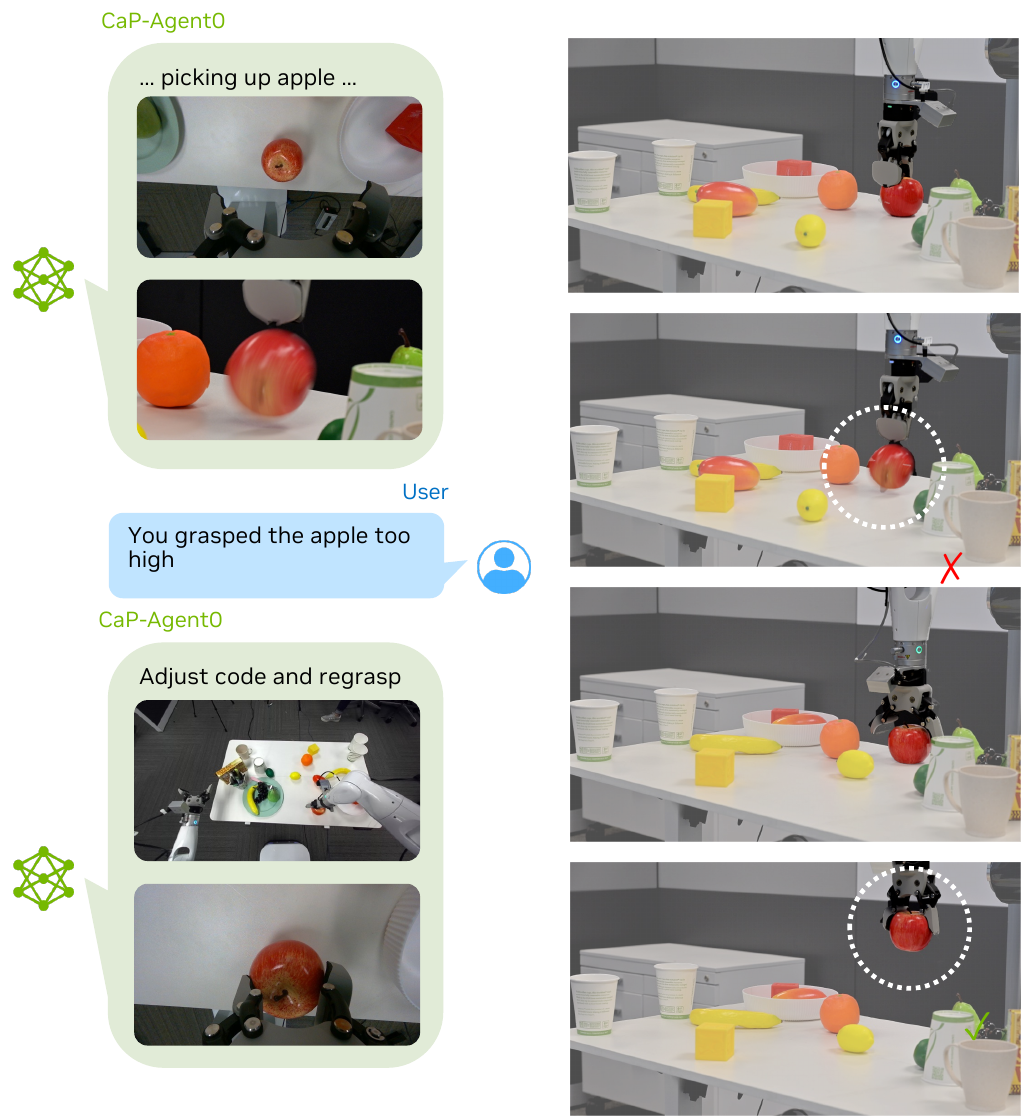}
    \caption{Example of picking up an apple. While the first attempt fails, \sysname{} takes human feedback as input, adjusts the code and completes the task successfully.}
    \label{fig:feedback}
\end{figure}

\sysname{} is able to incorporate human feedback and adjust its code generation accordingly to complete the task. As shown in~\cref{fig:feedback}, the robot is instructed to pick up an apple but fails in the initial attempt due to an excessively high grasp pose. After receiving the feedback ``grasped the apple too high,'' \sysname{} modifies the generated code and successfully completes the task on the second attempt.

\subsubsection{Embodied Reasoning}
\begin{figure}
    \centering
    \includegraphics[width=0.9\linewidth]{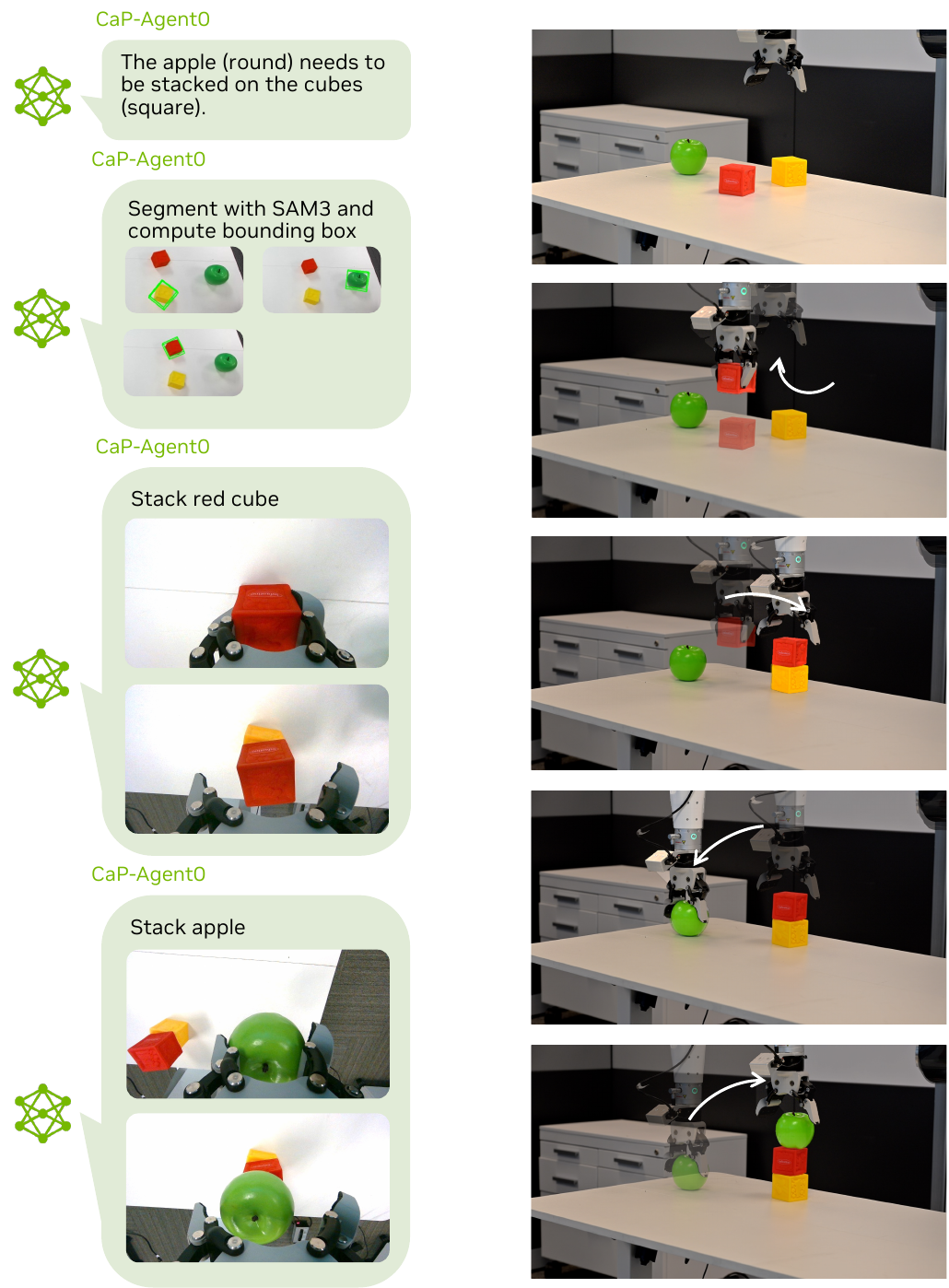}
    \caption{Example of \sysname{} stacking round objects on top of square objects to form the tallest possible stable stack.}
    \label{fig:everything}
\end{figure}

This task evaluates the embodied reasoning ability of \sysname{}, where the robot is asked to ``stack objects as high as possible''. Because the scene contains both square and round objects, the robot must reason about their physical properties to determine a stable stacking strategy. As shown in~\cref{fig:everything}, \sysname{} learns to place round objects on top of square ones to maximize stability.

\subsubsection{Tool Generalization with Domain Knowledge}
\begin{figure}[h!]
    \centering
    \includegraphics[width=0.8\linewidth]{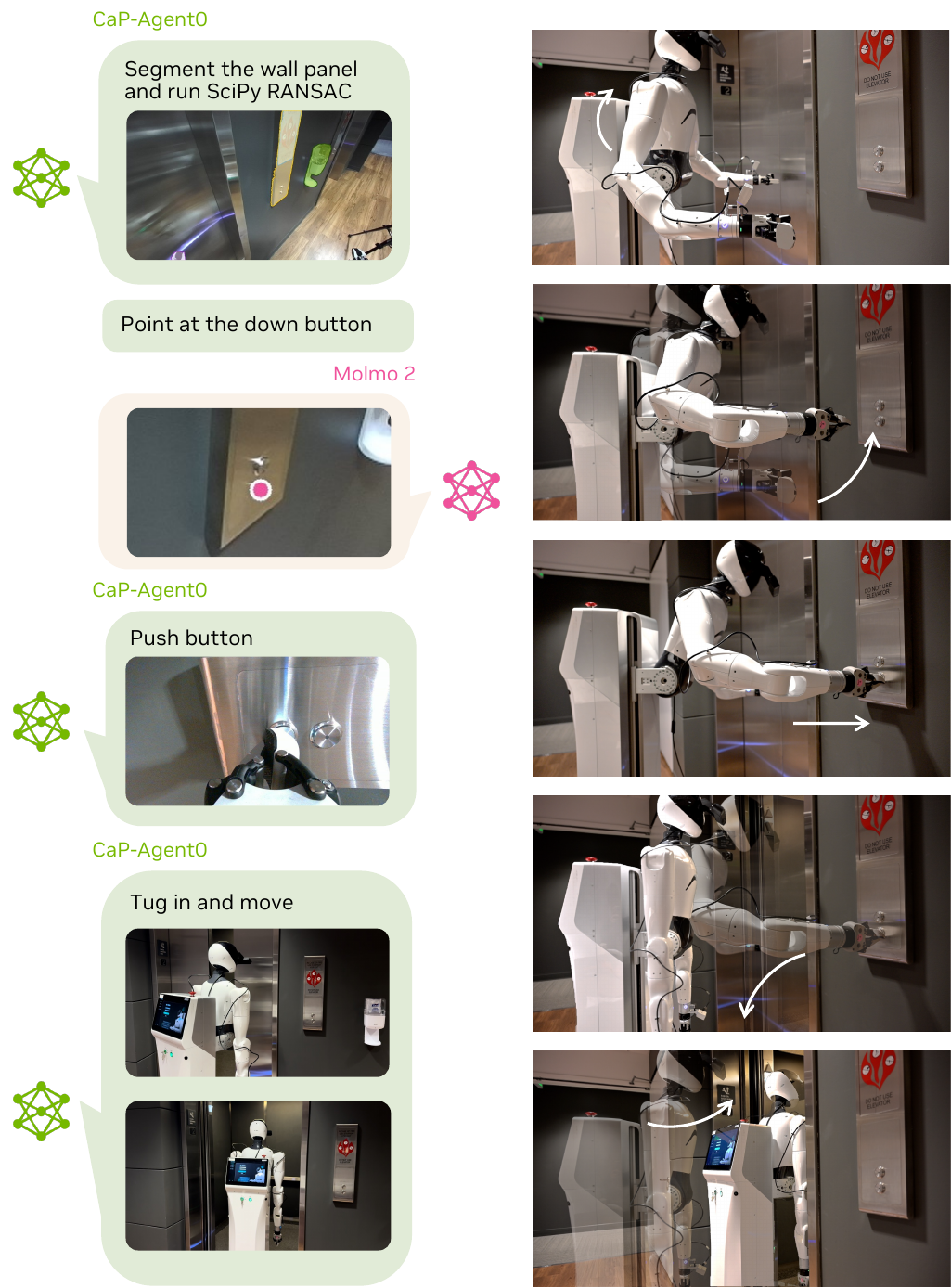}
    \caption{Example of robot pushing the elevator button.}
    \label{fig:elevator}
\end{figure}

This task demonstrates the versatility of the code-as-policy interface, where \sysname{} can use tools from arbituary Python packages to assist with the task.
We ask the robot to ``take the elevator downstairs'', which suggests locating and pressing the button, and moving into the elevator. Since the robot is positioned at an angle with respect to the wall, it is not immediately obvious in which direction the button should be pushed. \sysname{} is able to invoke SciPy RANSAC algorithm on the segmented wall pointcloud, and compute the surface normal direction, shown in~\cref{fig:elevator}.

\clearpage
\newpage
\section{Full Benchmark Table}
We present the full results from \benchname{} in \cref{fig:fullbenchtable}. From top to bottom are code compilation success rate, average dense reward, and average task success rate.
\fullbenchtable{h}

\newpage
\section{Additional Takeaways}

\textbf{In-Context Examples of API Usage improves performance}. In the docstring of each API used in S4, we provided a brief description of the API, the expected datatype, and shape for each input and output. However, due to low code compilation successes from open source models in the initial experiments, we hypothesize that additional prompting may be necessary to increase model performance. Therefore, in addition to the components above, we added API usage examples in the docstring. 
\iclapiusage{H}
We present the results in \cref{fig:abstraction_increase} and \cref{fig:codecompilation}, and a per-model breakdown in \cref{fig:iclapiusage}. We observe that almost all models benefited from adding additional context of examples of how to use the low-level APIs. 
\newpage

\section{Qualitative Analysis of CaP-RL Post-Training Effects}\label{app:reinforcement_learning}
In this section, we analyze the qualitative behavioral shifts induced by on-policy reinforcement learning (GRPO) on the \texttt{Qwen2.5-Coder-7B-Instruct} base model. We focus on the \textit{Stack Red Cube on Green Cube} task to illustrate how RL mitigates fundamental instruction-following failures and induces robust geometric reasoning. Methodologically, CaP-RL leverages the privileged S1 tier to stabilize training by removing perception noise. Below we show the code block examples before and after CaP-RL Post-Training. 

\textbf{Before CaP-RL Post-Training Model Behavior.} Qwen2.5-Coder-7B-Instruct base model code generation example on the task \textit{Stack Red Cube on Green Cube} prior to on-policy reinforcement learning with CaP-RL:
\begin{lstlisting}
import numpy as np

# Get the pose of the green cube
green_cube_position, green_cube_quaternion_wxyz, _ = get_object_pose("green cube")

# Get the pose of the red cube
red_cube_position, red_cube_quaternion_wxyz, _ = get_object_pose("red cube")

# Calculate the placement position for the red cube
placement_position = np.array(green_cube_position) + np.array([0, 0, 0.05])  # 5 cm above the green cube

# Move the robot to the placement position
goto_pose(placement_position, (0, 0, 1, 0), z_approach=0.1)  # Use a slight z_approach for precision

# Open the gripper to release the red cube
open_gripper()
\end{lstlisting}

\textbf{After CaP-RL Post-Training Model Behavior.} Code generation example of Qwen2.5-Coder-7B-Instruct model after on-policy reinforcement learning with CaP-RL on the task \textit{Stack Red Cube on Green Cube} :
\begin{lstlisting}
import numpy as np

# Get the positions and extents of the red and green cubes
red_cube_pos, _, red_cube_extent = get_object_pose("red cube", return_bbox_extent=True)
green_cube_pos, _, green_cube_extent = get_object_pose("green cube", return_bbox_extent=True)

# Calculate the placement position on top of the green cube
placement_pos = green_cube_pos + np.array([0, 0, red_cube_extent[2] / 2 + green_cube_extent[2] / 2])

# Sample a grasp pose for the red cube
grasp_pos, grasp_quat = sample_grasp_pose("red cube")

# Move to the grasp pose and approach
goto_pose(grasp_pos, grasp_quat, z_approach=0.1)

# Close the gripper to grasp the red cube
close_gripper()

# Move to the placement position
goto_pose(placement_pos, (0, 0, 1, 0), z_approach=0.1)

# Open the gripper to release the red cube
open_gripper()
\end{lstlisting}

\prepostrlreal{h!}

\textbf{Pre-RL Failure Modes: Step Skipping and Hallucinated State.}
Prior to RL post-training, the base model frequently exhibits "step skipping", a failure mode where the agent attempts to satisfy the final goal state without executing the necessary prerequisites. As shown in the pre-RL snippet, the model correctly identifies the target location (\texttt{placement\_position}) but fails to grasp the red cube. It attempts to move the gripper directly to the placement target and open it, seemingly hallucinating that it is already holding the object.

\textbf{Post-RL Improvement: Causal Sequencing and Geometric Generalization.}
After GRPO training, two critical improvements emerge:
\begin{enumerate}
    \item \textbf{Causal Sequencing:} The model correctly synthesizes the full manipulation chain: \textit{Identify $\rightarrow$ Grasp $\rightarrow$ Transport $\rightarrow$ Release}. It learns through environment interaction and reward feedback signals that the causal dependency that an object must be grasped (via \texttt{close\_gripper}) before it can be placed.
    \item \textbf{Dynamic Geometric Reasoning:} Instead of using hard-coded offsets, the RL-trained model utilizes the \texttt{return\_bbox\_extent=True} parameter to dynamically calculate the stacking height based on object dimensions (\texttt{red\_extent[2]/2 + green\_extent[2]/2}). This indicates a shift from memorization to grounded geometric reasoning.
\end{enumerate}

\textbf{Zero-shot Generalization to Non-privileged Physical World Setups.} Although the model is only post-trained on the privileged S1 tier, the resulting policies demonstrate robust zero-shot transfer to the non-privileged S2 tier. Consequently, the model functions effectively in physical real-world setups, as shown in Figure~\ref{fig:RL_real_franka}.
\postrlgeneralization{h!}

\textbf{Zero-Shot Generalization to Task Variants.}
We observe instruction-following and zero-shot generalization to similar task variants after RL post-training, as seen in \ref{fig:postRLgeneral}. The same policy successfully executes variants such as \textit{"put the tennis ball on the green cube"} or tasks involving randomized object colors, as the underlying logic relies on abstract properties (bounding boxes) rather than overfitting to specific entity names or training instance scales. 

\newpage

\section{\sysname{} Case Studies}
\label{app:case_study}
\subsection{Real-World Case Studies}
\subsubsection{Common Sense Physics-Aware Task Decomposition}
\label{app:case_study_physics_reasoning}
\tennisballcasestudy{h!}
We validate \sysname{} on a real-world Franka Panda platform using a set of geometrically heterogeneous objects: a yellow cube, a blue cube, and a tennis ball. The agent was given the open-ended instruction: \textit{"stack these as high as you can."} with no additional prompt context outside of API usage documentation and the multimodal agent initial scene description.

Crucially, the task prompt provides neither a specific stacking sequence nor \textit{any} semantic description of the scene or its objects. While a standard visuomotor policy might attempt to stack objects in detection order, potentially attempting to balance a cube on top of the spherical ball, \sysname{} demonstrated common-sense embodied physical reasoning by deriving the only stable construction order.

This behavior is enabled by the \sysname{} auxiliary VDM multimodal agent that extracts task-relevant image information before code generation begins. Prompted to "describe the initial state of the environment with the goal of the task in mind," the VLM emits the following context for the coding agent's initial turn:

\begin{quote} \textit{"...The cubes appear to have flat surfaces suitable for stacking, while the tennis ball is spherical and would likely need to be placed on top or handled carefully."} \end{quote}

Leveraging this textual grounding, the coding agent explicitly decomposed the problem based on stability constraints. The full code generated by the \sysname{} agent for completing this task is found in Appendix~\ref{app:code_stack_as_high_as_you_can}.

\subsubsection{Implicit Multi-Step Reasoning for Obstructed Goals}
\label{app:case_study_obstructed_stack}
\restackrealcasestudy{h!}

We evaluate \textit{\sysname{}} on a task where the \textit{Franka Panda} must place a blue cube on top of a yellow cube. In the initial state, the cubes form a three-object tower: blue is at the base, supporting yellow, which in turn supports a green cube (Figure~\ref{fig:restack_real}).

The goal prompt \textit{"place the blue cube on top of the yellow cube"} is linguistically underspecified; it contains no mention of the green obstruction nor the fact that the blue target currently supports the yellow destination.

\sysname{} resolves this dependency through its auxiliary multimodal perception layer. Before code generation, the VLM provides the coding agent with a high-level strategic prior: \begin{quote} \textit{"...the blue cube is at the bottom of the stack, underneath the yellow cube, and the yellow cube is underneath a green cube. This implies a complex manipulation task where the robot will likely need to: 1.  Remove the top (green) cube. 2.  Remove the middle (yellow) cube to access the blue one, or move the stack. 3.  Ultimately re-arrange them such that the blue cube ends up on top of the yellow one. The straightforward description is that the target blue cube is currently obstructed and supporting the target destination object (the yellow cube)."} \end{quote}

This behavior demonstrates that programmatic agents, when augmented with a dedicated perception-to-text grounding layer, can derive complex manipulation strategies traditionally requiring extensive task-specific imitation learning data or manual state-machine design. The full executable rollout is provided in Appendix~\ref{app:code_restack_real}.

\subsection{Simulation Case Studies}
\subsubsection{Verification and Retry Behavior Enabled by Multiturn Design}
\label{app:retry_behavior}
This case study comes from the "pick up the red cube" task from Robosuite ported into CaP-Gym. Unlike one-shot code generation frameworks, the multi-turn architecture of \sysname{} allows for emergent agentic resilience. In several simulation trials, we observed instances where a planned grasp was physically executed but failed to successfully lift the object due to physics-engine stochasticity or sub-optimal contact points. Rather than terminating the task upon failure, the agent utilized the subsequent turn's observation to verify that the goal condition (e.g., $z_{object} > z_{table}$) remained unfulfilled. As shown in the generated logic, the agent autonomously synthesized a recovery routine branch. This capacity to reason about execution failures and reset the environment state demonstrates that \sysname{} can function as a closed-loop reactive agent rather than a static script generator.
\begin{lstlisting}
    ...
    
masks = segment_sam3_text_prompt(rgb, "red cube")

is_lifted = False
if not masks:
    print("Could not find red cube to verify state. Assuming failed and moving to home pose.")
    is_lifted = False
else:
    best_mask = max(masks, key=lambda x: x["score"])["mask"]
    
    y_idxs, x_idxs = np.where(best_mask)
    if len(y_idxs) > 0:
        stride = 10
        y_sample = y_idxs[::stride]
        x_sample = x_idxs[::stride]
        
        z_sample = depth[y_sample, x_sample]
        if z_sample.ndim > 1: z_sample = z_sample.flatten()
        
        # Filter invalid depth
        valid = z_sample > 0
        if np.any(valid):
            z_sample = z_sample[valid]
            x_sample = x_sample[valid]
            y_sample = y_sample[valid]
            
            fx = intrinsics[0, 0]
            fy = intrinsics[1, 1]
            cx = intrinsics[0, 2]
            cy = intrinsics[1, 2]
            
            x_c = (x_sample - cx) * z_sample / fx
            y_c = (y_sample - cy) * z_sample / fy
            
            points_c = np.stack([x_c, y_c, z_sample], axis=1)
            points_c_homo = np.hstack([points_c, np.ones((len(points_c), 1))])
            points_w = (cam_pose @ points_c_homo.T).T[:, :3]
            
            mean_z = np.mean(points_w[:, 2])
            print(f"Estimated mean Z of object: {mean_z:.4f}")
            
            # If Z > 0.15, we assume it's lifted (table usually ~0, cube size ~0.05)
            if mean_z > 0.15:
                is_lifted = True
                print("Object is confirmed to be lifted.")
            else:
                print("Object appears to be on the table.")

if is_lifted:
    print("Task completed successfully.")
else:
    print("Retrying grasp sequence...")
    open_gripper()
    
    # Move to home pose to get a better view (avoid occlusion from gripper)
    home_pos = np.array([0.5, 0.0, 0.5]) 
    home_quat = np.array([0.0, 1.0, 0.0, 0.0]) # w,x,y,z (Approx gripper pointing down)
    
    move_to_joints(solve_ik(home_pos, home_quat))
    
    # Re-perceive
    obs = get_observation()
    rgb = obs["robot0_robotview"]["images"]["rgb"]
    depth_im = obs["robot0_robotview"]["images"]["depth"]
    intrinsics = obs["robot0_robotview"]["intrinsics"]
    cam_pose = obs["robot0_robotview"]["pose_mat"]
    
    masks = segment_sam3_text_prompt(rgb, "red cube")
    if not masks:
        print("No red cube found after reset.")
    else:
        best_mask = max(masks, key=lambda x: x["score"])["mask"]
        if depth_im.ndim == 3: depth_map = depth_im[:, :, 0]
        else: depth_map = depth_im
        
        grasp_poses, grasp_scores = plan_grasp(depth_map, intrinsics, best_mask.astype(np.int32))
        
        if len(grasp_poses) == 0:
            print("No grasps found.")
        else:
            best_idx = np.argmax(grasp_scores)
            best_pose_cam = grasp_poses[best_idx]
            
            pose_world = cam_pose @ best_pose_cam
            target_pos = pose_world[:3, 3]
            target_rot = pose_world[:3, :3]
            target_quat = rot_to_quat(target_rot)
            
            # Grasp execution
            # 1. Move to Pre-grasp
            pre_pos = target_pos + np.array([0, 0, 0.1])
            move_to_joints(solve_ik(pre_pos, target_quat))
            
            # 2. Move to Grasp
            move_to_joints(solve_ik(target_pos, target_quat))
            
            # 3. Close
            close_gripper()
            
            # 4. Lift
            lift_pos = target_pos + np.array([0, 0, 0.25])
            move_to_joints(solve_ik(lift_pos, target_quat))
            print("Retry finished.")

    ...
\end{lstlisting}
\subsubsection{Perception Module Cascade Failure Fallbacks}
\label{app:perception_fallbacks}
This case study comes from the "wipe up the brown spill" task from Robosuite ported into CaP-Gym. The model frequently constructs robust perception pipelines that utilize hierarchical fallbacks. It understands that a generic text prompt might fail or be imprecise. In several trials, the model employs a VLM (Molmo) to visually ground the target ("brown spill") into a specific 2D point, and then uses that point to prompt a segmentation model (SAM3) for a precise mask. If this high-precision method fails, it falls back to broader text prompts like "stain" or "brown object".

\begin{lstlisting}
    ...
    
    # 2. Try detection using Molmo (VLM) for better semantic understanding
    print("Attempting to detect spill using Molmo...")
    molmo_res = point_prompt_molmo(rgb, "brown spill")
    # Molmo returns dict[str, tuple], getting the first/only value
    spill_pt = None
    if molmo_res:
        # Check if we got a valid coordinate
        for key, pt in molmo_res.items():
            if pt[0] is not None and pt[1] is not None:
                spill_pt = pt
                print(f"Molmo found spill at: {spill_pt}")
                break
    masks = []
    if spill_pt:
        # Use the point to segment with SAM3
        print(f"Segmenting with SAM3 point prompt at {spill_pt}")
        masks = segment_sam3_point_prompt(rgb, spill_pt)
    else:
        # Fallback to text prompt with different wording if Molmo fails
        print("Molmo failed. Trying SAM3 text prompt 'stain'...")
        masks = segment_sam3_text_prompt(rgb, "stain")
        if not masks:
            print("Fallback to 'brown liquid'...")
            masks = segment_sam3_text_prompt(rgb, "brown liquid")
    if not masks:
        print("No masks found after all attempts. Cannot proceed.")
        return
    ...
\end{lstlisting}

\subsubsection{Geometric Reasoning for Grasp Selection}
\label{app:geometric_reasoning} This case study illustrates how \sysname{} enhances sub-symbolic planners by injecting context-aware strategy. While modules like \textit{Contact-GraspNet} infer many 6D grasp poses along with confidence scores, the learned scoring lacks awareness of the broader task setting. For this \textit{tabletop} scenario, the agent correctly identified that a top-down approach strategy is inherently more reliable than side-approaches, regardless of their raw grasp scores.

To enforce this strategy, the agent autonomously synthesized a geometric constraint wrapper. By computing the alignment between grasp approach vectors and the world vertical, the generated code filtered out high-scoring but suboptimal side-grasps, explicitly selecting a vertical candidate to maximize stability. The agent also synthesized an autonomous fallback branch: if this strict strategy yielded no candidates, the logic reverted to the global maximum score to prevent execution stagnation. This demonstrates the agent's capacity to wrap generic tools in context-specific logic, bridging the gap between raw geometric perception and high-level task semantics.

Crucially, the automatically synthesized skill library (Appendix \ref{app:skill_lib}) contains an analogous grasp filter, serving as a primary example of the agent's capacity to synthesize task-agnostic geometric logic into its persistent codebase.
\begin{lstlisting}
    ...
    
    # 2. Grasp Planning
    print("Planning grasp...")
    grasp_poses, grasp_scores = plan_grasp(depth, intrinsics, seg_map)
    if len(grasp_poses) == 0:
        print("Error: No grasp poses found.")
        return
        
    # 3. Filter and Selection
    # Transform all grasps to world frame to check orientation
    best_score = -1.0
    best_grasp_world = None
    # We iterate through high-scoring grasps
    indices = np.argsort(grasp_scores)[::-1] # Descending order
    print(f"Found {len(indices)} candidates. Evaluating top 10...")
    for idx in indices[:20]:
        grasp_camera = grasp_poses[idx]
        grasp_world = transform_camera_to_world(grasp_camera, camera_pose)
        # Check approach vector (Z-axis of rotation matrix)
        # We want the gripper to approach somewhat downwards (negative Z in world)
        rot_mat = grasp_world[:3, :3]
        approach_vec = rot_mat[:, 2] # The Z-axis
        # Dot product with world down vector (0, 0, -1)
        # If dot product is positive (angle < 90), it's pointing somewhat down.
        # Actually, if approach is (0,0,-1), dot with (0,0,-1) is 1.
        # We want approach_z to be negative.
        if approach_vec[2] < -0.5: # At least 60 degrees downward tilt
            print(f"Candidate {idx}: Score {grasp_scores[idx]:.3f}, Approach Z: {approach_vec[2]:.3f} (Good)")
            best_grasp_world = grasp_world
            break
        else:
            # print(f"Candidate {idx}: Score {grasp_scores[idx]:.3f}, Approach Z: {approach_vec[2]:.3f} (Skipping, too horizontal)")
            pass
    if best_grasp_world is None:
        print("Warning: No steep top-down grasp found. Falling back to highest score.")
        best_grasp_world = transform_camera_to_world(grasp_poses[indices[0]], camera_pose)
    target_pos = best_grasp_world[:3, 3]
    target_rot = best_grasp_world[:3, :3]
    target_quat = rotation_matrix_to_quaternion(target_rot)
    ...
\end{lstlisting}

\subsubsection{Code Generation Ensembling Increases Robustness}
\label{app:model_ensemble_case_study_robust}
In this case study, we analyze the differences in code generation for a VDM agent which uses a single query vs a model ensemble. All generated code is for the cube restack task. The following is a portion of the initial generation from single query.
\begin{lstlisting}
    ...
    import numpy as np
    
    # Step 1: Unstack the green cube from the red cube.
    # Get grasp pose for the green cube (currently on top)
    green_grasp_pos, green_grasp_quat = sample_grasp_pose('green cube')
    
    # Approach and pick up the green cube
    open_gripper()
    goto_pose(green_grasp_pos, green_grasp_quat, z_approach=0.1)
    close_gripper()
    ...
\end{lstlisting}
Notice how there is no fallback for the \texttt{sample\_grasp\_pose()} generation. When the initial code was executed, the outputted grasp was invalid. The model reasoned that the grasp was invalid because it was too close to the origin, so in the subsequent generation, the model retroactively implemented a fallback which only prevents the failure case it just encountered.

\begin{lstlisting}
    ...
    # Based on the logs, the previous execution seems to have proceeded through the logic,
    # but looking at the sampled grasp pose for the red cube: [ 0.04235514  0.08625826 -0.04337643]
    # compared to the get_object_pose for red cube: [ 0.68666898  0.12055038 -0.09239209]
    # The sampled grasp pose for the red cube seems completely wrong (close to robot base origin 0,0,0).
    # This likely caused the robot to fail picking up the red cube or crash/move incorrectly in Step 2.
    ...
    # Get poses
    red_pos, _, _ = get_object_pose('red cube')
    green_pos, _, green_bbox = get_object_pose('green cube', return_bbox_extent=True)
    
    # Sample grasp for red cube
    red_grasp_pos, red_grasp_quat = sample_grasp_pose('red cube')
    
    # Sanity check: if grasp pose is near 0,0,0, fall back to object pose + offset
    if np.linalg.norm(red_grasp_pos) < 0.2:
        # Fallback: use object position directly for grasp center, oriented top-down
        print("Warning: Sampled grasp pose seems invalid. Using object position fallback.")
        red_grasp_pos = red_pos.copy()
        red_grasp_quat = np.array([0, 0, 1, 0]) # Top down for fallback
    ...
\end{lstlisting}

On the other hand, generations from a model ensemble tend to anticipate failure cases and preemptively implement fallbacks as shown through the following initial generation:

\begin{lstlisting}
    ...
    def pick_object(mask, depth, intrinsics, extrinsics, z_lift=0.2):
        """Plans and executes a grasp for the object defined by the mask."""
        # Plan grasp
        grasps, scores = plan_grasp(depth, intrinsics, mask.astype(np.int32))
        best_grasp_world, _ = select_top_down_grasp(grasps, scores, extrinsics, vertical_threshold=0.7)
        
        if best_grasp_world is None:
            # Fallback to centroid grasp if grasp planner fails
            print("Grasp planner failed, falling back to centroid grasp.")
            stats = get_object_stats(mask, depth, intrinsics, extrinsics)
            if stats is None:
                raise RuntimeError("Cannot compute object stats for fallback grasp.")
            
            # Create a default top-down orientation
            R_down = np.array([[1.0, 0.0, 0.0], [0.0, -1.0, 0.0], [0.0, 0.0, -1.0]])
            quat = rotation_matrix_to_quaternion(R_down)
            pos = stats["center"].copy()
            pos[2] = stats["max_z"] - 0.02  # Grasp slightly below top
        else:
            pos, quat = decompose_transform(best_grasp_world)
        
        # Execute sequence
        pre_grasp = pos.copy()
        pre_grasp[2] += 0.12  # 12cm above
        
        open_gripper()
        move_tcp(pre_grasp, quat)
        move_tcp(pos, quat)
    ...
\end{lstlisting}

\section{Low-Level Perception and Control Primitives}
\label{app:apis}
This section provides the complete specifications for the low-level perception and control primitives used in CaP-Bench, including function signatures, interface definitions, and documentation strings. These primitives constitute the exact specifications provided to agents in Tier S3. Tier S4 utilizes an identical primitive set, but with in-context usage examples (the Example: sections within docstrings) stripped. Tiers S1 and S2 operate on high-level abstractions that are constructed by composing these fundamental primitives.

\begin{lstlisting}
from typing import Any
import numpy as np
import open3d as o3d
import viser.transforms as vtf
from PIL import Image
from scipy.spatial.transform import Rotation as SciRotation
from capx.envs.base_env import BaseEnv
from capx.integrations import pyroki_snippets as pks  # type: ignore
from capx.integrations.base_api import ApiBase
from capx.integrations.grasp_graspnet import init_contact_graspnet
from capx.integrations.molmo import init_molmo
from capx.integrations.pyroki import init_pyroki
from capx.integrations.sam3 import init_sam3, init_sam3_point_prompt

class S3(ApiBase):
    """
    Robot perception and control primitives for the S3 CaP-Bench Tier
    """
    def __init__(
        self,
        env: BaseEnv,
        tcp_offset: list[float] | None = [0.0, 0.0, -0.107],
        bimanual: bool = False,
    ) -> None:
        super().__init__(env)
        self._TCP_OFFSET = np.array(tcp_offset, dtype=np.float64)
        self.grasp_net_plan_fn = init_contact_graspnet()  
        self.sam3_seg_fn = init_sam3()
        self.sam3_point_prompt_fn = init_sam3_point_prompt()
        self.molmo_point_fn = init_molmo()

        self.ik_solve_fn = init_pyroki()
        self.trajopt_plan_fn = init_pyroki_trajopt()
        self.cfg = None
        self.bimanual = bimanual

    def functions(self) -> dict[str, Any]:
        fns = {
            "get_observation": self.get_observation,
            "segment_sam3_text_prompt": self.segment_sam3_text_prompt,
            "segment_sam3_point_prompt": self.segment_sam3_point_prompt,
            "point_prompt_molmo": self.point_prompt_molmo,
            "plan_grasp": self.plan_grasp,
            "get_oriented_bounding_box_from_3d_points": self.get_oriented_bounding_box_from_3d_points,
        }
        if self.bimanual:
            fns["solve_ik_arm0"] = self.solve_ik_arm0
            fns["solve_ik_arm1"] = self.solve_ik_arm1
            fns["move_to_joints_both"] = self.move_to_joints_both
            fns["move_to_joints_arm0"] = self.move_to_joints_arm0
            fns["move_to_joints_arm1"] = self.move_to_joints_arm1
            fns["open_gripper_arm0"] = self.open_gripper_arm0
            fns["close_gripper_arm0"] = self.close_gripper_arm0
            fns["open_gripper_arm1"] = self.open_gripper_arm1
            fns["close_gripper_arm1"] = self.close_gripper_arm1
        else:
            fns["solve_ik"] = self.solve_ik
            fns["move_to_joints"] = self.move_to_joints
            fns["open_gripper"] = self.open_gripper
            fns["close_gripper"] = self.close_gripper

        return fns

    def get_observation(self) -> dict[str, Any]:
        """Get the observation of the environment.
        Returns:
            observation:
                A dictionary containing the observation of the environment.
                The dictionary contains the following keys:
                - ["robot0_robotview"]["images"]["rgb"]: Current color camera image as a numpy array of shape (H, W, 3), dtype uint8.
                - ["robot0_robotview"]["images"]["depth"]: Current depth camera image as a numpy array of shape (H, W, 1), dtype float32.
                - ["robot0_robotview"]["intrinsics"]: Camera intrinsic matrix as a numpy array of shape (3, 3), dtype float64.
                - ["robot0_robotview"]["pose_mat"]: Camera extrinsic matrix as a numpy array of shape (4, 4), dtype float64.

        """
        return self._env.get_observation()

    # --------------------------------------------------------------------- #
    # Vision models: Sam3 segmentation
    # --------------------------------------------------------------------- #

    def segment_sam3_point_prompt(
        self,
        rgb: np.ndarray,
        point_coords: tuple[float, float],
    ) -> list[dict[str, Any]]:
        """Run SAM3 segmentation on an RGB image, optionally conditioned on an image coordinate point prompt.

        Args:
            rgb:
                RGB image array of shape (H, W, 3), dtype uint8.
            point_coords:
                (x, y) pixel coordinates of the point prompt.

        Returns:
            masks:
                A list of dictionaries. Each dict may contain:

                  - "mask":  np.ndarray of shape (H, W), dtype bool or uint8,
                              where True/1 means the pixel belongs to the instance.
                  - "score": float confidence score.

        Example:
            >>> rgb = obs["robot0_robotview"]["images"]["rgb"]
            >>> masks = segment_sam3_point_prompt(rgb, (100, 100))
        """
        return self.sam3_point_prompt_fn(Image.fromarray(rgb), point_coords)

    def segment_sam3_text_prompt(
        self,
        rgb: np.ndarray,
        text_prompt: str,
    ) -> list[dict[str, Any]]:
        """Run SAM3 segmentation on an RGB image conditioned on a text prompt.

        Args:
            rgb:
                RGB image array of shape (H, W, 3), dtype uint8.
            text_prompt:
                Text prompt for SAM3 segmentation.

        Returns:
            masks:
                A list of dictionaries. Each dict may contain:

                  - "mask":  np.ndarray of shape (H, W), dtype bool or uint8,
                              where True/1 means the pixel belongs to the instance.
                  - "box": list [x1, y1, x2, y2] in pixel coordinates.
                  - "score": float confidence score.

        Example:
            >>> rgb = obs["robot0_robotview"]["images"]["rgb"]
            >>> masks = segment_sam3(rgb, text_prompt="red mug")
        """
        return self.sam3_seg_fn(rgb, text_prompt=text_prompt)

    # --------------------------------------------------------------------- #
    # Molmo point prompt
    # --------------------------------------------------------------------- #
    def point_prompt_molmo(
        self,
        image: np.ndarray,
        text_prompt: str,
    ) -> dict[str, tuple[int | None, int | None]]:
        """Use Molmo to point to a coordinate in the image based on a text prompt.

        Args:
            image: np.ndarray: The RGB image to process. Shape: (H, W, 3), dtype uint8.
            text_prompt: str: The text prompt to point to.

        Returns:
            dict[str, tuple[int | None, int | None]]: Pixel coordinates for each
            object query; (None, None) if parsing failed.
        """
        return self.molmo_point_fn(Image.fromarray(image), objects=[text_prompt])

    def get_oriented_bounding_box_from_3d_points(self, points: np.ndarray) -> dict[str, Any]:
        """Get the oriented bounding box from 3D points.

        Args:
            points: np.ndarray: The 3D points to get the oriented bounding box from.
                Shape: (N, 3), dtype float64.

        Returns:
            dict[str, Any]: The oriented bounding box. The dictionary contains the following keys:
                - "center": np.ndarray: The center of the oriented bounding box in point cloud frame.
                - "extent": np.ndarray: The extent of the oriented bounding box.
                - "R": np.ndarray: The rotation matrix of the oriented bounding box in point cloud frame.
        """
        o3d_points = o3d.geometry.PointCloud()
        o3d_points.points = o3d.utility.Vector3dVector(points)
        o3d_points, ind = o3d_points.remove_statistical_outlier(nb_neighbors=20, std_ratio=2.0)
        obb = o3d_points.get_oriented_bounding_box()
        return {
            "center": obb.center,
            "extent": obb.extent,
            "R": obb.R,
        }

    # --------------------------------------------------------------------- #
    # Grasp planner (Contact-GraspNet)
    # --------------------------------------------------------------------- #
    def plan_grasp(
        self,
        depth: np.ndarray,
        intrinsics: np.ndarray,
        segmentation: np.ndarray,
    ) -> tuple[np.ndarray, np.ndarray]:
        """Plan grasp candidates using Contact-GraspNet for a single instance.

        This is a thin wrapper around the Contact-GraspNet planner. It does not
        apply any camera/world transforms or TCP offsets: the caller is
        responsible for transforming the resulting grasp poses into the desired
        frame and applying TCP offsets if necessary.

        Args:
            depth:
                Depth image in meters.
                Shape: (H, W) or (H, W, 1), dtype float32/float64.
            intrinsics:
                Camera intrinsic matrix.
                Shape: (3, 3), dtype float64.
            segmentation:
                Instance segmentation map where each integer > 0 corresponds to a
                unique object instance ID.
                Shape: (H, W) or (H, W, 1), dtype int32/int64.

        Returns:
            grasp_poses:
                np.ndarray of shape (K, 4, 4), dtype float64.
                Homogeneous transforms for each candidate grasp IN THE CAMERA FRAME.
            grasp_scores:
                np.ndarray of shape (K,), dtype float64.
                Confidence score for each candidate grasp.

        Example:
            >>> cam = obs["robot0_robotview"]
            >>> rgb = cam["images"]["rgb"]
            >>> depth = cam["images"]["depth"][:, :, 0]
            >>> sam3_results = sam3_seg_fn(rgb, text_prompt="red mug")
            >>> best = max(sam3_results, key=lambda d: d["score"])
            >>> mask = best["mask"]
            >>> K = cam["intrinsics"]
            >>> grasp_sample_tf, grasp_scores = plan_grasp(
            ...     depth=depth,
            ...     intrinsics=K,
            ...     segmentation=mask,
            ... )
            >>> best_idx = grasp_scores.argmax()
            >>> best_T = grasp_poses[best_idx]  # (4, 4)
            >>> camera_extrinsics = cam["pose_mat"]
            >>> grasp_sample_world_frame = camera_extrinsics @ best_T
        """
        if depth.ndim == 3 and depth.shape[-1] == 1:
            depth = depth[:, :, 0]
        if segmentation.ndim == 3 and segmentation.shape[-1] == 1:
            segmentation = segmentation[:, :, 0]

        grasp_sample, grasp_scores, _ = self.grasp_net_plan_fn(
            depth,
            intrinsics,
            segmentation,
            1,
            z_range=[0.2, 3.5] if self.is_handover else [0.2, 2.0],
            forward_passes=1 if self.is_handover else 3,
        )
        
        grasp_sample_tf = (
            vtf.SE3.from_matrix(grasp_sample) @ vtf.SE3.from_translation(np.array([0, 0, 0.12]))
        ).as_matrix()

        return grasp_sample_tf, grasp_scores

    # --------------------------------------------------------------------- #
    # IK / motion primitives
    # --------------------------------------------------------------------- #
    def solve_ik(
        self,
        position: np.ndarray,
        quaternion_wxyz: np.ndarray,
    ) -> np.ndarray:
        """Solve inverse kinematics for the panda_hand link.

        Args:
            position:
                Target position in world frame.
                Shape: (3,), dtype float64.
            quaternion_wxyz:
                Target orientation as a unit quaternion in world frame.
                Shape: (4,), [w, x, y, z], dtype float64.

        Returns:
            joints:
                np.ndarray of shape (7,), dtype float64.
                Joint angles for the 7 DoF Franka arm.

        Example:
            >>> target_pos = np.array([0.5, 0.0, 0.3])
            >>> target_quat = np.array([1.0, 0.0, 0.0, 0.0])  # identity, wxyz
            >>> joints = solve_ik(target_pos, target_quat)
            >>> move_to_joints(joints)
        """
        pos = np.asarray(position, dtype=np.float64).reshape(3)
        quat_wxyz = np.asarray(quaternion_wxyz, dtype=np.float64).reshape(4)
        quat_xyzw = np.array(
            [quat_wxyz[1], quat_wxyz[2], quat_wxyz[3], quat_wxyz[0]], dtype=np.float64
        )
        rot = SciRotation.from_quat(quat_xyzw)
        offset_pos = pos + rot.apply(self._TCP_OFFSET)

        prev_cfg = self.cfg

        for i in range(15):  # run w/ multiple iterations when using vel_cost ik solver
            self.cfg = self.ik_solve_fn(
                target_pose_wxyz_xyz=np.concatenate([quat_wxyz, offset_pos]),
                prev_cfg=prev_cfg,
            )
            if prev_cfg is not None:
                if np.allclose(self.cfg, prev_cfg, atol=1e-3):
                    break
                else:
                    prev_cfg = self.cfg
        joints = np.asarray(self.cfg[:-1], dtype=np.float64).reshape(7)
        return joints

    # Single arm control APIs

    def move_to_joints(self, joints: np.ndarray) -> None:
        """Move the robot to a given joint configuration in a blocking manner.

        Args:
            joints:
                Target joint angles for the 7-DoF Franka arm.
                Shape: (7,), dtype float64.

        Returns:
            None

        Example:
            >>> joints = np.array([0.0, -0.5, 0.0, -2.0, 0.0, 1.5, 0.8])
            >>> move_to_joints(joints)
        """
        joints = np.asarray(joints, dtype=np.float64).reshape(7)
        self._env.move_to_joints_blocking(joints)

    def open_gripper(self) -> None:
        """Open gripper fully.

        Args:
            None
        """
        self._env._set_gripper(1.0)
        for _ in range(30):
            self._env._step_once()

    def close_gripper(self) -> None:
        """Close gripper fully.

        Args:
            None
        """
        self._env._set_gripper(0.0)
        for _ in range(30):
            self._env._step_once()

        
    # Dual arm control APIs
    def move_to_joints_both(self, joints0: np.ndarray, joints1: np.ndarray) -> None:
        """Move the arms 0 and 1 to a given joint configuration in a blocking manner simultaneously.

        Args:
            joints0:
                Target joint angles for the 7-DoF Franka arm 0.
                Shape: (7,), dtype float64.
            joints1:
                Target joint angles for the 7-DoF Franka arm 1.
                Shape: (7,), dtype float64.
        """
        self._env.move_to_joints_blocking_both(joints0, joints1)

    def move_to_joints_arm0(self, joints: np.ndarray) -> None:
        """Move the robot arm 0 to a given joint configuration in a blocking manner.

        Args:
            joints:
                Target joint angles for the 7-DoF Franka arm 0.
                Shape: (7,), dtype float64.
        """
        joints = np.asarray(joints, dtype=np.float64).reshape(7)
        self._env.move_to_joints_blocking(joints)
        
    def move_to_joints_arm1(self, joints: np.ndarray) -> None:
        """Move the robot arm 1 to a given joint configuration in a blocking manner.

        Args:
            joints:
                Target joint angles for the 7-DoF Franka arm 1.
                Shape: (7,), dtype float64.
        """
        joints = np.asarray(joints, dtype=np.float64).reshape(7)
        self._env.move_to_joints_blocking_arm1(joints)
        
    def open_gripper_arm0(self) -> None:
        """Open gripper fully for Arm 0 (robot0).
        Args:
            None
        Returns:
            None
        """
        self._env._set_gripper(1.0)
        for _ in range(30):
            self._env._step_once()

    def close_gripper_arm0(self) -> None:
        """Close gripper fully for Arm 0 (robot0).
        Args:
            None
        Returns:
            None
        """
        self._env._set_gripper(0.0)
        for _ in range(30):
            self._env._step_once()
    def open_gripper_arm1(self) -> None:
        """Open gripper fully for Arm 1 (robot1).
        Args:
            None
        Returns:
            None
        """
        self._env._set_gripper_arm1(1.0)
        for _ in range(30):
            self._env._step_once()

    def close_gripper_arm1(self) -> None:
        """Close gripper fully for Arm 1 (robot1).
        Args:
            None
        Returns:
            None
        """
        self._env._set_gripper_arm1(0.0)
        for _ in range(30):
            self._env._step_once()

    def solve_ik_arm0(self, position: np.ndarray, quaternion_wxyz: np.ndarray) -> np.ndarray:
        """Solve inverse kinematics for the panda_hand link for Arm 0 (robot0)."""
        pos = np.asarray(position, dtype=np.float64).reshape(3)
        quat_wxyz = np.asarray(quaternion_wxyz, dtype=np.float64).reshape(4)
        quat_xyzw = np.array(
            [quat_wxyz[1], quat_wxyz[2], quat_wxyz[3], quat_wxyz[0]], dtype=np.float64
        )
        rot = SciRotation.from_quat(quat_xyzw)
        offset_pos = pos + rot.apply(self._TCP_OFFSET)

        prev_cfg = self.cfg
        for i in range(15):
            self.cfg = self.ik_solve_fn(
                target_pose_wxyz_xyz=np.concatenate([quat_wxyz, offset_pos]),
                prev_cfg=prev_cfg,
            )
            if prev_cfg is not None:
                if np.allclose(self.cfg, prev_cfg, atol=1e-3):
                    break
                else:
                    prev_cfg = self.cfg

        joints = np.asarray(self.cfg[:-1], dtype=np.float64).reshape(7)
        return joints

    def solve_ik_arm1(self, position: np.ndarray, quaternion_wxyz: np.ndarray) -> np.ndarray:
        """Solve inverse kinematics for the panda_hand link for Arm 1 (robot1)."""
        if not hasattr(self._env, "move_to_joints_blocking_arm1"):
            raise RuntimeError("Environment does not support Arm 1 control")

        if not hasattr(self._env, "base_link_wxyz_xyz_0") or not hasattr(
            self._env, "base_link_wxyz_xyz_1"
        ):
            raise RuntimeError("Environment does not provide base transforms.")

        pose_arm0_base = vtf.SE3.from_rotation_and_translation(
            rotation=vtf.SO3(wxyz=quaternion_wxyz),
            translation=position,
        )
        base0_transform = vtf.SE3(wxyz_xyz=self._env.base_link_wxyz_xyz_0)
        pose_world = base0_transform @ pose_arm0_base

        base1_transform = vtf.SE3(wxyz_xyz=self._env.base_link_wxyz_xyz_1)
        base1_transform_inv = base1_transform.inverse()
        pose_arm1_base = base1_transform_inv @ pose_world

        pos = np.asarray(pose_arm1_base.translation(), dtype=np.float64).reshape(3)
        quat_wxyz = np.asarray(pose_arm1_base.rotation().wxyz, dtype=np.float64).reshape(4)
        quat_xyzw = np.array(
            [quat_wxyz[1], quat_wxyz[2], quat_wxyz[3], quat_wxyz[0]], dtype=np.float64
        )
        rot = SciRotation.from_quat(quat_xyzw)
        offset_pos = pos + rot.apply(self._TCP_OFFSET)

        prev_cfg = self.cfg
        for i in range(15):
            self.cfg = self.ik_solve_fn(
                target_pose_wxyz_xyz=np.concatenate([quat_wxyz, offset_pos]),
                prev_cfg=prev_cfg,
            )
            if prev_cfg is not None:
                if np.allclose(self.cfg, prev_cfg, atol=1e-3):
                    break
                else:
                    prev_cfg = self.cfg
        joints = np.asarray(self.cfg[:-1], dtype=np.float64).reshape(7)
        return joints

\end{lstlisting}

\section{\sysname{} Details}
\subsection{Synthesized Task-Agnostic Skill Library}
\label{app:skill_lib}

In this section, we list all nine agent-synthesized function definitions, interfaces, documentation strings, and low-level implementations added to the \sysname{} skill library.

\begin{lstlisting}
    def rotation_matrix_to_quaternion(R: np.ndarray) -> np.ndarray:
        """
        Convert a 3x3 rotation matrix to a unit quaternion [w, x, y, z].

        Implements the robust Sheppard's method (checking trace and diagonal elements)
        to avoid numerical instability when the trace is close to zero.
        Args:
            R: (3, 3) rotation matrix.
        Returns:
            np.array: [w, x, y, z] unit quaternion.
        """
        tr = np.trace(R)
        if tr > 0:
            S = np.sqrt(tr + 1.0) * 2
            w = 0.25 * S
            x = (R[2, 1] - R[1, 2]) / S
            y = (R[0, 2] - R[2, 0]) / S
            z = (R[1, 0] - R[0, 1]) / S
        elif (R[0, 0] > R[1, 1]) and (R[0, 0] > R[2, 2]):
            S = np.sqrt(1.0 + R[0, 0] - R[1, 1] - R[2, 2]) * 2
            w = (R[2, 1] - R[1, 2]) / S
            x = 0.25 * S
            y = (R[0, 1] + R[1, 0]) / S
            z = (R[0, 2] + R[2, 0]) / S
        elif R[1, 1] > R[2, 2]:
            S = np.sqrt(1.0 + R[1, 1] - R[0, 0] - R[2, 2]) * 2
            w = (R[0, 2] - R[2, 0]) / S
            x = (R[0, 1] + R[1, 0]) / S
            y = 0.25 * S
            z = (R[1, 2] + R[2, 1]) / S
        else:
            S = np.sqrt(1.0 + R[2, 2] - R[0, 0] - R[1, 1]) * 2
            w = (R[1, 0] - R[0, 1]) / S
            x = (R[0, 2] + R[2, 0]) / S
            y = (R[1, 2] + R[2, 1]) / S
            z = 0.25 * S
        return np.array([w, x, y, z])

    def decompose_transform(T: np.ndarray) -> tuple[np.ndarray, np.ndarray]:
        """
        Decompose a 4x4 homogeneous transformation matrix into position and quaternion.
        Args:
            T: (4, 4) homogeneous transformation matrix.
        Returns:
            tuple:
                - position: (3,) np.array
                - quaternion: (4,) np.array [w, x, y, z]
        """
        position = T[:3, 3]
        R = T[:3, :3]
        quat = rotation_matrix_to_quaternion(R)
        return position, quat

    def depth_to_point_cloud(depth_img: np.ndarray, intrinsics: np.ndarray) -> np.ndarray:
        """
        Convert a depth image to a 3D point cloud in the Camera Frame.
        Args:
            depth_img: (H, W) depth map in meters.
            intrinsics: (3, 3) camera intrinsic matrix.
        Returns:
            np.array: (H, W, 3) image of 3D coordinates.
        """
        if depth_img.ndim == 3:
            depth_img = depth_img[:, :, 0]

        h, w = depth_img.shape
        fx = intrinsics[0, 0]
        fy = intrinsics[1, 1]
        cx = intrinsics[0, 2]
        cy = intrinsics[1, 2]

        # Vectorized grid generation
        y_grid, x_grid = np.mgrid[0:h, 0:w]

        z = depth_img
        x = (x_grid - cx) * z / fx
        y = (y_grid - cy) * z / fy

        return np.dstack((x, y, z))

    def mask_to_world_points(
        mask: np.ndarray, depth: np.ndarray, intrinsics: np.ndarray, extrinsics: np.ndarray
    ) -> np.ndarray:
        """
        Convert specific pixels defined by a binary mask into 3D points in the World Frame.
        Args:
            mask: (H, W) binary mask (0 or 1).
            depth: (H, W) depth map.
            intrinsics: (3, 3) camera intrinsics.
            extrinsics: (4, 4) camera-to-world pose matrix.
        Returns:
            np.array: (N, 3) array of valid 3D points in world coordinates.
        """
        # Get pixel coordinates
        ys, xs = np.where(mask > 0)
        if len(ys) == 0:
            return np.empty((0, 3))

        z_vals = depth[ys, xs]

        # Filter invalid depth
        valid = z_vals > 0
        ys = ys[valid]
        xs = xs[valid]
        z = z_vals[valid]

        fx = intrinsics[0, 0]
        fy = intrinsics[1, 1]
        cx = intrinsics[0, 2]
        cy = intrinsics[1, 2]

        # Deproject to Camera Frame
        x_cam = (xs - cx) * z / fx
        y_cam = (ys - cy) * z / fy

        # Stack to (N, 3)
        points_cam = np.stack([x_cam, y_cam, z], axis=-1)

        # Transform to World Frame
        # Create homogeneous coordinates (N, 4)
        points_cam_hom = np.hstack([points_cam, np.ones((len(points_cam), 1))])
        points_world_hom = (extrinsics @ points_cam_hom.T).T

        return points_world_hom[:, :3]

    def pixel_to_world_point(
        u: int, v: int, z: float, intrinsics: np.ndarray, extrinsics: np.ndarray
    ) -> np.ndarray:
        """
        Deproject a single pixel to a 3D world point.
        Args:
            u, v: Pixel coordinates (col, row).
            z: Depth at that pixel.
            intrinsics: (3, 3) matrix.
            extrinsics: (4, 4) matrix.
        Returns:
            np.array: [x, y, z] in world frame.
        """
        fx = intrinsics[0, 0]
        fy = intrinsics[1, 1]
        cx = intrinsics[0, 2]
        cy = intrinsics[1, 2]

        x_cam = (u - cx) * z / fx
        y_cam = (v - cy) * z / fy

        p_cam = np.array([x_cam, y_cam, z, 1.0])
        p_world = extrinsics @ p_cam
        return p_world[:3]

    def transform_points(points: np.ndarray, transform_matrix: np.ndarray) -> np.ndarray:
        """
        Apply a 4x4 homogeneous transform to a set of 3D points.
        Args:
            points: (N, 3) or (H, W, 3) array of points.
            transform_matrix: (4, 4) homogeneous transformation matrix.
        Returns:
            np.array: Transformed points with same shape as input.
        """
        original_shape = points.shape
        # Flatten to (N, 3)
        points_reshaped = points.reshape(-1, 3)

        # Convert to homogeneous (N, 4)
        ones = np.ones((points_reshaped.shape[0], 1))
        points_hom = np.hstack((points_reshaped, ones))

        # Apply transform: (4,4) @ (4,N) -> (4,N) -> Transpose back to (N,4)
        points_transformed = (transform_matrix @ points_hom.T).T

        # Return to (N, 3) and original shape
        return points_transformed[:, :3].reshape(original_shape)

    def interpolate_segment(
        p1: np.ndarray, p2: np.ndarray, step: float = 0.03
    ) -> list[np.ndarray]:
        """
        Generate waypoints along a line segment between two 3D points.
        Args:
            p1: Start point (3,).
            p2: End point (3,).
            step: Distance between waypoints in meters.
        Returns:
            list[np.ndarray]: List of points including p1 and p2.
        """
        dist = np.linalg.norm(p2 - p1)
        if dist < 1e-6:
            return [p1]
        num_points = int(np.ceil(dist / step))
        # Using linspace to ensure we hit the start and end exactly
        return [p1 + (p2 - p1) * t for t in np.linspace(0, 1, num_points + 1)]

    def normalize_vector(v: np.ndarray) -> np.ndarray:
        """
        Normalize a vector to unit length.
        Args:
            v: (3,) vector.
        Returns:
            np.array: (3,) unit vector.
        """
        norm = np.linalg.norm(v)
        if norm < 1e-6:
            return v
        return v / norm

    def select_top_down_grasp(
        grasps: np.ndarray,
        scores: np.ndarray,
        cam_to_world: np.ndarray,
        vertical_threshold: float = 0.8,
    ) -> tuple:
        """
        Selects the best grasp that aligns the gripper vertically (Top-Down).
        Args:
            grasps: (N, 4, 4) Grasp poses in camera frame.
            scores: (N,) Grasp scores.
            cam_to_world: (4, 4) Extrinsics matrix.
            vertical_threshold: Dot product threshold (1.0 is perfectly vertical).
        Returns:
            tuple: (best_grasp_world_matrix, best_score) or (None, -inf)
        """
        best_grasp = None
        best_score = -np.float64("inf")

        # World Z axis (vertical)
        world_z = np.array([0, 0, 1])

        for i, g_camera in enumerate(grasps):
            # Transform grasp to world frame
            g_world = cam_to_world @ g_camera

            # Extract rotation
            R = g_world[:3, :3]

            # Assuming Gripper Z or Y is the approach vector depending on gripper definition.
            # For Franka/Robotiq, the approach vector is usually the Z-axis of the end effector.
            gripper_approach = R[:, 2]

            # Check alignment with negative World Z (pointing down)
            # Dot product should be close to -1 for top-down
            alignment = -np.dot(gripper_approach, world_z)

            if alignment > vertical_threshold:
                if scores[i] > best_score:
                    best_score = scores[i]
                    best_grasp = g_world

        return best_grasp, best_score
\end{lstlisting}

\subsection{Model Ensemble Temperature Details}
\label{app:model_ensemble_details}
In both single and multi-model settings, 9 candidates responses are generated. For single model, we query Gemini-3-Pro 9 times with temperatures 0.1, 0.2, 0.3,..., 0.9. For multi-model, we query Gemini-3-Pro, Claude-Opus-4.5, and GPT-5.2 3 times each with temperatures 0.1, 0.5, and 0.9.

\subsection{Model Ensemble Prompt}
\label{app:model_ensemble_prompt}
This section contains the prompts the coding agent uses to generate the final solution from the candidates for both the initial generation and subsequent multi-turn attempts. Text inside curly braces "\{\}" represent fstring placeholders.

User prompt for both initial and subsequent generations:
\begin{lstlisting}[
    language={},
    basicstyle=\ttfamily\small,
    breaklines=true,
    breakatwhitespace=true,
    columns=fullflexible,
]
    Synthesize the best solution.

    <original_task_description>
    {original prompt for candidates}
    </original_task_description>

    <candidate_solutions>
    {candidate generations}
    </candidate_solutions>
\end{lstlisting}

System prompt for initial generation:
\begin{lstlisting}[
    language={},
    basicstyle=\ttfamily\small,
    breaklines=true,
    breakatwhitespace=true,
    columns=fullflexible,
]
    You are synthesizing {# of candidate generations} candidate Python solutions into one optimal program.
    
    SYNTHESIS RULES:
    1. Analyze critically and assume no candidate is fully correct
    2. Prefer explicit checks over assumptions
    3. Combine the best ideas from multiple candidates when appropriate
    4. If candidates disagree fundamentally, choose the more robust approach

    OUTPUT FORMAT (strict):
    You may include reasoning before the fenced code block.
    Output ONLY ONE fenced code block (```python...```) containing the complete final solution.
    Do NOT include any other code blocks or code snippets outside this single block.
\end{lstlisting}

System prompt for subsequent generations:
\begin{lstlisting}[
    language={},
    basicstyle=\ttfamily\small,
    breaklines=true,
    breakatwhitespace=true,
    columns=fullflexible
]
    You are synthesizing {# of generations} candidate responses for a multi-turn robot control task.

    DECISION ANALYSIS:
    - {regenerate_count} candidates voted REGENERATE
    - {finish_count} candidates voted FINISH

    SYNTHESIS RULES:
    1. Analyze critically and assume no candidate is fully correct
    2. Prefer explicit checks over assumptions
    3. Combine the best ideas from multiple candidates when appropriate
    4. If candidates disagree fundamentally, choose the more robust approach
    5. Combine best code ideas from REGENERATE candidates

    OUTPUT FORMAT (strict):
    - You may include brief reasoning first
    - Then output "REGENERATE" on its own line followed by exactly ONE fenced code block, OR output "FINISH" on its own line
\end{lstlisting}

\subsection{Multi-turn prompt incentivizing debugging}
We noticed that one failure case was inaccurate verification of task completion. Therefore, we experimented with a modified multi-turn prompt which incentivizes verification and debugging, however this did not empirically improve success rate.

\begin{table}[h]
\centering
\small
\begin{tabular}{l*{8}{c}}
\hline
\textbf{} & \textbf{Cube} & \textbf{Cube} & \textbf{Spill} & \textbf{Peg} & \textbf{Cube} & \textbf{Two Arm} & \textbf{Two Arm} & \textbf{Avg.} \\
 & \textbf{Lift} & \textbf{Stack} & \textbf{Wipe} & \textbf{Insert} & \textbf{Restack} & \textbf{Lift} & \textbf{Handover} & \\
\hline
3M & 97 & 98 & 100 & 0 & 89 & 74 & 20 & 68.29 \\
3M + debug & 94 & 100 & 98 & 0 & 88 & 66 & 12 & 65.43 \\
\hline
\end{tabular}
\caption{Task completions between 3M and 3M + debug.}
\label{tab:performance}
\end{table}

Modified multi-turn prompt:
\begin{lstlisting}[
    language={},
    basicstyle=\ttfamily\small,
    breaklines=true,
    breakatwhitespace=true,
    columns=fullflexible
]
    You are acting as experienced debugger and task completion verifier. You will make no assumptions without explicit evidence.
      Your task is to determine whether the program has actually completed the intended task, and to fix all bugs if any exist.
      You can treat the observed differences between the current and previous state of the environment that are provided to you as a source of evidence, but they are not guaranteed to be accurate.

      The following code blocks have been executed so far:
      ```python
      {executed_code}
      ```
      The current console stdout from the most recent code execution is:
      ```
      {console_stdout}
      ```
      The current console stderr from the most recent code execution is:
      ```
      {console_stderr}
      ```

      Do NOT trust printed outputs or logs blindly.

      Proceed as follows to verify task completion:
      1. Line-by-line inspect the executed code
      2. Cross-check the code's behavior against the console stdout and stderr
      3. Identify explicit, concrete evidence that each required step of the task was completed
      4. Treat missing evidence, implicit assumptions, or partial signals as failure

      Decision Rules:
      - If the task is verifiably complete, respond with the single word 'FINISH'.
      - If the task is not verifiably complete:
        1. Identify specific bugs, failures, missing steps, or incorrect assumptions.
        2. Explain why each issue prevents task completion, citing evidence from stdout, stderr, or code behavior.
        3. If the same approach has been attempted multiple times without success, you MUST try a fundamentally different strategy.
        4. Brainstorm multiple strategies to fix or verify these issues in the next code generation.
        5. Synthesize your brainstorming into a single improved Python program that addresses all identified issues and verifiably completes the task.
        6. Respond with the single word 'REGENERATE' followed immediately by new Python code in a fenced code block (```python...```)
      
      If you choose to regenerate code, you must include your reasoning process only AS COMMENTS at the start of the code explaining:
      - What strategies/APIs have already been tried and why they were insufficient
      - How your solution effectively addresses the identified issues

      To reiterate, you must only respond with EXACTLY ONE of the following:
      - The word 'FINISH' if you decide to stop generating code
      - The word 'REGENERATE' followed immediately by new Python code in a SINGLE fenced code block (```python...```).
\end{lstlisting}

\subsection{Model Ensemble Decreases Average Turn Count}
We observed that applying a model ensemble for code generations decreases the average turn count. We noticed that M4 tends to retroactively implement bug fixes and API fallbacks, while the model ensemble preemptively anticipates failures beforehand, resulting in more robust code and lower turn count. See Appendix~\ref{app:model_ensemble_case_study_robust} for a case study on retroactive vs preemptive behavior.

\begin{figure}[h]
\centering
\includegraphics[width=0.5\textwidth]{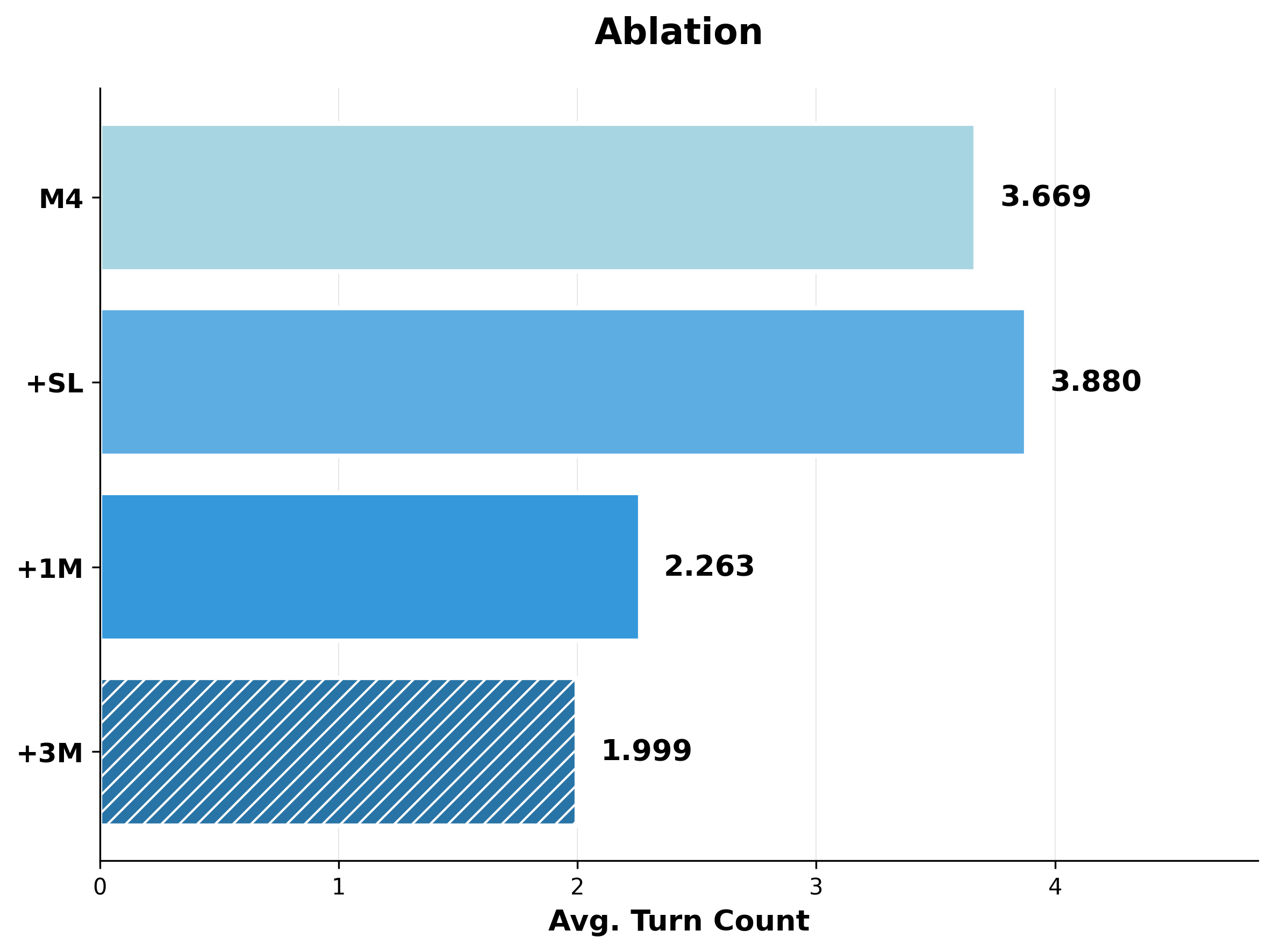}
\caption{CaP-Agent0 decreases average turn count compared to M4}
\label{fig:agent_ablation_turn_count}
\end{figure}

\section{Select Case Studies - Full Generated Code}

\subsection{Stack These as High as You Can}
\label{app:code_stack_as_high_as_you_can}
Generated by Gemini-3-Pro as the coding agent and Gemini-3-Pro as the VDM agent.
\begin{lstlisting}
import numpy as np

def get_best_mask(masks):
    if not masks:
        return None
    return max(masks, key=lambda x: x["score"])

def get_object_metrics(mask, depth, intrinsics, extrinsics):
    """Returns center (x,y,z), top_z, bottom_z for a given object mask."""
    points = mask_to_world_points(mask, depth, intrinsics, extrinsics)
    if points.shape[0] == 0:
        return None, None, None
    
    # Get Oriented Bounding Box
    bbox = get_oriented_bounding_box_from_3d_points(points)
    center = bbox["center"]
    
    # Calculate robust top and bottom Z from points
    zs = points[:, 2]
    if len(zs) == 0:
        return None, None, None
        
    top_z = np.percentile(zs, 98) # Use 98th percentile to filter outliers
    bottom_z = np.percentile(zs, 2) # Use 2nd percentile
    
    return center, top_z, bottom_z

def execute_pose_move(pose_matrix):
    """Moves robot to the specified 4x4 pose matrix."""
    pos, quat = decompose_transform(pose_matrix)
    joints = solve_ik(pos, quat)
    move_to_joints(joints)

def pick_and_place(obj_mask, target_center, target_top_z, depth, intrinsics, cam_pose):
    """
    Picks object defined by obj_mask and places it on target defined by target_center and target_top_z.
    """
    # 1. Analyze Object to be picked
    obj_center, obj_top, obj_bottom = get_object_metrics(obj_mask, depth, intrinsics, cam_pose)
    if obj_center is None:
        print("Could not calculate object metrics.")
        return

    # 2. Plan Grasp
    seg_map = obj_mask.astype(np.int32)
    # plan_grasp returns poses in CAMERA FRAME
    grasp_poses_cam, grasp_scores = plan_grasp(depth, intrinsics, seg_map)
    
    if len(grasp_poses_cam) == 0:
        print("No grasps generated by planner.")
        return

    # Select best top-down grasp. 
    # select_top_down_grasp takes grasps in CAMERA FRAME and returns the best in WORLD FRAME.
    best_grasp_world, best_score = select_top_down_grasp(grasp_poses_cam, grasp_scores, cam_pose, vertical_threshold=0.8)
    
    if best_grasp_world is None:
        print("No strictly vertical grasp found. Trying relaxed threshold.")
        best_grasp_world, best_score = select_top_down_grasp(grasp_poses_cam, grasp_scores, cam_pose, vertical_threshold=0.5)

    if best_grasp_world is None:
        print("No valid grasp found even with relaxed threshold.")
        return

    print(f"Executing grasp with score {best_score}")

    # Calculate offset from TCP to bottom of object
    # TCP Z at grasp - Object Bottom Z
    grasp_z = best_grasp_world[2, 3]
    tcp_to_bottom_offset = grasp_z - obj_bottom

    # 3. Execute Pick
    # Move to Pre-Grasp (Safety Hover)
    pre_grasp = best_grasp_world.copy()
    pre_grasp[2, 3] += 0.10
    execute_pose_move(pre_grasp)

    # Move to Grasp
    execute_pose_move(best_grasp_world)
    close_gripper()

    # Lift Up
    lift_pose = best_grasp_world.copy()
    lift_pose[2, 3] += 0.20
    execute_pose_move(lift_pose)

    # 4. Execute Place
    # Calculate Place Pose
    place_pose = best_grasp_world.copy()
    place_pose[0, 3] = target_center[0]
    place_pose[1, 3] = target_center[1]
    # Target Z for TCP = Target Surface Z + Offset + Padding
    # Adding a small padding (1cm) to ensure we don't collide hard, 
    # but close enough to drop safely.
    place_pose[2, 3] = target_top_z + tcp_to_bottom_offset + 0.015 

    # Move to Pre-Place (Safety Hover)
    pre_place = place_pose.copy()
    pre_place[2, 3] += 0.15
    execute_pose_move(pre_place)

    # Move to Place
    execute_pose_move(place_pose)
    open_gripper()

    # Move to Post-Place (Lift Up)
    execute_pose_move(pre_place)


# --- Main Execution ---

# 1. Reset
open_gripper()

# 2. Perception & Strategy: Stack Blue Cube on Yellow Cube
obs = get_observation()
rgb = obs["robot0_robotview"]["images"]["rgb"]
depth = obs["robot0_robotview"]["images"]["depth"]
intrinsics = obs["robot0_robotview"]["intrinsics"]
cam_pose = obs["robot0_robotview"]["pose_mat"]

# Find Yellow Cube (Base)
masks_yellow = segment_sam3_text_prompt(rgb, "yellow cube")
yellow_data = get_best_mask(masks_yellow)

# Find Blue Cube (Object 1)
masks_blue = segment_sam3_text_prompt(rgb, "blue cube")
blue_data = get_best_mask(masks_blue)

if yellow_data is not None and blue_data is not None:
    # Get base metrics
    yellow_center, yellow_top, _ = get_object_metrics(yellow_data["mask"], depth, intrinsics, cam_pose)
    
    if yellow_center is not None:
        print("Stacking Blue Cube on Yellow Cube...")
        pick_and_place(blue_data["mask"], yellow_center, yellow_top, depth, intrinsics, cam_pose)
    else:
        print("Could not compute metrics for Yellow Cube")
else:
    print("Could not find Blue or Yellow cube")


# 3. Perception & Strategy: Stack Ball on Blue Cube
# Re-observe because the blue cube has moved
obs = get_observation()
rgb = obs["robot0_robotview"]["images"]["rgb"]
depth = obs["robot0_robotview"]["images"]["depth"]
intrinsics = obs["robot0_robotview"]["intrinsics"]
cam_pose = obs["robot0_robotview"]["pose_mat"]

# Find Blue Cube again (New Base)
masks_blue_new = segment_sam3_text_prompt(rgb, "blue cube")
blue_data_new = get_best_mask(masks_blue_new)

# Find Tennis Ball (Object 2)
masks_ball = segment_sam3_text_prompt(rgb, "yellow tennis ball")
ball_data = get_best_mask(masks_ball)

if blue_data_new is not None and ball_data is not None:
    # Get base metrics (blue cube is now the base)
    blue_center_new, blue_top_new, _ = get_object_metrics(blue_data_new["mask"], depth, intrinsics, cam_pose)
    
    if blue_center_new is not None:
        print("Stacking Tennis Ball on Blue Cube...")
        pick_and_place(ball_data["mask"], blue_center_new, blue_top_new, depth, intrinsics, cam_pose)
    else:
        print("Could not compute metrics for Blue Cube (new base)")
else:
    print("Could not find Blue Cube or Tennis Ball for second stack")
\end{lstlisting}

\subsection{Place the Blue Cube on Top of the Yellow Cube}
\label{app:code_restack_real}
Generated by Gemini-3-Pro as the coding agent and Gemini-3-Pro as the VDM agent.
\begin{lstlisting}
import numpy as np

# --- Helper functions ---

def get_object_centroid(obs, text_prompt):
    """
    Finds the 3D centroid of an object specified by a text prompt.
    Returns the centroid (x, y, z) in world frame and the mask.
    """
    rgb = obs["robot0_robotview"]["images"]["rgb"]
    depth = obs["robot0_robotview"]["images"]["depth"]
    intrinsics = obs["robot0_robotview"]["intrinsics"]
    extrinsics = obs["robot0_robotview"]["pose_mat"]
    
    # Segment object
    results = segment_sam3_text_prompt(rgb, text_prompt)
    if not results:
        print(f"Object '{text_prompt}' not found.")
        return None, None
        
    best_result = max(results, key=lambda x: x["score"])
    mask = best_result["mask"]
    
    # Get 3D points
    points = mask_to_world_points(mask, depth, intrinsics, extrinsics)
    if points.shape[0] == 0:
        print(f"No valid depth points for '{text_prompt}'.")
        return None, None
        
    centroid = np.mean(points, axis=0)
    return centroid, mask

def pick_object(obs, text_prompt, z_offset=0.0):
    """
    Standard pick routine:
    1. Segment object
    2. Plan grasp
    3. Execute pick trajectory (pre-grasp -> grasp -> close -> lift)
    """
    rgb = obs["robot0_robotview"]["images"]["rgb"]
    depth = obs["robot0_robotview"]["images"]["depth"]
    intrinsics = obs["robot0_robotview"]["intrinsics"]
    extrinsics = obs["robot0_robotview"]["pose_mat"]

    # 1. Segment
    results = segment_sam3_text_prompt(rgb, text_prompt)
    if not results:
        print(f"Could not find {text_prompt} to pick.")
        return False
    
    # Get mask with highest score
    best_res = max(results, key=lambda x: x["score"])
    mask = best_res["mask"]

    # 2. Plan Grasp
    # Note: depth needs to be (H, W) or (H, W, 1) - API doc says (H, W) usually preferred for planner wrappers, 
    # but the function signature accepts (H, W, 1).
    if depth.ndim == 3 and depth.shape[2] == 1:
        depth_map = depth[:, :, 0]
    else:
        depth_map = depth

    grasp_poses, grasp_scores = plan_grasp(depth_map, intrinsics, mask)
    
    if len(grasp_scores) == 0:
        print(f"No grasps found for {text_prompt}.")
        return False
    
    # Select best top-down grasp
    best_pose, best_score = select_top_down_grasp(grasp_poses, grasp_scores, extrinsics)
    
    if best_pose is None:
        print("No valid top-down grasp found, falling back to highest score.")
        best_idx = np.argmax(grasp_scores)
        best_pose_cam = grasp_poses[best_idx]
        best_pose = extrinsics @ best_pose_cam # Convert to world frame

    # Decompose grasp pose
    grasp_pos, grasp_quat = decompose_transform(best_pose)

    # 3. Execute Pick
    # Pre-grasp (hover 10cm above)
    pre_grasp_pos = grasp_pos + np.array([0, 0, 0.1])
    
    # Move to pre-grasp
    joints = solve_ik(pre_grasp_pos, grasp_quat)
    move_to_joints(joints)
    
    # Open gripper
    open_gripper()
    
    # Move to grasp
    joints = solve_ik(grasp_pos, grasp_quat)
    move_to_joints(joints)
    
    # Close gripper
    close_gripper()
    
    # Lift up (20cm)
    lift_pos = grasp_pos + np.array([0, 0, 0.2])
    joints = solve_ik(lift_pos, grasp_quat)
    move_to_joints(joints)
    
    return True

def place_at_position(position, height_offset=0.05):
    """
    Place currently held object at a specific world position.
    Target orientation is usually top-down (gripper pointing down).
    """
    # Standard top-down orientation (gripper pointing -z)
    # A common quaternion for top-down is pointing down z-axis. 
    # Let's assume the current grasp orientation is maintained or we define a fixed top-down.
    # For simplicity, we often maintain the orientation we picked with, or reset to a known neutral top-down.
    # Here, we will define a fixed top-down orientation [0, 1, 0, 0] (x-axis rotation 180 deg) or similar.
    # However, to be safe, let's just use the current robot configuration's orientation or a hardcoded one.
    # Often [0, 1, 0, 0] is top down for Panda.
    place_quat = np.array([0.0, 1.0, 0.0, 0.0]) 

    target_pos = position + np.array([0, 0, height_offset])
    
    # Move over target (high)
    hover_pos = target_pos + np.array([0, 0, 0.15])
    joints = solve_ik(hover_pos, place_quat)
    move_to_joints(joints)
    
    # Move down to place
    joints = solve_ik(target_pos, place_quat)
    move_to_joints(joints)
    
    # Open gripper
    open_gripper()
    
    # Move back up
    joints = solve_ik(hover_pos, place_quat)
    move_to_joints(joints)

def main():
    obs = get_observation()
    
    # --- Strategy ---
    # The stack is Green (top) -> Yellow (mid) -> Blue (bottom).
    # Goal: Blue on Yellow.
    # 1. Move Green to the side.
    # 2. Move Yellow to another spot (target base).
    # 3. Pick Blue.
    # 4. Place Blue on top of Yellow.

    # Defined drop zones relative to the table center or just hardcoded offsets based on workspace knowledge.
    # Let's dynamically find a spot by looking at the initial stack position.
    
    # Find the stack location (using the green cube at the top)
    green_center, _ = get_object_centroid(obs, "green cube")
    if green_center is None: return

    # Define temporary drop zones relative to the stack
    # Drop zone 1 (for green): 20cm to the left
    drop_zone_green = green_center + np.array([-0.2, 0.0, -green_center[2] + 0.02]) # Z is roughly table height
    # Drop zone 2 (for yellow): 20cm to the right (this will be the final destination for yellow)
    drop_zone_yellow = green_center + np.array([0.2, 0.0, -green_center[2] + 0.02])
    
    # --- Step 1: Remove Green Cube ---
    print("Moving Green Cube...")
    success = pick_object(obs, "green cube")
    if success:
        place_at_position(drop_zone_green, height_offset=0.05) # Place on table
    else:
        print("Failed to move green cube.")
        return

    # Refresh observation
    obs = get_observation()

    # --- Step 2: Move Yellow Cube ---
    # Now Yellow should be exposed.
    print("Moving Yellow Cube...")
    success = pick_object(obs, "yellow cube")
    if success:
        # We place it at drop_zone_yellow. This is where we want to stack the blue one later.
        place_at_position(drop_zone_yellow, height_offset=0.05) 
    else:
        print("Failed to move yellow cube.")
        return

    # Refresh observation to find the new position of the yellow cube and the exposed blue cube
    obs = get_observation()
    
    # Get precise location of yellow cube now that it's moved
    yellow_center, _ = get_object_centroid(obs, "yellow cube")
    if yellow_center is None:
        print("Lost track of yellow cube.")
        return

    # --- Step 3: Pick Blue Cube ---
    print("Picking Blue Cube...")
    success = pick_object(obs, "blue cube")
    if not success:
        print("Failed to pick blue cube.")
        return

    # --- Step 4: Place Blue on Yellow ---
    print("Placing Blue on Yellow...")
    # Target is yellow center, but offset Z by cube height (approx 5cm usually for these cubes)
    # We add a small buffer.
    cube_height_approx = 0.05
    place_at_position(yellow_center, height_offset=cube_height_approx + 0.02)

    # Done, gripper is already opened in place_at_position
    print("Task completed.")

main()
\end{lstlisting}

\subsection{Additional Analysis Plots}
\begin{figure}[h!]
    \centering
    \begin{minipage}{0.48\textwidth}
        \centering
        \includegraphics[width=\textwidth]{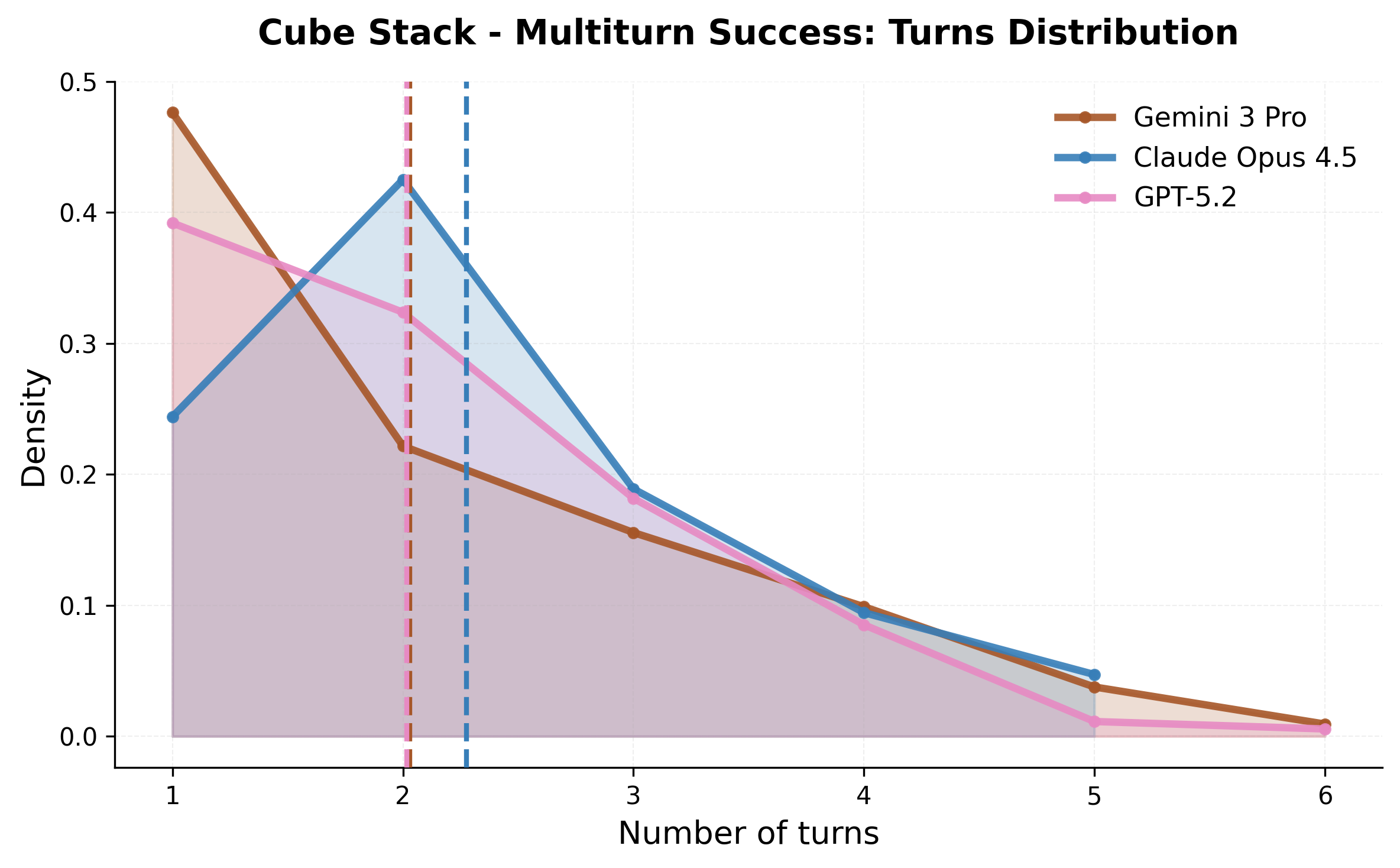}
        \caption{Distribution of multi-turn cube stack successes over turns. Results are averaged across M1, M2, and M3.}
        \label{fig:cube_stack_turns}
    \end{minipage}
    \hfill
    \begin{minipage}{0.48\textwidth}
        \centering
        
        \includegraphics[width=\textwidth]{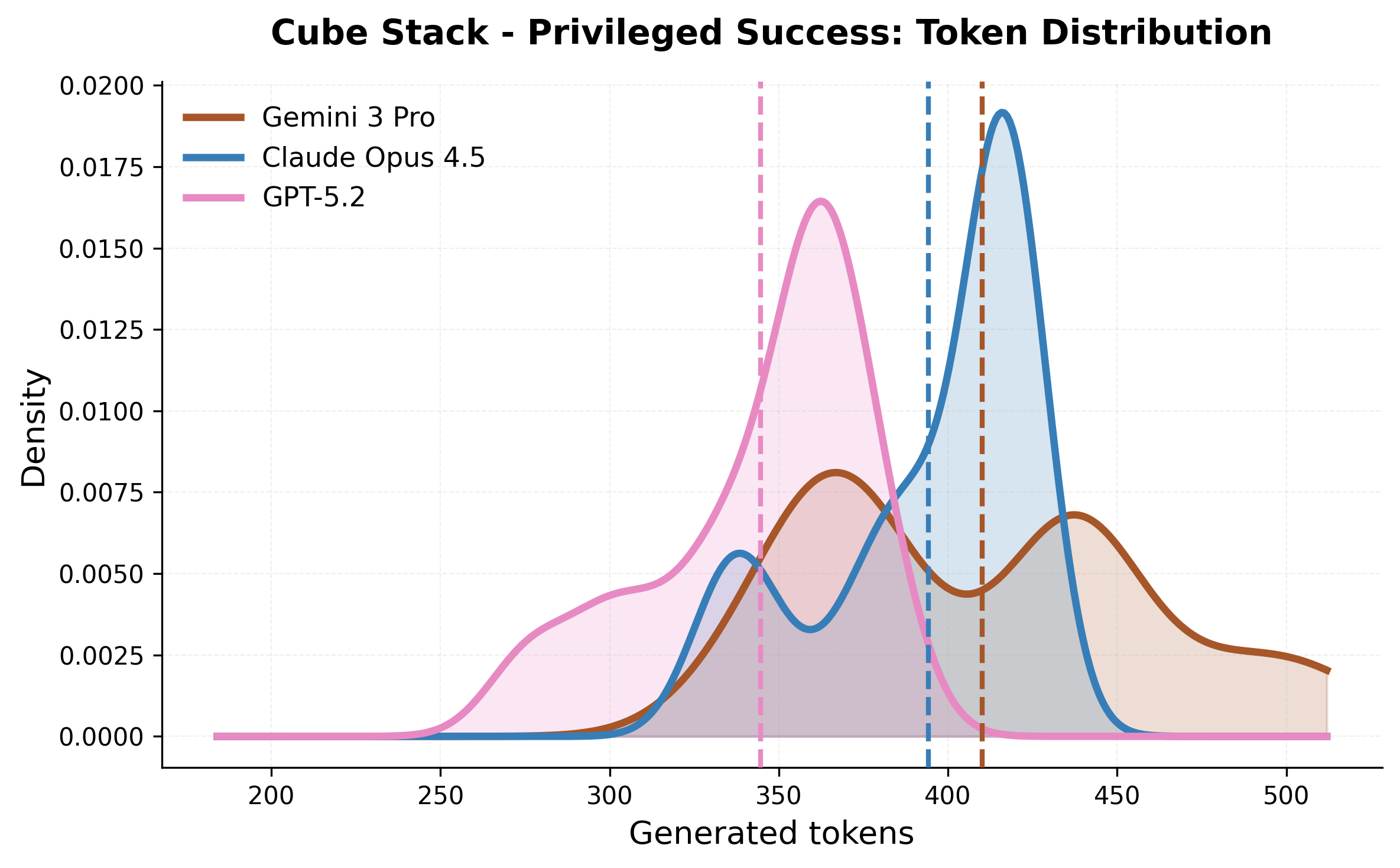}
        \caption{Distribution of cube stack successes (S1) over tokens}
        \label{fig:cube_stack_tokens}
    \end{minipage}
\end{figure}

\begin{figure}[h!]
    \begin{minipage}{0.48\textwidth}
        \centering
        \includegraphics[width=\textwidth]{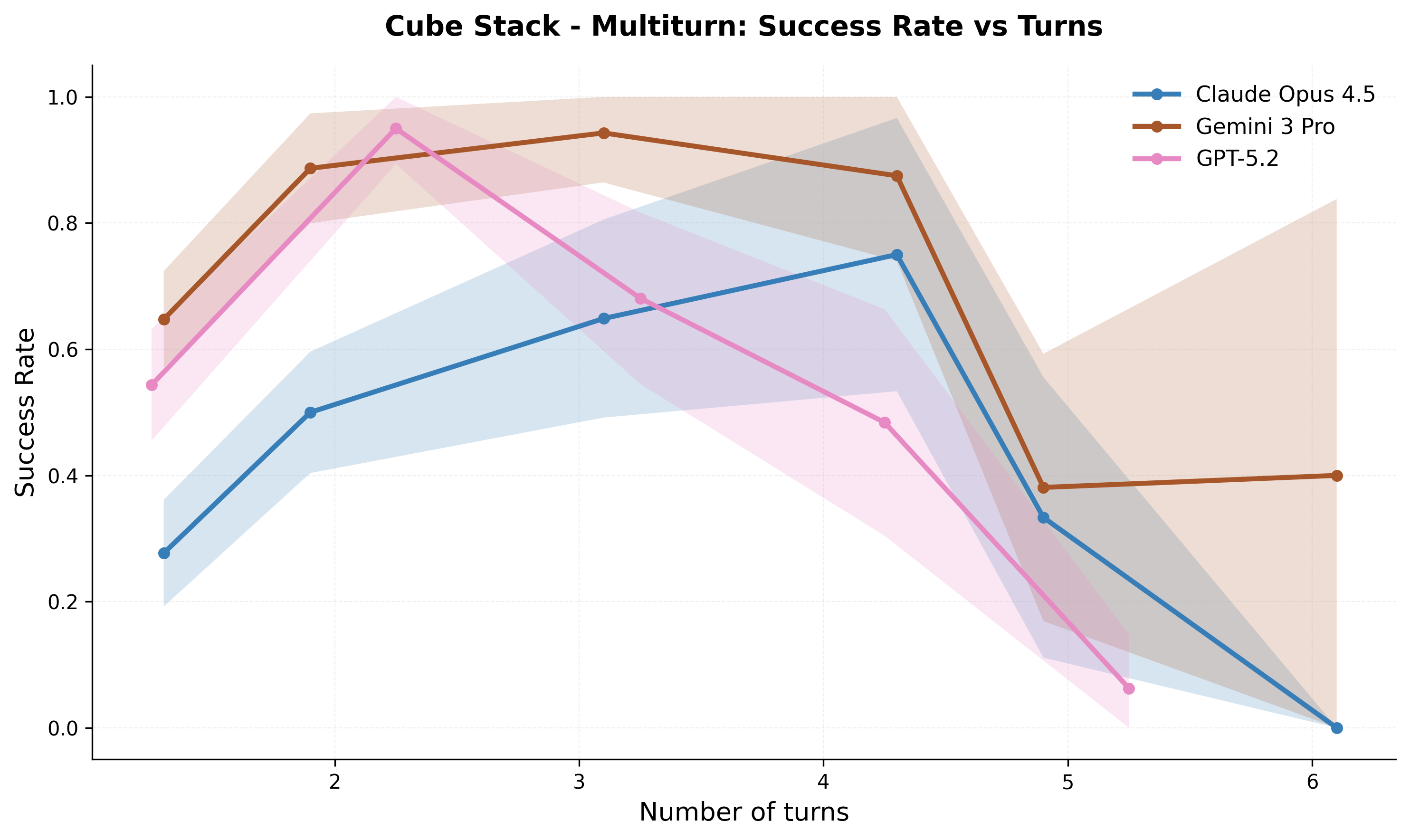}
        \caption{Cube stack multi-turn success rate vs. turns. Trials with too few or too many turns have lower success rate. Results are averaged across M1, M2, and M3.}
        \label{fig:cube_stack_success_turns}
    \end{minipage}
\end{figure}

We perform additional quantitative analysis on the cube stack task to understand the impact of multi-turn on success rate. We focus our analysis on Gemini 3 Pro~\cite{google_deepmind_gemini3pro_modelcard_2025}, GPT 5.2~\cite{openai_gpt5_2_2025}, and Claude Opus 4.5~\cite{anthropic_claude_opus_4_5_2025}, the three strongest performers on \benchname{}. In particular, we plot the probability density function (PDF) of success over turns (\cref{fig:cube_stack_turns}), PDF of success over tokens (\cref{fig:cube_stack_tokens}), multi-turn success rate vs turn for the cube stacking task in \cref{fig:cube_stack_success_turns}. We find that for the cube stacking task, these models would perform the task on the first try, and use the subsequent turns for recovery. Successes for GPT 5.2 and Claude Opus 4.5 seem to have shorter code length than that of Gemini 3 Pro.

\newpage
\section{LIBERO-PRO Evaluation Results}
\label{app:liberoproresults}
We present detailed task-wise performance of OpenVLA, $\pi_0$, $\pi_{0.5}$, and \sysname{} on LIBERO-PRO, which evaluates model generalization under initial position perturbations (Pos) and instruction perturbations (Task). Results are summarized in~\cref{tab:liberoproOBJECTcomparison},~\cref{tab:liberoproGOALcomparison} and~\cref{tab:liberoproSPATIALcomparison}. Each task was executed over 50 trials. In these tasks, \sysname{} struggles with failures in perception, grasp generation, control APIs. For example, queries of "alphabet soup can" to SAM 3 often results in segmentations of the "tomato sauce can" also present in the scene. Also due to the occluded nature of these scenes and the camera not being top down, many grasps generated on desired objects will be at an angle, this, when combined with the lack of collision-aware motion planning often results in other objects in the scene being knocked over during execution, preventing a secure grasp on the desired object.  
\liberoproOBJECTcomparison{}
\liberoproGOALcomparison{}
\liberoproSPATIALcomparison{}

\newpage
\section{Additional Clarifications}
\label{app:reviewer_revisions}
This section consolidates clarifications and added details requested by reviewers during the discussion period. It expands the protocol behind the human-expert baseline, the computational cost of \sysname{}, and the choice of VDM backbone.

\subsection{Human Expert Baseline Protocol}
\label{app:human_expert}
The human-expert curve in \cref{fig:splash_fig} is \emph{not} a single-shot human attempt. The baseline was written by a subset of the paper authors ($N{=}7$), each with $2+$ years of robotics programming experience. For each task and each tier, an author wrote a single Python script using exactly the same API primitives available to the model at that tier, and iterated through normal trial-and-error---reading execution traces, fixing bugs, and updating the script---until the program achieved high reliability. Human-written code uses the same primitives as the model; what differs is the iterative refinement loop performed offline by a human.

\textbf{Resulting upper bound.} The iterated human reference achieves $88.5\%$ average success on single-turn tiers, which we treat as a near-upper-bound for what is achievable when a robotics engineer hand-writes static code with full access to development-time iteration.

\textbf{Human effort budget.} The effort required to reach this near-upper-bound varies sharply with task complexity:
\begin{itemize}
    \item \textbf{Simple pick-and-place tasks} (Cube Lift, Cube Stack): arrived at a working solution in a few hours ($<$1 day), with both low-level primitive and high-level code implementations, hill-climbed with a few rounds of trial-and-error.
    \item \textbf{Contact-rich and bimanual tasks} (Peg Insertion, Two-Arm Handover): identifying the right primitives and the right overall strategy took 2--3 weeks of iterative development per task. This estimate also includes the overhead of comparing alternative primitives and updating implementations as the underlying toolchain evolved (e.g., the Molmo $\rightarrow$ Molmo2 transition).
\end{itemize}

\textbf{Comparison to multi-turn agent recovery.} In contrast, \sysname{}'s multi-turn recovery is fast at evaluation time: a complete evaluation of \sysname{} on a Robosuite task takes approximately 2 minutes per trial (see Appendix~\ref{app:compute_cost}). The intended comparison is therefore between \emph{static, single-turn code iterated on extensively by an experienced human} and an \emph{interactive agent capable of online monitoring and recovery at deployment time}---which is precisely the practical gap that \sysname{} is designed to address.

\subsection{Computational Cost Analysis}
\label{app:compute_cost}
We report code-generation time and total trial time (LLM + execution) averaged over $N{=}20$ trials of cube stacking using Gemini-3-Pro. All measurements were collected on the same hardware with identical perception and control stacks.

\begin{table}[h]
\centering
\small
\begin{tabular}{lcc}
\toprule
Tier & Avg LLM Code-Gen Time & Avg Trial Time \\
\midrule
S1 (privileged) & 6.8 s & 12.9 s \\
S2 (non-privileged) & 9.6 s & 20.0 s \\
S3 (reduced API) & 23.8 s & 28.4 s \\
M1 (multi-turn, no VDM) & 20.1 s & 60.8 s \\
M3 (multi-turn + VDM) & 15.8 s & 113.6 s \\
\bottomrule
\end{tabular}
\caption{Per-trial timing breakdown across CaP-Bench tiers (Gemini-3-Pro, cube stack, $N{=}20$).}
\label{tab:compute_cost}
\end{table}

\textbf{Comparison to VLAs.} For reference, $\pi_0$~\cite{black2024pi0visionlanguageactionflowmodel} reports $\sim$73~ms per action chunk and OpenVLA~\cite{kim2024openvlaopensourcevisionlanguageactionmodel} reports $\sim$167~ms per action. VLAs are substantially faster per inference step because they emit single motor commands at high frequency. \sysname{} operates at a different level of abstraction: each code-generation iteration (6.8--23.8~s) produces an \emph{entire manipulation sequence}, not a single action. Direct latency comparison is therefore misleading; the relevant axis is \emph{cost per task attempt} rather than \emph{cost per action}.

\subsection{Visual Differencing Module: Model Choice}
\label{app:vdm_choice}
The Visual Differencing Module (VDM) used throughout the main paper is implemented with Gemini-3-Pro~\cite{google_deepmind_gemini3pro_modelcard_2025}, which at the time of writing was the state-of-the-art on major multimodal benchmarks (MMMU-Pro: $81\%$; Video-MMMU: $87.6\%$). We choose the strongest available VLM to upper-bound the effect of language-grounded perception, ensuring that the observed performance gains in M3/M4 reflect the value of the \emph{mechanism} (text grounding via an auxiliary VLM) rather than weakness of an under-powered captioner.

The VDM is \emph{model-agnostic by design}: swapping the backbone requires a single configuration change in CaP-Gym, with no modifications to the coding-agent loop or to the rest of the benchmark. A systematic ablation across VDM backbones is an orthogonal axis of study that CaP-Bench supports out of the box and that we encourage the community to explore; the contribution of this paper is to demonstrate that VDM, as a \emph{method}, materially helps multi-turn recovery.

\end{document}